\documentclass[5p,sort,compress]{elsarticle}
\usepackage{graphicx}
\usepackage{amsmath, amssymb}
\DeclareMathOperator*{\concat}{\parallel}
\usepackage{caption, subcaption}
\usepackage{float}
\usepackage{multirow}
\usepackage[colorlinks=true,linkcolor=black,citecolor=black,urlcolor=black]{hyperref}
\usepackage{pgf-pie}
\usepackage{soul,color}
\usepackage{xcolor}
\usepackage{placeins}
\usepackage[normalem]{ulem}
\definecolor{darkgreen}{rgb}{0,.4,0}
\definecolor{darkcyan}{rgb}{0,.4,.4}
\newcommand{\REMOVE}[1]%
{{\color{blue}\sout{#1}}}

\newcommand{\COMMENT}[1]%
{{\color{darkgreen}\textbf{{Editor: }} {#1}}}
\begin{document}
\begin{frontmatter}
\title{Spatial and Spatial-Spectral Morphological Mamba for Hyperspectral Image Classification}
\author[1]{Muhammad Ahmad}
\author[2]{Muhammad Hassaan Farooq Butt}
\author[3]{Adil Mehmood Khan}
\author[4]{Manuel Mazzara}
\author[1]{Salvatore Distefano}
\author[5]{Muhammad Usama}
\author[6]{Swalpa Kumar Roy}
\author[7]{Jocelyn Chanussot}
\author[8]{Danfeng Hong\corref{cor}}
\ead{hongdf@aircas.ac.cn}

\address[1]{Dipartimento di Matematica e Informatica-MIFT, University of Messina, 98121 Messina, Italy. mahmad00@gmail.com; sdistefano@unime.it}
\address[2]{Institute of Artificial Intelligence, School of Mechanical and Electrical Engineering, Shaoxing University, Shaoxing 312000, China. (e-mail: hassaanbutt67@gmail.com)}
\address[3]{School of Computer Science, University of Hull, Hull HU6 7RX, UK. (e-mail: a.m.khan@hull.ac.uk)}
\address[4]{Institute of Software Development and Engineering, Innopolis University, 420500 Innopolis, Russia. (e-mail: m.mazzara@innopolis.ru)}
\address[5]{M. Usama is with the Department of Computer Science, National University of Computer and Emerging Sciences, Islamabad, Chiniot-Faisalabad Campus, Chiniot 35400, Pakistan. (e-mail: m.usama@nu.edu.pk)}
\address[6]{Department of Computer Science and Engineering, Alipurduar Government Engineering and Management College, West Bengal 736206, India (e-mail: swalpa@agemc.ac.in)}
\address[7]{Univ. Grenoble Alpes, Inria, CNRS, Grenoble INP, LJK, Grenoble, 38000, France, and also with the Aerospace Information Research Institute, Chinese Academy of Sciences, 100094 Beijing, China. (e-mail: jocelyn.chanussot@inria.fr)}
\address[8]{Aerospace Information Research Institute, Chinese Academy of Sciences, Beijing, 100094, China, and also with the School of Electronic, Electrical and Communication Engineering, University of Chinese Academy of Sciences, 100049 Beijing, China. (e-mail: hongdf@aircas.ac.cn)}
\cortext[cor]{Corresponding author}
\begin{abstract}
Recent advancements in transformers, specifically self-attention mechanisms, have significantly improved hyperspectral image (HSI) classification. However, these models often suffer from inefficiencies, as their computational complexity scales quadratically with sequence length. To address these challenges, we propose the morphological spatial mamba (SMM) and morphological spatial-spectral Mamba (SSMM) model (MorpMamba), which combines the strengths of morphological operations and the state space model framework, offering a more computationally efficient alternative to transformers. In MorpMamba, a novel token generation module first converts HSI patches into spatial-spectral tokens. These tokens are then processed through morphological operations such as erosion and dilation, utilizing depthwise separable convolutions to capture structural and shape information. A token enhancement module refines these features by dynamically adjusting the spatial and spectral tokens based on central HSI regions, ensuring effective feature fusion within each block. Subsequently, multi-head self-attention is applied to further enrich the feature representations, allowing the model to capture complex relationships and dependencies within the data. Finally, the enhanced tokens are fed into a state space module, which efficiently models the temporal evolution of the features for classification. Experimental results on widely used HSI datasets demonstrate that MorpMamba achieves superior parametric efficiency compared to traditional CNN and transformer models while maintaining high accuracy. The code will be made publicly available at \url{https://github.com/mahmad000/MorpMamba}.
\end{abstract}
\end{frontmatter}
\section{\textbf{Introduction}}
\label{Intr}

Hyperspectral Image (HSI) classification plays a critical role in a wide array of applications, including remote sensing \cite{ahmad2021hyperspectral}, Earth observation \cite{hong2024multimodal}, urban planning \cite{li2024HD}, agriculture \cite{deng2024rustqnet}, and environmental monitoring \cite{hong2023cross, pande2023application}. The ability of HSIs to capture detailed spectral information across a wide range of wavelengths provides insights that traditional imaging cannot, enabling precise material identification and classification across these domains. However, effectively analyzing the high-dimensional data inherent in HSIs poses significant challenges, particularly in terms of developing algorithms that can manage and interpret the vast spectral and spatial information without overwhelming computational resources \cite{hong2024spectralgpt, 10647404}.

Recent advances in deep learning, particularly convolutional neural networks (CNNs), have shown promise in extracting meaningful spatial and spectral features from HSIs \cite{ahmad2020fast, hong2023decoupled}. While CNNs can learn hierarchical representations crucial for HSI classification, they are limited by their local receptive fields, which fail to capture the global spatial context needed for complex classification tasks \cite{li2024learning, jaiswal2021critical}. This limitation often results in suboptimal performance, especially in high-dimensional hyperspectral data. Moreover, CNNs require large labeled datasets for effective training, which is a significant constraint given the scarcity of annotated HSI datasets \cite{10423094}.

Transformer architectures, leveraging self-attention mechanisms, have emerged as a promising alternative due to their ability to model long-range dependencies and global contextual relationships \cite{10604879}. This has led to notable improvements in HSI classification by capturing the intricate relationships between spectral bands and spatial regions \cite{yao2023extended, wu2021convolutional, 10685113}. However, transformers also introduce substantial computational complexity, as their operations scale quadratically with the sequence length, making them less practical for processing large-scale HSI data, which are inherently high-dimensional and require extensive computational resources \cite{9868046, 10681622}.

To address the limitations of both CNNs and transformers, the Mamba architecture, based on the state space model (SSM), has emerged as a more efficient alternative for sequence modeling \cite{gu2023mamba}. Mamba replaces the attention mechanism with a state space formulation, achieving linear complexity scaling with sequence length and offering substantial computational savings \cite{mamba-explained}. This makes Mamba particularly well-suited for HSI classification, where efficiently managing long spectral sequences is essential \cite{10767233}.

Several recent studies have applied the Mamba framework to HSI classification. SpectralMamba \cite{yao2024spectralmamba} introduced a gated spatial-spectral merging (GSSM) module and piece-wise sequential scanning (PSS) strategy to address inefficiencies in sequence modeling. Similarly, spatial-spectral Mamba (SSMamba) \cite{huang2024spectral} incorporated a spectral-spatial token generation module to fuse spatial and spectral information effectively. While these architectures improved classification performance, they face challenges in handling extremely high-dimensional HSI data and generalizing across diverse datasets. Wang et al. \cite{wang2024s} introduced the \(S^2\)Mamba architecture for HSI classification, which integrates spatial and spectral features using SSMs. This model employs Patch cross-scanning and Bi-directional spectral scanning to capture spatial and spectral features, respectively, merging them with a spatial-spectral mixture gate to enhance classification accuracy. He et al. \cite{he20243dss} developed the 3D spectral-spatial Mamba (3DSSMamba) architecture for HSI classification. This framework utilizes the strengths of the Mamba architecture by incorporating a spectral-spatial token generation module (SSTG) and a 3D spectral-spatial selective scanning (3DSS) mechanism, enabling the model to effectively capture global spectral-spatial contextual dependencies while maintaining linear computational complexity. Despite improved classification performance, the authors stressed the need for further optimization to robustly handle high-dimensional data and explore additional strategies for better generalization across diverse HSI datasets.

Sheng et al. \cite{sheng2024dualmamba} introduced DualMamba, a spatial-spectral Mamba architecture for HSI classification. This design integrates Mamba with CNNs, effectively capturing complex spectral-spatial relationships while ensuring computational efficiency. However, the authors noted limitations, including redundancy in multi-directional scanning strategies and challenges in fully leveraging spectral information. Similarly, Zhou et al. \cite{zhou2024mamba} developed another spatial-spectral Mamba architecture, featuring a centralized Mamba-Cross-Scan mechanism that transforms HSI data into diverse sequences, enhancing feature extraction through a Tokenized Mamba encoder. Despite its strengths, this method is sensitive to variations in peripheral pixels and requires significant computational resources for larger patches. Yang et al. \cite{yang2024graphmamba} proposed GraphMamba, which enhances spatial-spectral feature extraction through components like HyperMamba and SpatialGCN, addressing efficiency and contextual awareness. Nevertheless, the authors identified shortcomings, such as optimizing encoding modules to accommodate diverse HSI datasets and potential overfitting in high-dimensional spaces.

To overcome the limitations of the existing models, this paper proposes a morphological spatial and spatial-spectral Mamba (MorpMamba) architecture, which integrates morphological operations into the Mamba framework. Morphological operations, specifically erosion, and dilation, are well-suited for capturing structural and shape-related features in spatial-spectral data. These operations enhance the tokenization process by emphasizing boundaries, filling gaps, and smoothing out noise, leading to more robust and meaningful token representations. In a nutshell, the following contributions are made in this study. 

\textbf{Firstly}, erosion highlights the boundaries of objects, enabling better distinction between different regions in HSIs, while dilation enhances structural continuity by connecting disjoint parts. These morphological operations effectively reduce noise and extract prominent spatial-spectral features. By incorporating them in the token generation process, MorpMamba ensures that tokens represent both fine details and global structures. \textbf{Secondly}, the token enhancement module further refines the tokens by dynamically adjusting the spatial and spectral features based on the central regions of the HSI. This results in more context-aware and stable token representations, making the model less sensitive to noise and improving feature robustness. \textbf{Lastly}, the multi-head self-attention mechanism and state space model (SSM) work in tandem to capture long-range dependencies and efficiently model the temporal evolution of features. The self-attention mechanism focuses on different aspects of the spatial-spectral features, while the state space model ensures efficient and interpretable feature progression, contributing to superior classification accuracy.

In short, MorpMamba leverages the strengths of morphological operations, token enhancement, and the Mamba framework to create a robust, efficient, and scalable model for HSI classification. The combination of these techniques leads to better feature extraction, reduced computational complexity, and improved generalization across diverse HSI datasets.

\begin{figure*}[!hbt]
	\centering
	\includegraphics[width=0.98\textwidth]{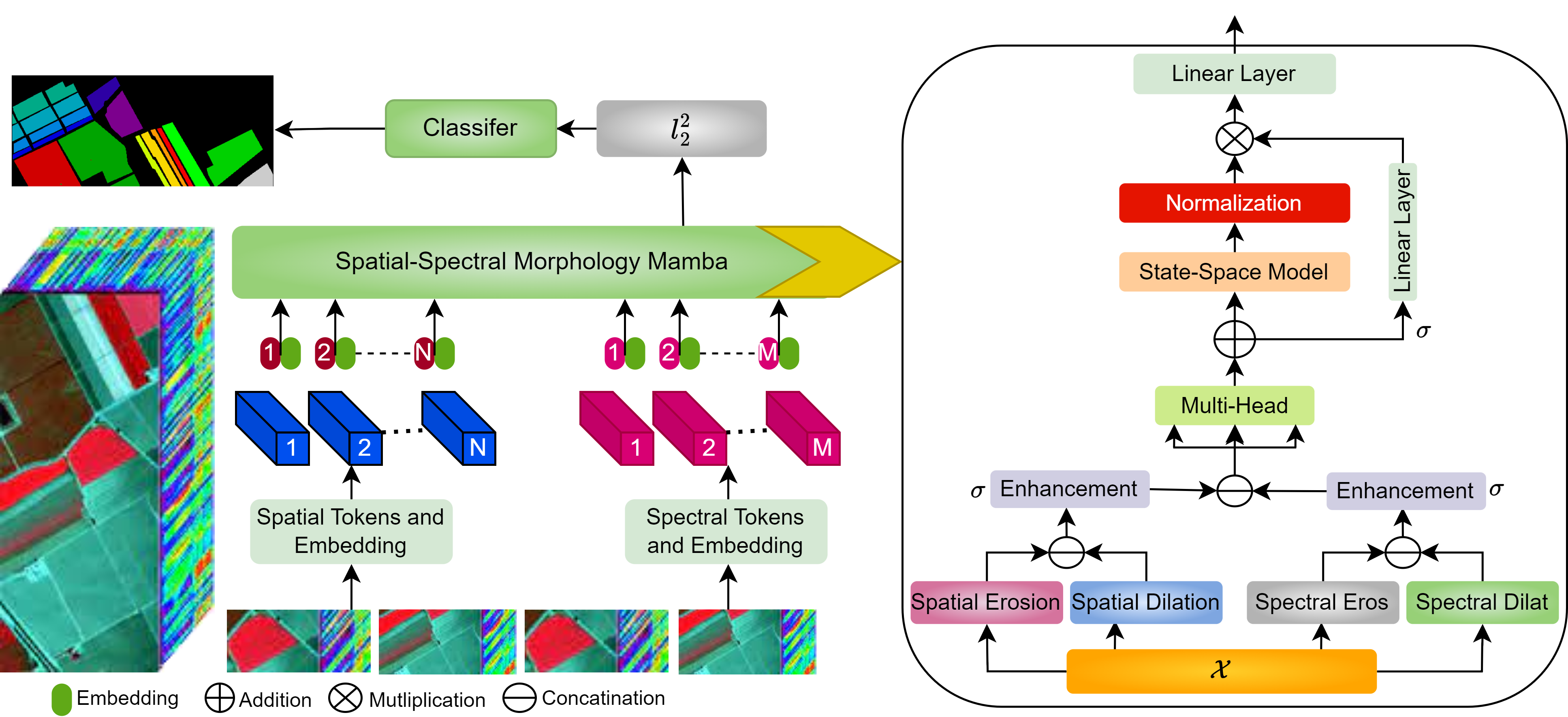}
        \caption{A joint spatial-spectral feature token is first computed from the HSI using morphological operations. These tokens are then integrated into the MorpMamba model, which includes Erosion and Dilation operations, token enhancement, and a multi-head attention module. This method allows a more selective and effective representation of information as compared to standard fixed-dimension encodings. The output is then processed through an SSM, followed by feature normalization and a linear layer with $l_2^2$ regularization. Finally, this output is passed to the classification head for generating the ground truth.}
	\label{Fig1}
\end{figure*}

\section{Proposed Methodology}
\label{PM}

Given the HSI data $\mathcal{X} \in \mathcal{R}^{H, W, C}$, where $H$ and $W$ represent the spatial dimensions (height and width), and $C$ denotes the number of spectral bands, the goal is to classify pixels using both spectral and spatial information. We divide the HSI cube $\mathcal{X}$ into overlapping 3D patches, each capturing spatial-spectral data for further processing. 

\begin{equation}
    N = \bigg(\frac{H}{P} \times \frac{W}{P}\bigg)
\end{equation}
where $\mathcal{X} \in \mathcal{R}^{N \times (P \times P \times C)}$, where $N$ is the total number of 3D patches, and each patch of size $P \times P \times C$ is used as input to subsequent modules. Spatial and spectral patches are processed independently using morphological operations (erosion and dilation) to extract structural features. The complete model structure is presented in Figure \ref{Fig1}. 

\subsection{Morphological Operations}

Morphological operations are employed to refine spatial and spectral information from HSI patches. Erosion reduces the shape of objects and eliminates minor details, while dilation expands object boundaries and enhances structural integrity. These operations enable the extraction of both fine details and large-scale features. In our model, morphological operations are applied separately to both spatial and spectral dimensions, which are processed independently. These operations help refine the spatial and spectral structures in HSI by highlighting boundaries and enhancing structural features. The erosion operation on a patch $\mathcal{X}$ is defined as:

\begin{equation}
    \varepsilon_\mathbf{k}(\mathcal{X}) = \min_{\mathbf{i} \in \mathcal{N}(\mathbf{j})} {\mathcal{X}(\mathbf{i}) \boxminus \mathbf{k}(\mathbf{i} - \mathbf{j})}
    \label{Eq1}
\end{equation}
where $\mathcal{N}(\mathbf{j})$ is the neighborhood of pixels $\textbf{j}$, $\boxminus$ denotes element-wise subtraction, and $\mathbf{k}$ is the structuring element (SE). The dilation operation is defined as:

\begin{equation}
    \delta_\mathbf{k}(\mathcal{X}) = \max_{\mathbf{i} \in \mathcal{N}(\mathbf{j})} {\mathcal{X}(\mathbf{i}) \boxplus \mathbf{k}(\mathbf{i} - \mathbf{j})}
    \label{Eq2}
\end{equation}
where $\boxplus$ denotes element-wise addition, allowing dilation to expand the foreground object in the patch. These operations are performed using depthwise separable convolutions, with kernels representing SEs, applied separately along spatial and spectral dimensions. In other words, the SE kernel's size affects the apparent texture size for various regions within the patch and token. The erosion (Equation \ref{Eq1}) and dilation (Equation \ref{Eq2}) operations are performed per channel using depthwise 2D convolution layers with a kernel of size $(5 \times 5)$ and the same padding, initialized with weights set to one (representing the SE) and the sign is inverted. 

\subsection{Spatial-Spectral Token Generation}

Once morphological operations are applied, we generate spatial and spectral tokens separately. These tokens are formed by concatenating the results of the erosion and dilation operations. For spatial token generation, the operations are applied along the height and width dimensions:

\begin{equation}
    \mathcal{X}_{\text{spatial}}^\text{eroded} = \varepsilon\mathbf{k}(\mathcal{X})~,~ \mathcal{X}_{\text{spatial}}^\text{dilated} = \delta\mathbf{k}(\mathcal{X})
\end{equation}

The eroded and dilated spatial features are then concatenated and passed through a depthwise convolution to produce spatial tokens:

\begin{equation}
    \mathbf{t}_{\text{spatial}} = \text{Conv2D$_\text{Depthwise}$}(\concat(\mathcal{X}_{\text{spatial}}^\text{eroded}, \mathcal{X}_{\text{spatial}}^\text{dilated}), \mathcal{W})
\end{equation}

For spectral token generation, the input data is transposed to treat the spectral channels as the spatial dimension, and similar morphological operations are applied:

\begin{equation}
    \mathcal{X}_{\text{spectral}}^\text{eroded} = \varepsilon\mathbf{k}(\mathcal{X}^\top)~, ~ \mathcal{X}_{\text{spectral}}^\text{dilated} = \delta\mathbf{k}(\mathcal{X}^\top)
\end{equation}

The concatenated eroded and dilated features are passed through a depthwise convolution to generate spectral tokens:

\begin{equation}
    \mathbf{t}_{\text{spectral}} = \text{Conv2D$_\text{Depthwise}$}(\concat(\mathcal{X}_{\text{spectral}}^\text{eroded}, \mathcal{X}_{\text{spectral}}^\text{dilated}), \mathcal{W})
\end{equation}

The final output from the token generation module consists of the spatial and spectral tokens $(\mathbf{t}_{\text{spatial}}, \mathbf{t}_{\text{spectral}})$, providing a comprehensive spatial-spectral representation of the HSI data.

\subsection{Token Enhancement and Multi-head Attention}

To refine the tokens generated, a gating mechanism is applied, which adjusts the spatial and spectral tokens based on contextual information from the center region of the HSI patch. Specifically, the model extracts center tokens from the spatial tokens and uses them to modulate both the spatial and spectral token importance:

\begin{figure}[!hbt]
	\centering
	\includegraphics[width=0.48\textwidth]{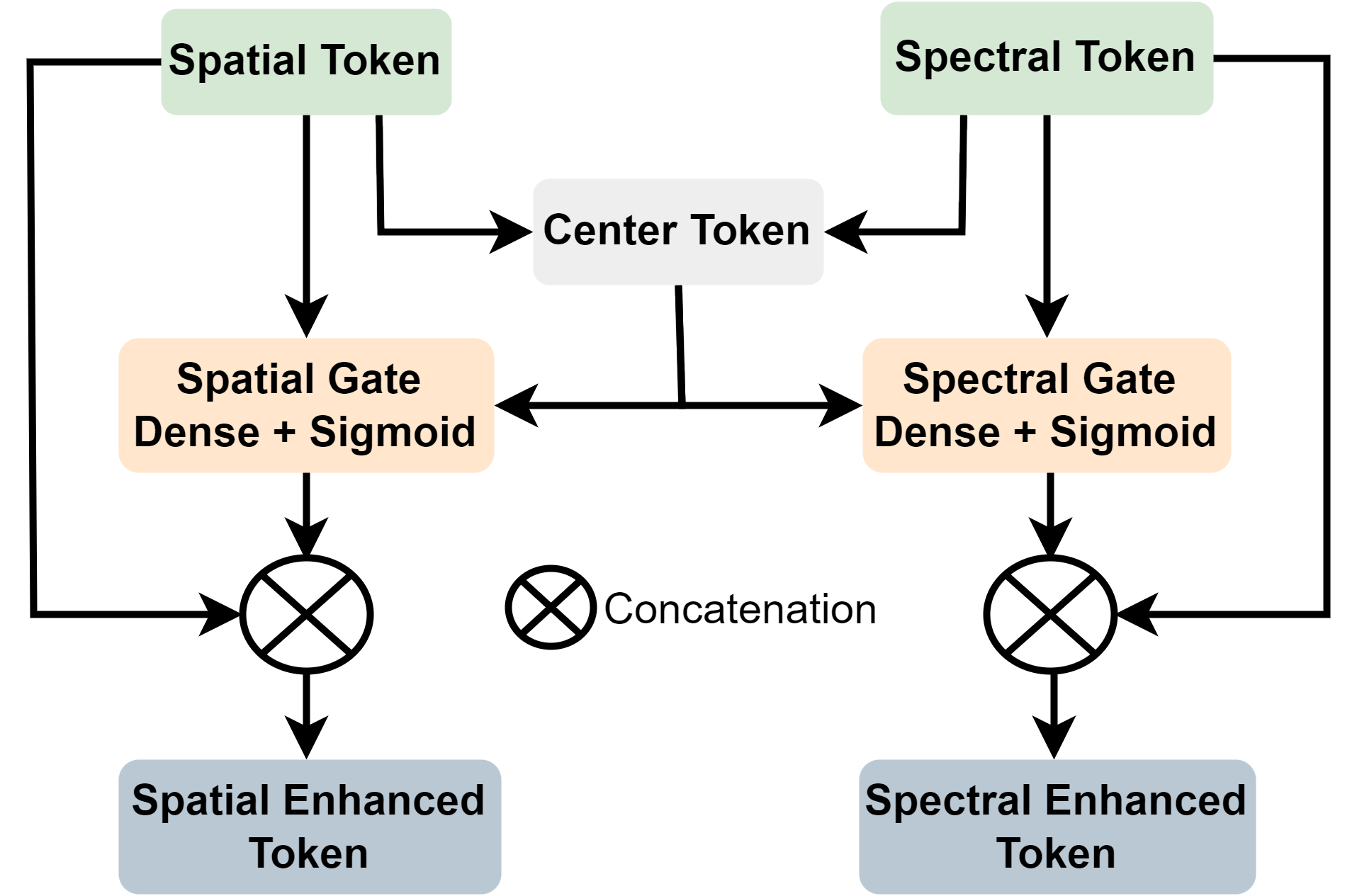}
	\caption{Spatial-Spectral token enhancement module adopted to refine and enhancement of the extracted spatial and spectral features.}
	\label{FigB}
\end{figure}

\begin{equation}
\widetilde{\mathbf{t}}_{\text{spectral}}^{(l)} = \mathbf{t}_{\text{spectral}}^{(l)} \odot \sigma(\mathbf{W}_{spectral}~ \mathbf{c} + \mathbf{b}_{\text{spectral}})
\end{equation}

\begin{equation}
\widetilde{\mathbf{t}}_{\text{spatial}}^{(l)} = \mathbf{t}_{\text{spatial}}^{(l)} \odot \sigma(\mathbf{W}_{spatial}~ \mathbf{c} + \mathbf{b}_{\text{spatial}})
\end{equation}
where $\mathbf{W}_{spectral}$ and $\mathbf{W}_{spatial}$ are weight matrices, $\mathbf{c}$ is the center region of the patch, and $\sigma$ denotes the sigmoid function as shown in Figure \ref{FigB}. After this enhancement, a multi-head self-attention mechanism is applied, which allows the model to focus on different regions of the data simultaneously, further refining the tokenized features:

\begin{equation}
A_i = \text{softmax}\left(\frac{\mathbf{Q}_i \mathbf{K}_i^\top}{\sqrt{d_k}}\right)~,~O_i = A_i \textbf{V}_i
\end{equation}
where $\textbf{Q}_i = \widetilde{\mathbf{t}}_{\text{spectral}}^{(l)} W^Q_i$, $\textbf{K}_i = \widetilde{\mathbf{t}}_{\text{spatial}}^{(l)} W^K_i$, and $\textbf{V}_i = \widetilde{\mathbf{t}}_{\text{spatial}}^{(l)} W^V_i$ are the query, key, and value projections of the tokens, respectively, and $O_i$ represents the output for each attention head. The combined output is a refined feature representation across spatial-spectral dimensions as shown in Figure \ref{Fig3}.

\begin{figure}[!hbt]
	\centering
	\includegraphics[width=0.48\textwidth]{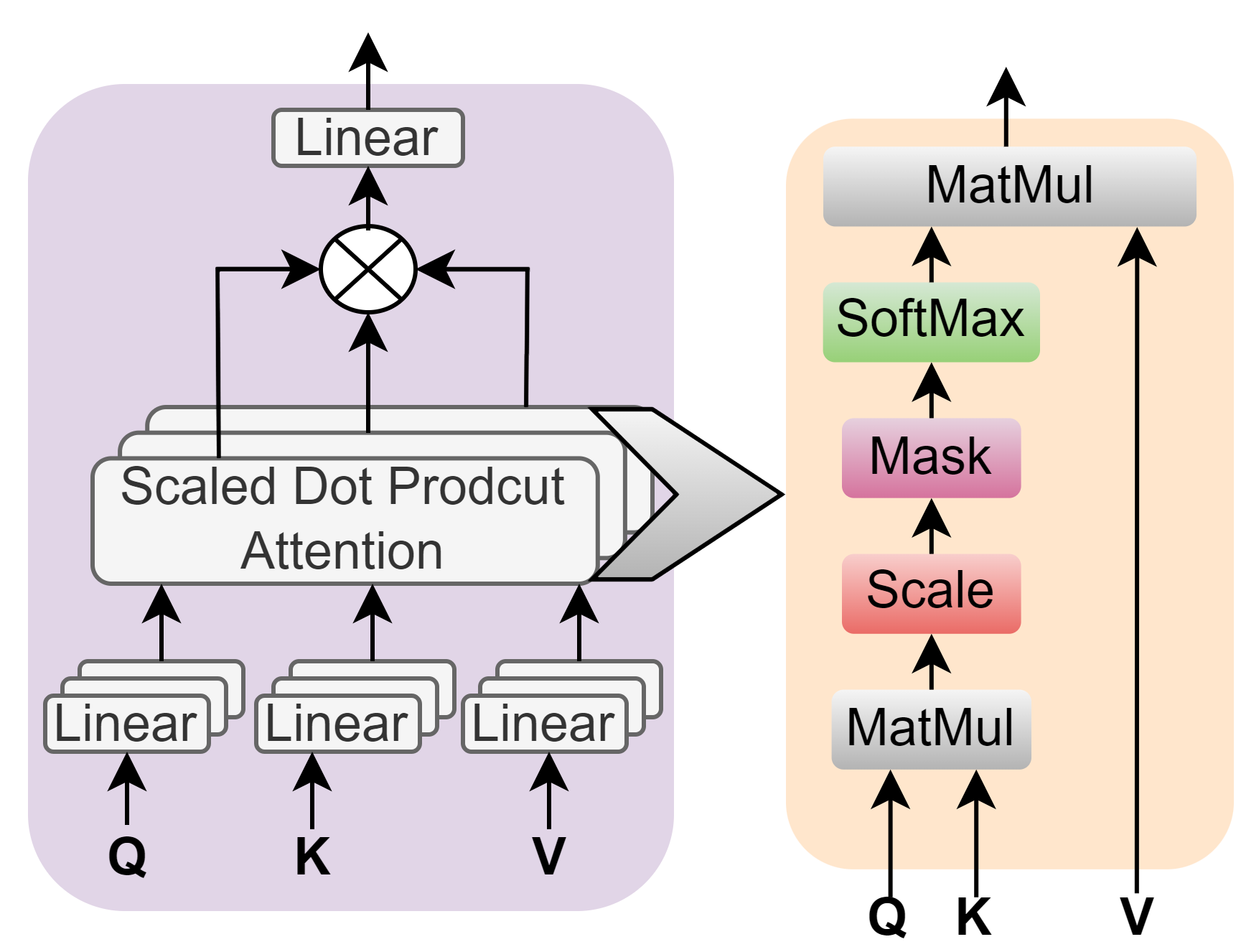}
	\caption{Multi-head self-attention module adapted to interact with the enhanced spatial and spectral features.}
	\label{Fig3}
\end{figure}

\subsection{State Space Model (SSM)}

The final stage involves processing the enhanced tokens through a state space model (SSM), which captures long-term dependencies and models the temporal evolution of the features:

\begin{equation}
h_t = \text{ReLU}(W_{\text{trans}} h_{t-1} + W_{\text{update}} E_t)
\end{equation}
where $W_{\text{trans}}$ and $W_{\text{update}}$ are learned weights, and $h_t$ is the hidden state at time $t$. The final classification output is produced using a linear classifier with $l_2^2$ regularization:

\begin{equation}
y = \sigma(h_t W_{\text{classifier}} - \lambda |W_{\text{classifier}}|_2^2)
\end{equation}
where $\lambda$ controls the regularization strength. By integrating morphological tokenization, token enhancement, multi-head attention, and state space modeling, MorpMamba efficiently combines spatial and spectral information, leading to improved classification performance with reduced computational complexity.

\section{Experimental Datasets}
\label{ED}

To evaluate the performance of MorpMamba, we utilized several widely-used hyperspectral image (HSI) datasets: WHU-Hi-LongKou (LK), Pavia University (PU), Pavia Centre (PC), Salinas (SA), and the University of Houston (UH). These datasets span various geographic locations, sensor types, and spatial-spectral resolutions, providing a comprehensive benchmark for model evaluation. Table \ref{Tab.1} summarizes the key characteristics of each dataset.

\begin{table}[!hbt] 
\centering 
\caption{Summary of the HSI datasets used for experimental evaluation.} 
\resizebox{\columnwidth}{!}{\begin{tabular}{l||c||c||c||c||c} \hline 
--- & \textbf{SA} & \textbf{PU} & \textbf{PC} & \textbf{UH} & \textbf{LK}\\ \hline 
\textbf{Source} & AVIRIS & ROSIS-03 & ROSIS-03 & CASI & UAV \\ 
\textbf{Spatial} & $512 \times 217$ & $610 \times 610$ & $1096 \times 1096$ & $340 \times 1905$ & $550 \times 400$ \\
\textbf{Spectral Bands} & 224 & 103 & 102 & 144 & 270 \\
\textbf{Wavelength (nm)} & 350-1050 & 430-860 & 430-860 & 350-1050 & 400-1000 \\
\textbf{Ground Samples} & 111,104 & 372,100 & 1,201,216 & 647,700 & 220,000 \\
\textbf{Classes} & 16 & 9 & 9 & 15 & 9 \\
\textbf{Resolution (m)} & 3.7 & 1.3 & 1.3 & 2.5 & 0.463 \\ \hline 
\end{tabular}} 
\label{Tab.1} 
\end{table} 

The\textbf{ WHU-Hi-LongKou (LK)} dataset was collected in Longkou Town, Hubei, China, using a Headwall Nano-Hyperspec sensor mounted on a DJI M600 Pro UAV. Captured in July 2018 at an altitude of 500 meters, the dataset consists of 550 $\times$ 400 pixels with 270 spectral bands ranging from 400 to 1000 nm. The scene includes six crop types: corn, cotton, sesame, broad-leaf soybean, narrow-leaf soybean, and rice. The high-resolution dataset has a spatial resolution of 0.463 meters per pixel, making it suitable for fine-grained agricultural monitoring.

The \textbf{Salinas (SA)} dataset, collected by the AVIRIS sensor over Salinas Valley, California, consists of 512$\times$217 pixels with 224 spectral bands covering the 0.35 to 1.05 $\mu$m wavelength range. The dataset includes diverse land covers such as vegetables, bare soils, and vineyards, with ground truth labels for 16 distinct classes. Its 3.7-meter spatial resolution makes it ideal for high-precision agricultural and environmental monitoring.

Both \textbf{Pavia University (PU)} and \textbf{Pavia Center (PC)} datasets were acquired using the ROSIS-03 sensor over northern Italy. The Pavia University scene contains 610$\times$610 pixels with 103 spectral bands, while the Pavia Centre is a larger 1096$\times$1096-pixel image with 102 spectral bands. Both datasets feature a spatial resolution of 1.3 meters and differentiate between nine land-cover classes. These datasets are frequently used benchmarks for urban land-cover classification tasks.

The \textbf{University of Houston (UH)} dataset, published as part of the IEEE GRSS Data Fusion Contest, was captured using the Compact Airborne Spectrographic Imager (CASI). This dataset spans 340$\times$1905 pixels with 144 spectral bands, covering wavelengths from 0.38 to 1.05 $\mu$m. The spatial resolution of 2.5 meters per pixel, combined with 15 land-cover classes, provides a challenging dataset for urban and land-cover classification.

\section{Ablation Study and Discussion}
\label{QRD1}

This section outlines a series of experiments designed to evaluate MorpMamba's performance across various scenarios.

\begin{enumerate}
    \item Without morphological operation (NM)
    \item Only spatial morphology (SMM)
    \item joint spatial-spectral morphology within the Mamba model (SSMM)
    \item Training sample ratios (1\%, 2\%, 5\%, 10\%, 15\%, 20\%, and 25\%). 5)
    \item Different patch sizes $2 \times 2$, $4 \times 4$, $6 \times 6$, $8 \times 8$, and $10 \times 10$)
    \item Number of Attention heads (2, 4, 6, and 8)
    \item Kernel sizes ($3 \times 3$, $5 \times 5$, $7 \times 7$, $9 \times 9$, and $11 \times 11$) for morphological operations.
    \item Computational time for Training samples, patch sizes, number of heads, and kernel sizes. 
    \item t-SNE feature representations.
\end{enumerate}

The model with spatial morphological operations is referred to as Spatial morphological Mamba (SMM), and the model with spatial-spectral morphological operations is referred to as SSMM, while the version without these operations is called SSMamba (token generation, enhancement, and multi-head attention component remains intact). The study aims to evaluate the impact of morphological operations on the overall accuracy (OA), average accuracy (AA), and kappa ($\kappa$) coefficient across various datasets.

Table \ref{Tab2} present the results for the Mamba model variants: No Morphology (NM--SSMamba), SMM, and SSMM across multiple datasets. As shown in the Table, SSMM consistently outperforms the other models. However, it is important to highlight the performance of SMM, which demonstrates competitive results as well. For instance, in the PU dataset, SMM achieved an OA of 96.52\%, which is an improvement over SSMamba (95.67\%) and just slightly lower than SSMM (97.67\%). SMM also demonstrated a strong $\kappa$ coefficient of 95.40\%, compared to 94.25\% for NM, although still trailing behind SSMM's 96.91\%. In the PC dataset, SMM again performed well with an OA of 99.52\%, which is marginally higher than NM (99.48\%) but slightly below SSMM’s 99.71\%. Similarly, the $\kappa$ coefficient for SMM in this dataset was 99.32\%, higher than NM (99.27\%) and just under SSMM (99.59\%). These results demonstrate that even spatial morphological operations alone (without the spectral component) significantly enhance model performance over the baseline No Morphology model.

\begin{table*}[!hbt]
    \centering
    \caption{Results of the Mamba model without Morphology (SSMamba -- NM), the Spatial Morphology Mamba (SMM), and the spatial-spectral Morphology Mamba (SSMM). Each of these models is trained using a $4 \times 4$ patch size and 20\% of the training samples.}
    \resizebox{\textwidth}{!}{\begin{tabular}{c||ccc||ccc||c||ccc||ccc||ccc} \hline 
    
       \multirow{2}{*}{\textbf{Class}} & \multicolumn{3}{c||}{\textbf{Salinas}} & \multicolumn{3}{c||}{\textbf{University of Houston}} & \multirow{2}{*}{\textbf{Class}} & \multicolumn{3}{c||}{\textbf{Pavia University}} & \multicolumn{3}{c||}{\textbf{Pavia Centre}} & \multicolumn{3}{c}{\textbf{WHU-Hi-LongKou}} \\ \cline{2-7} \cline{9-17}
       
       & NM & SMM & SSMM & NM & SMM & SSMM && NM & SMM & SSMM & NM & SMM & SSMM & NM & SMM & SSMM \\ \hline  
       
        1 & 100 & 100 & 100 & 100 & 100 & 99 & 1 & 96 & 94 & 98 & 100 & 100 & 100 & 100 & 99 & 100 \\
        2 & 100 & 100 & 100 & 100  & 100 & 99 & 2 & 98 & 98 & 99 & 99 & 99 & 98 & 99 & 95 & 99 \\
        3 & 99 & 99 & 100 & 99 & 99 & 100 & 3 & 82 & 91 & 90 & 97 & 90 & 96 & 100 & 94 & 100 \\
        4 & 100 & 99 & 100 & 98 & 99 & 99 & 4 & 93 & 96 & 97 & 95 & 98 & 100 & 99 & 99 & 100 \\
        5 & 99 & 99 & 99 & 100 & 99 & 100 & 5 & 100 & 99 & 100 & 98 & 98 & 100 & 95 & 95 & 98 \\
        6 & 100 & 99 & 100 & 96 & 99 & 99 & 6 & 97 & 97 & 98 & 99 & 99 & 99 & 100 & 99 & 100 \\
        7 & 100 & 100 & 100 & 96 & 100 & 98 & 7 & 92 & 96 & 96 & 99 & 99 & 99 & 100 & 99 & 100 \\
        8 & 89 & 93 & 97 & 94 & 93 & 98 & 8 & 88 & 88 & 92 & 100 & 99 & 100 & 98 & 96 & 98 \\
        9 & 100 & 99 & 100 & 96 & 99 & 96 & 9 & 99 & 98 & 99 & 100 & 99 & 100 & 98 & 95 & 99 \\ \cline{8-17}
        10 & 99 & 98 & 99 & 99 & 98 & 99 & OA & 95.67 & 96.52 & \textbf{97.67} & 99.48 & 99.52 & \textbf{99.71} & 99.51 & 99.25 & \textbf{99.70} \\
        11 & 100 & 99 & 100 & 99 & 99 & 97 & AA & 93.20 & 95.94 & \textbf{96.93} & 98.60 & 98.28 & \textbf{98.95} & 98.45 & 97.53 & \textbf{99.25} \\
        12 & 100 & 100 & 100 & 95 & 100 & 98 & $\kappa$ & 94.25 & 95.40 & \textbf{96.91} & 99.27 & 99.32 & \textbf{99.59} & 99.36 & 99.02 & \textbf{99.61} \\  \cline{8-17}
        13 & 100 & 100 & 100 & 86 & 100 & 92 \\
        14 & 99 & 99 & 99 & 99 & 99 & 99 \\
        15 & 82 & 87 & 96 & 100 & 87 & 100 \\
        16 & 100 & 99 & 98 & --- & 99 & --- \\ \cline{1-7}
        OA & 95.20 & 96.78 & \textbf{98.52} & 90.36 & 96.78 & \textbf{98.28} \\
        AA & 97.81 & 98.48 & \textbf{99.25} & 90.54 & 97.48 & \textbf{97.91} \\
        $\kappa$ & 94.65 & 96.41 & \textbf{98.35} & 89.58 & 96.41 & \textbf{98.14} \\ \cline{1-7} 
    \end{tabular}}
    \label{Tab2}
\end{table*}
\begin{figure*}[!hbt]
    \centering
    \includegraphics[width=0.99\textwidth]{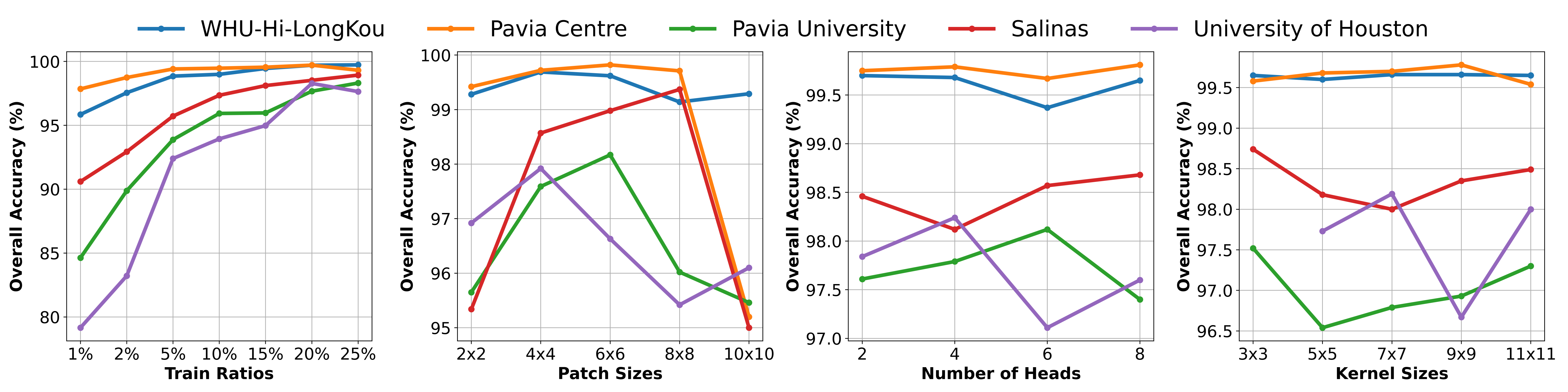}
    \caption{OA of MorpMamba across different training data ratios (1\%, 2\%, 5\%, 10\%, 15\%, 20\%, and 25\%), patch sizes ($4 \times 4$. Different patch sizes $2 \times 2$, $4 \times 4$, $6 \times 6$, $8 \times 8$, and $10 \times 10$), number of heads (2, 4, 6, and 8), and kernel sizes ($3 \times 3$, $5 \times 5$, $7 \times 7$, $9 \times 9$, and $11 \times 11$) over 50 epochs on WHU-Hi-LongKou, Pavia Centre, Pavia University, Salinas, and University of Houston datasets.}
    \label{overall_acc_sub_plots}
\end{figure*}
\begin{figure*}[!hbt]
    \centering
    \includegraphics[width=0.99\textwidth]{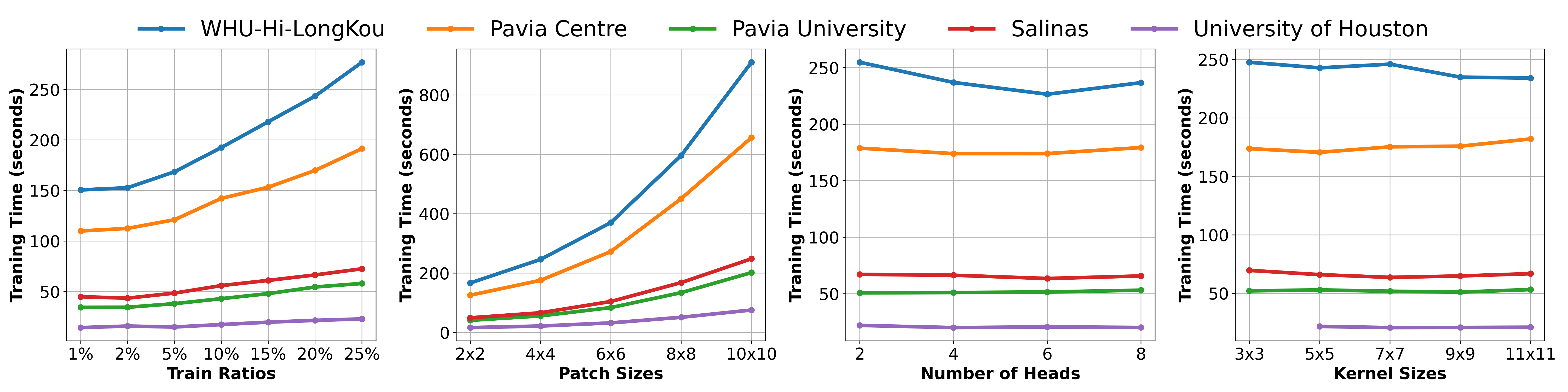}
    \caption{Training Time of MorpMamba across different training data ratios, patch sizes, number of heads, and kernel sizes over 50 epochs on WHU-Hi-LongKou, Pavia Centre, Pavia University, Salinas, and University of Houston datasets. The training ratio and patch size have a strong influence on computational time, whereas the head size within multi-head self-attention and the kernel size within morphological operations do not significantly affect the computational load. This demonstrates that the Mamba model maintains a linear computational load even after incorporating multi-head self-attention and morphological operations. However, these additions significantly improve performance, as shown in the subsequent sections.}
    \label{overall_time_sub_plots}
\end{figure*}
\begin{figure*}[!hbt]
    \centering
	\begin{subfigure}{0.32\textwidth}
		\includegraphics[width=0.99\textwidth]{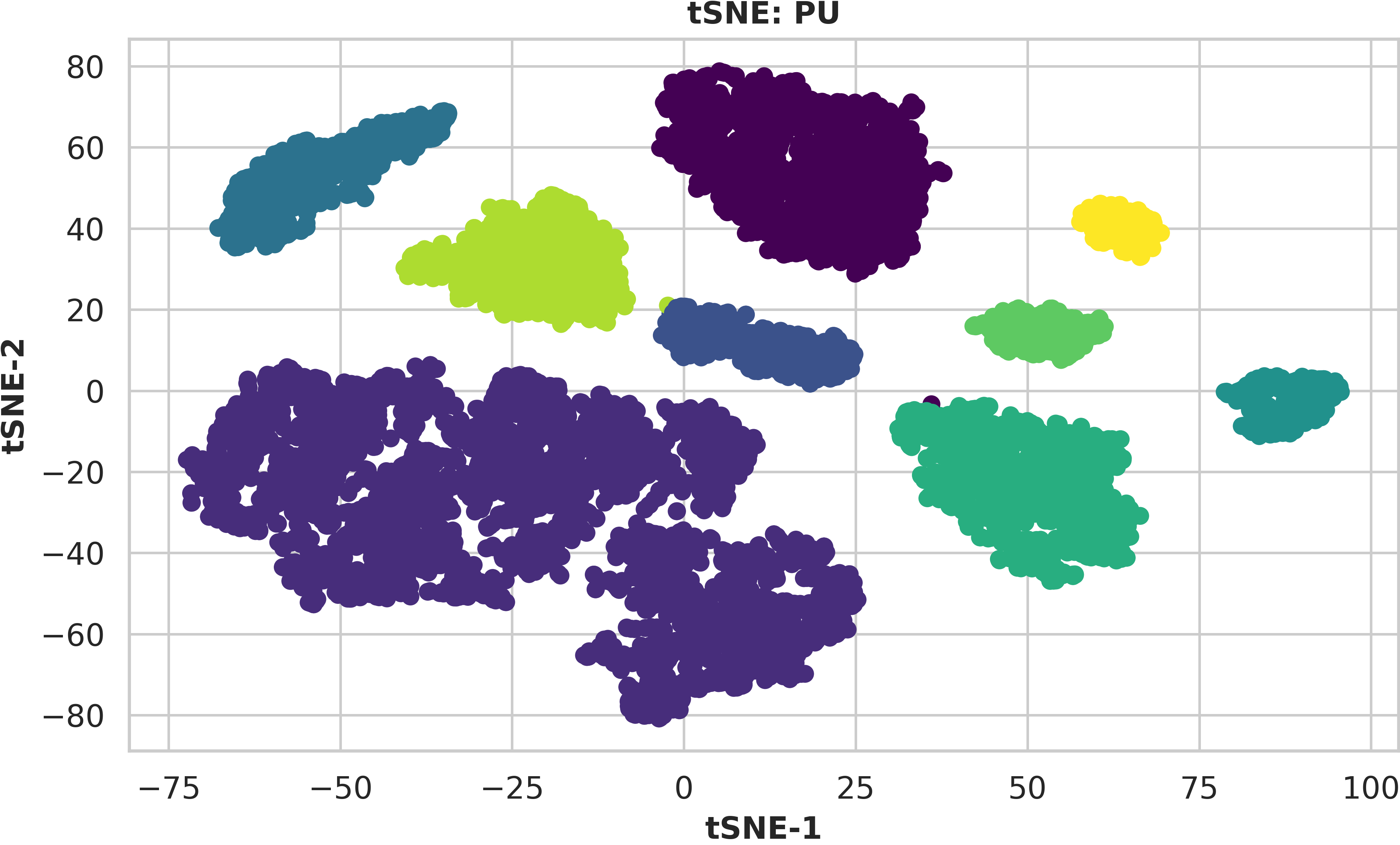}
		\caption{Pavia University} 
		\label{Fig2A}
	\end{subfigure}
    \begin{subfigure}{0.32\textwidth}
		\includegraphics[width=0.99\textwidth]{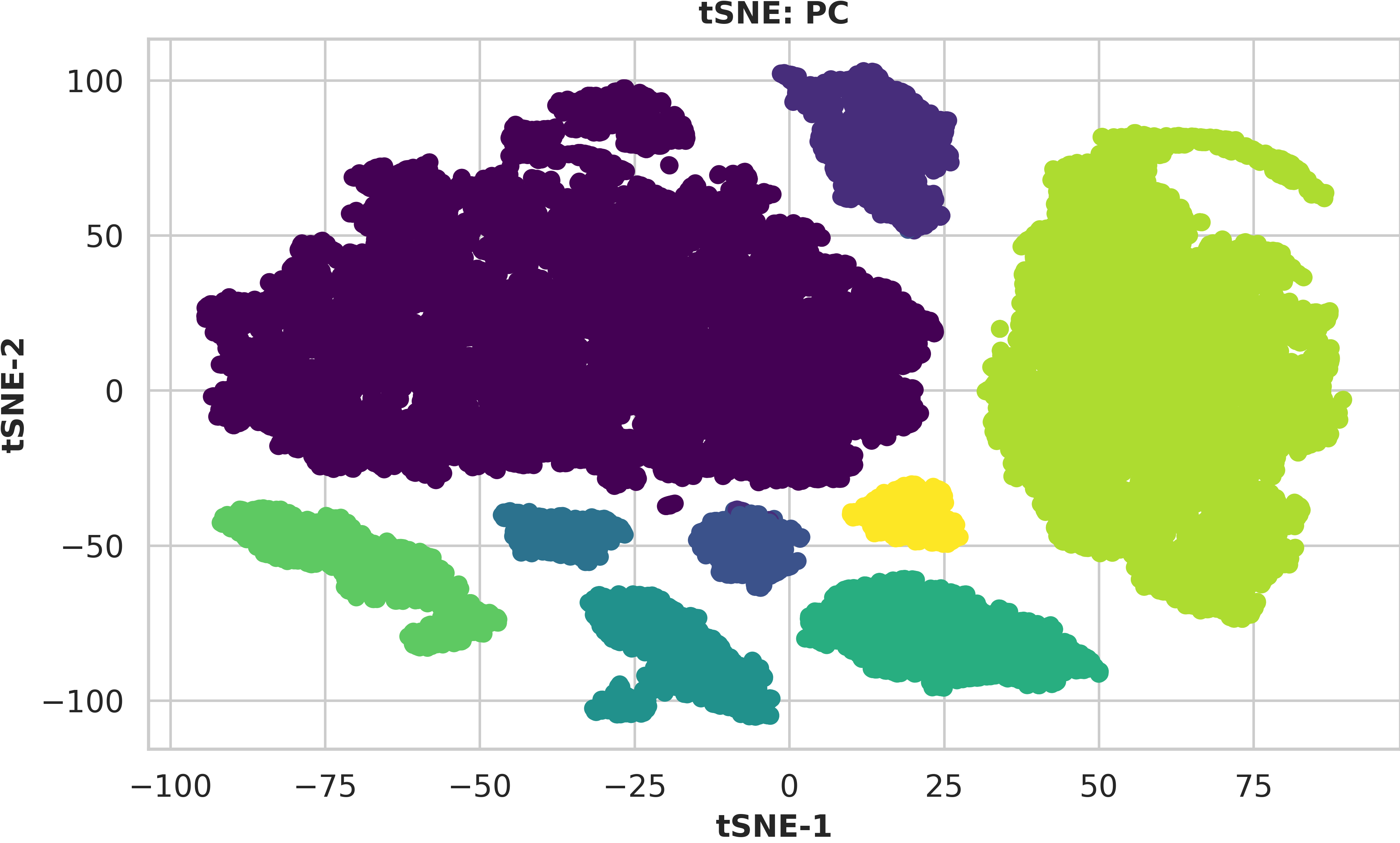}
		\caption{Pavia Center}
		\label{Fig2B}
	\end{subfigure}
	\begin{subfigure}{0.32\textwidth}
		\includegraphics[width=0.99\textwidth]{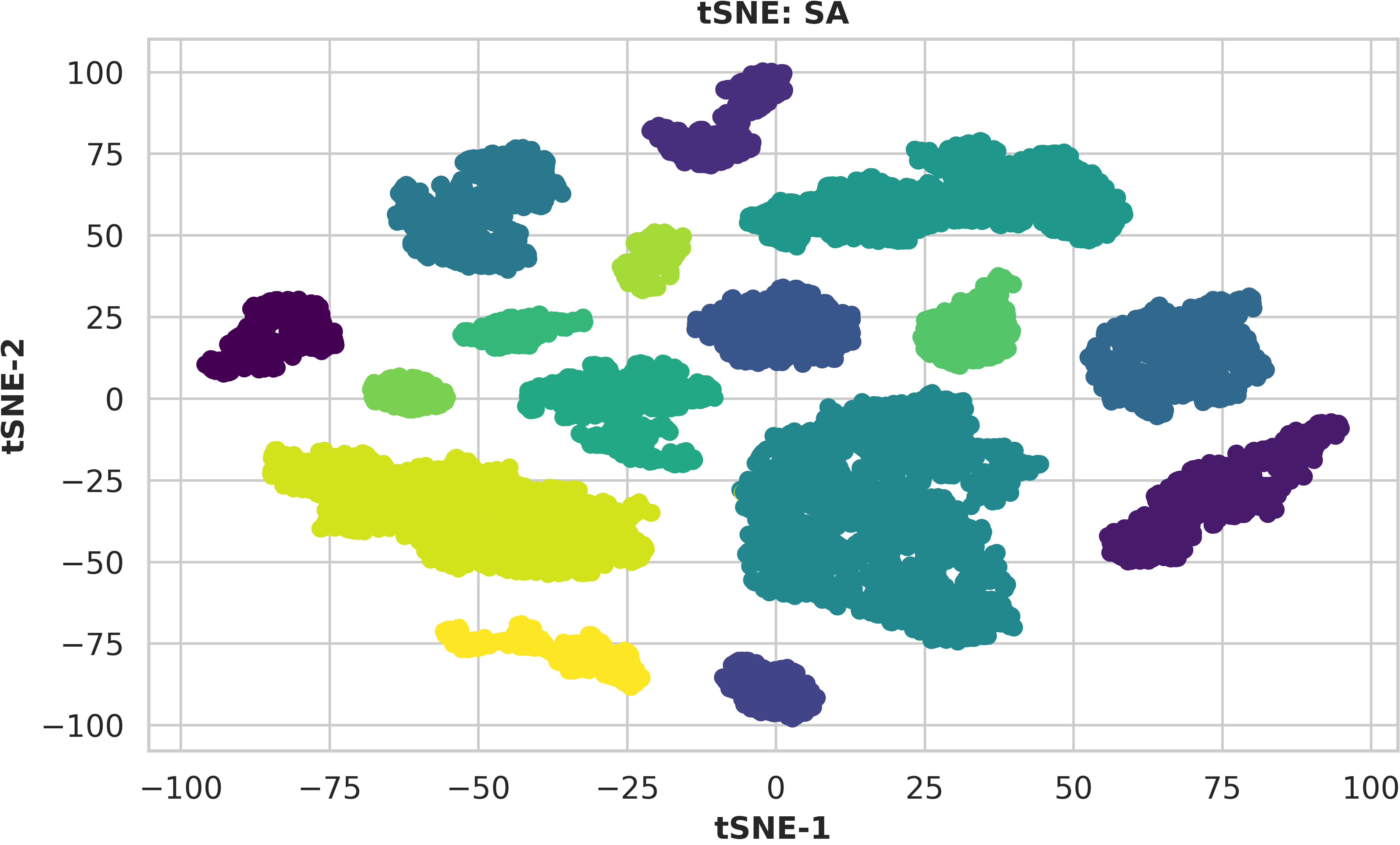}
		\caption{Salinas}
		\label{Fig2C}
	\end{subfigure} 
	\begin{subfigure}{0.32\textwidth}
		\includegraphics[width=0.99\textwidth]{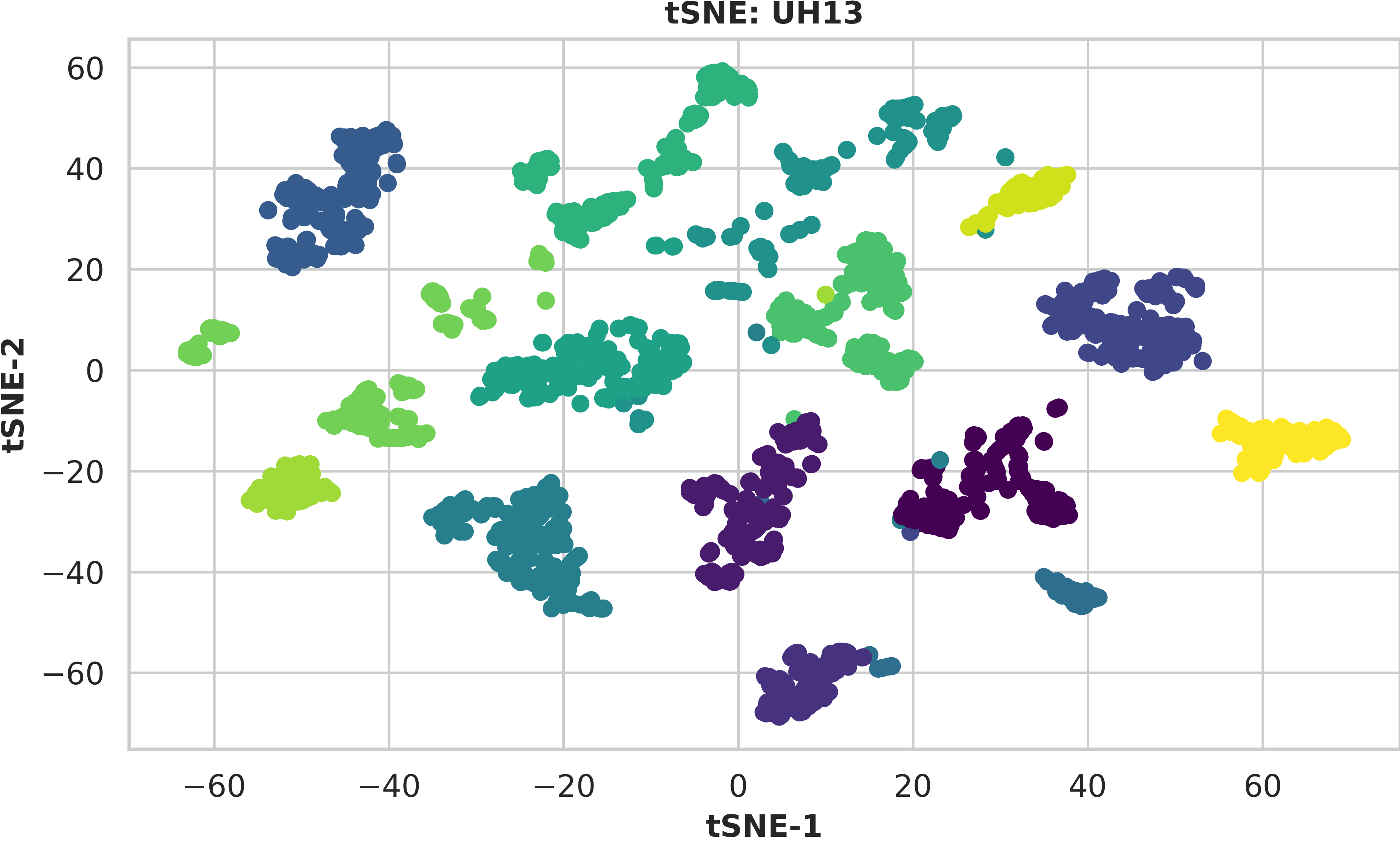}
		\caption{University of Houston}
		\label{Fig2D}
	\end{subfigure}
    \begin{subfigure}{0.32\textwidth}
		\includegraphics[width=0.99\textwidth]{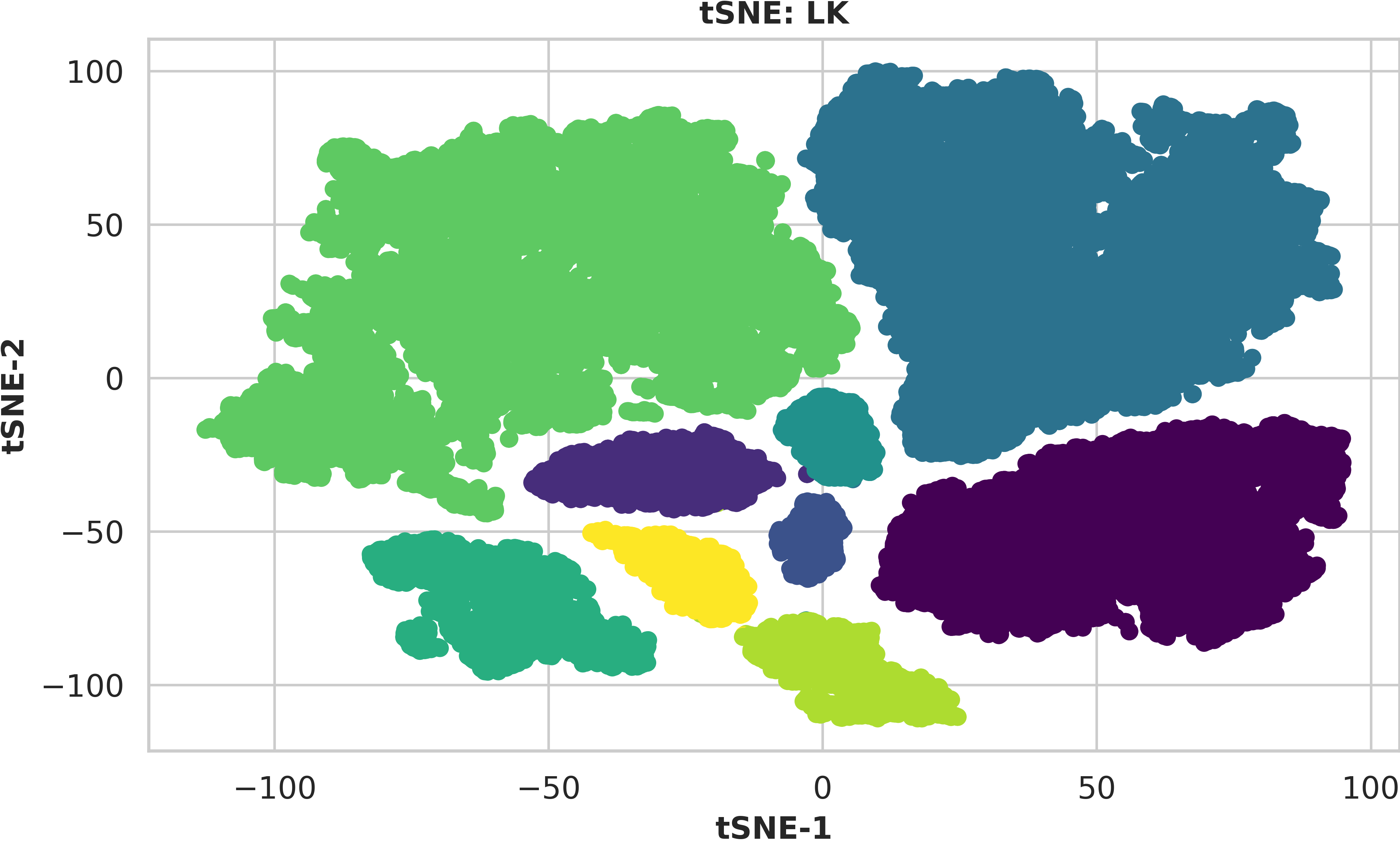}
		\caption{WHU-Hi-LongKou}
		\label{Fig2E}
	\end{subfigure}
\caption{Learned features representation for Pavia University, Pavia Center, Salinas, University of Houston, and WHU-Hi-LongKou Datasets. }
\label{Fig2}
\end{figure*}

For the LK dataset, SMM achieved an OA of 99.25\%, slightly lower than SMM's 99.70\% but still better than NM’s 99.51\%. The $\kappa$ coefficient for SMM was 99.02\%, slightly below SSMM (99.61\%) but superior to NM (99.36\%). In the SA dataset, SMM demonstrated a solid improvement over NM, achieving an OA of 96.78\% and a $\kappa$ of 96.41\%, both significantly higher than NM (95.20\% OA and 94.65\% $\kappa$), though still outperformed by SSMM (98.52\% OA and 98.35\% $\kappa$). SMM's performance in the UH dataset was also notable, achieving an OA of 96.78\% and a $\kappa$ of 96.41\%, which were significantly better than NM's OA of 90.36\% and $\kappa$ of 89.58\%. SSMM, however, further improved the results to an OA of 98.28\% and a $\kappa$ of 98.14\%. In short, while SSMM consistently achieved the highest accuracy and classification accuracy, SMM offers significant performance improvements over the No Morphology model, particularly in datasets where spatial information plays a vital role in classification accuracy. The incorporation of spatial morphological operations alone yields measurable benefits in OA, although combining both spatial and spectral morphology leads to the best results.

As stated above, this study also experimented with various hyperparameter settings, such as training sample ratios (1\%, 2\%, 5\%, 10\%, 15\%, 20\%, and 25\%), different patch sizes $2 \times 2$, $4 \times 4$, $6 \times 6$, $8 \times 8$, and $10 \times 10$), number of attention head (2, 4, 6, and 8), and kernel sizes ($3 \times 3$, $5 \times 5$, $7 \times 7$, $9 \times 9$, and $11 \times 11$) for morphological operations. As shown in Figure \ref{overall_acc_sub_plots}, the OA improves as the ratio of training samples increases. For instance, on the LK dataset, the OA increased from 97.20\% (at 5\% training samples) to 99.73\% (at 25\% training samples). Similarly, on the UH dataset, OA increased from 79.16\% to 97.64\%, reflecting the importance of larger training sets for achieving higher accuracy. These results highlight SSMM’s adaptability across different datasets and training scenarios.

\begin{figure}[!hbt]
    \centering
    \includegraphics[width=0.48\textwidth]{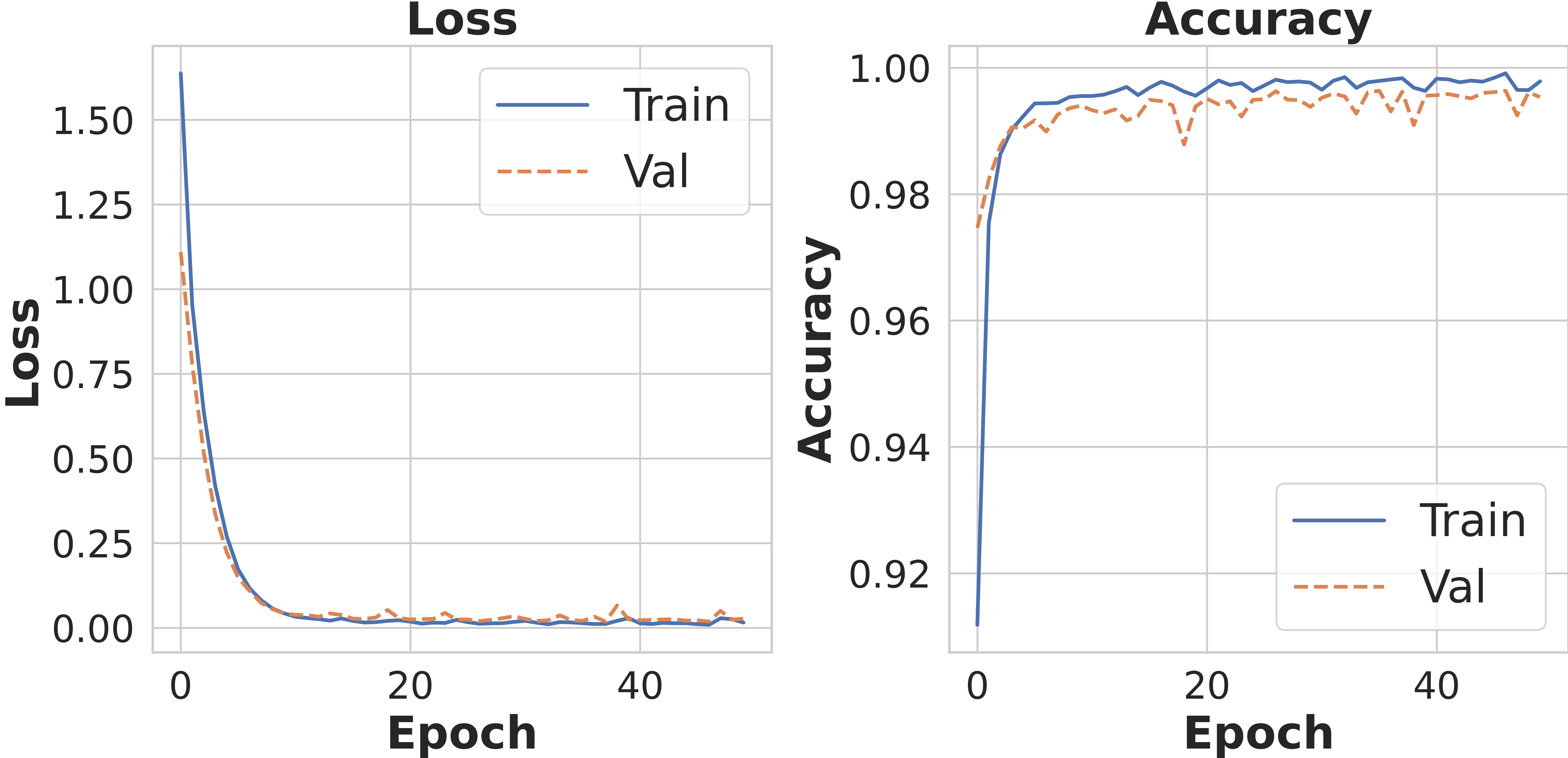}
    \caption{The accuracy and loss for both the training and validation sets were computed using a $4 \times 4$ patch and 20\% training samples over 50 epochs on the LK dataset.}
    \label{fig:lk_loss_acc}
\end{figure}

Additionally, Figure \ref{overall_time_sub_plots} illustrates the computational time required for different settings. As expected, larger kernel sizes for morphological operations and more attention heads led to increased computational time. However, the trade-off between accuracy and computational efficiency was manageable, and MorpMamba remained efficient even with complex configurations. The best settings from these experiments were used to evaluate the MorpMamba in comparison with other methods. All experiments were performed on an Intel i9-13900k machine with an RTX 4090 GPU and 32GB of RAM using Jupyter Notebook. Figure \ref{fig:lk_loss_acc} presents the convergence of loss and accuracy of the MorpMamba model over 50 epochs.

To further analyze the feature representations learned by MorpMamba, we used t-distributed Stochastic Neighbor Embedding (t-SNE) to visualize the high-dimensional data in a 2D space. t-SNE preserves local structures within the data, making it an effective tool for understanding how well the model captures spectral-spatial features. As shown in Figures \ref{Fig2A}, \ref{Fig2B}, \ref{Fig2C}, \ref{Fig2D}, and \ref{Fig2E}, t-SNE visualizations for the PU, PC, SA, UH, and LK datasets highlight the effectiveness of MorpMamba in separating feature clusters for different classes. The clear separation of clusters demonstrates MorpMamba’s ability to learn distinct class-specific representations, particularly in challenging datasets such as UH, where classes are difficult to distinguish due to complex spatial relationships.

\begin{table*}[!hbt]
    \centering
    \caption{Performance comparison of MorpMamba with SOTA models on various datasets. The metrics shown are OA, AA, and $\kappa$ coefficient. The datasets include the UH, LK, PU, PC, and SA. The number of parameters for each model is also provided. Moreover, this table presents the results of the spatial-spectral Mamba model (SSMamba), the spatial Morphology Mamba (SMM), and the spatial-spectral Morphology Mamba (SSMM). All of these models were trained using a $4 \times 4$ patch size and 20\% of the training samples.}
    \resizebox{\textwidth}{!}{\begin{tabular}{c|ccc|ccc|ccc|ccc|ccc||c} \hline 
        \multirow{2}{*}{\textbf{Model}} & \textbf{OA} & \textbf{AA} & \textbf{$\kappa$} & \textbf{OA} & \textbf{AA} & \textbf{$\kappa$} & \textbf{OA} & \textbf{AA} & \textbf{$\kappa$} & \textbf{OA} & \textbf{AA} & \textbf{$\kappa$} & \textbf{OA} & \textbf{AA} & \textbf{$\kappa$} & \multirow{2}{*}{\textbf{Parameters$\approx$}} \\ \cline{2-16}
        & \multicolumn{3}{c|}{\textbf{University of Houston}} & \multicolumn{3}{c|}{\textbf{WHU-Hi-LongKou (LK)}} & \multicolumn{3}{c|}{\textbf{Pavia University}} & \multicolumn{3}{c|}{\textbf{Pavia Center}} & \multicolumn{3}{c||}{\textbf{Salinas}} & \\ \hline 
        2DCNN & 97.49 & 97.04 & 97.29 & 99.71 & 99.29 & 99.62 & 97.97 & 96.99 & 97.30 & 99.66 & 99.06 & 99.52 & 97.54 & 98.82 & 97.26 & 322752 \\ \hline 
        3DCNN & 99.01 & 98.81 & 98.93 & 99.81 & 99.58 & 99.75 & 98.70 & 97.86 & 98.28 & 99.87 & 99.65 & 99.81 & 98.86 & 99.48 & 98.73 & 4042880 \\ \hline
        HybCNN & 98.93 & 98.71 & 98.84 & 99.84 & 99.60 & 99.79 & 43.59 & --- & --- & 99.78 & 99.45 & 99.69 & 98.76 & 99.42 & 98.62 & 594048  \\ \hline
        2DIN & 99.09 & 98.96 & 99.02 & 99.83 & 99.56 & 99.78 & 98.74 & 98.10 & 98.33 & 99.82 & 99.52 & 99.74 & 98.65 & 99.28 & 98.50 & 3285844 \\ \hline
        3DIN & 98.73 & 98.43 & 98.63 & 99.80 & 99.49 & 99.74 & 98.51 & 97.61 & 98.03 & 99.86  & 99.61 & 99.81 & 98.23 & 99.16 & 98.03 & 47448680 \\ \hline
        HybIN & 98.81 & 98.56 & 98.71 & 99.75 & 99.56 & 99.67 & 98.79 & 98.26 & 98.40 & 99.82 & 99.45 & 99.75 & 98.74 & 99.38 & 98.60 & 1349848 \\ \hline
        MorpCNN & 99.68 & 99.68 & 99.65 & 99.92 & 99.75 & 99.90 & 99.84 & 99.66 & 99.79 & 99.97 & 99.92 & 99.96 & 99.89 & 99.87 & 99.88 & 789071 \\ \hline 
        Hybrid-ViT & 98.45 & 97.85 & 98.33 & 99.75 & 99.36 & 99.68 & 98.15 & 97.24 & 97.55 & 99.71 & 99.13 & 99.59 & 97.99 & 99.05 & 97.76 & 790736 \\ \hline
        Hir-Transformer & 97.12 & 96.25 & 96.89 & 99.68 & 99.14 & 99.59 & 97.99 & 96.79 & 97.34 & 99.64 & 98.79 & 99.49 & 98.09 & 98.95 & 97.87 & 4219094 \\ \hline \hline 
        \textbf{SSMamba} & 90.36 & 90.54 & 89.58 & 99.51 & 98.45 & 99.36 & 95.67 & 93.20 & 94.25 & 99.48 & 98.60 & 99.27 & 95.20 & 97.81 & 94.65 & 49744 \\ \hline
        \textbf{SMM} & 96.46 & 95.93 & 96.17 & 99.25 & 97.53 & 99.02 & 96.52 & 95.94 & 95.40 & 99.52 & 98.28 & 99.32 & 96.78 & 98.48 & 96.41 & 62665 \\ \hline 
        \textbf{SSMM} & \textbf{98.28} & \textbf{97.91} & \textbf{98.14} & \textbf{99.70} & \textbf{99.25} & \textbf{99.61} & \textbf{97.67} & \textbf{96.93} & \textbf{96.91} & \textbf{99.71} & \textbf{98.85} & \textbf{99.59} & \textbf{98.52} & \textbf{99.25} & \textbf{98.35} & \textbf{67142}   \\ \hline        
    \end{tabular}}
    \label{my_label}
\end{table*}
\begin{figure*}[!htb]
    \centering
        \begin{subfigure}{0.15\textwidth}
            \includegraphics[width=0.99\textwidth]{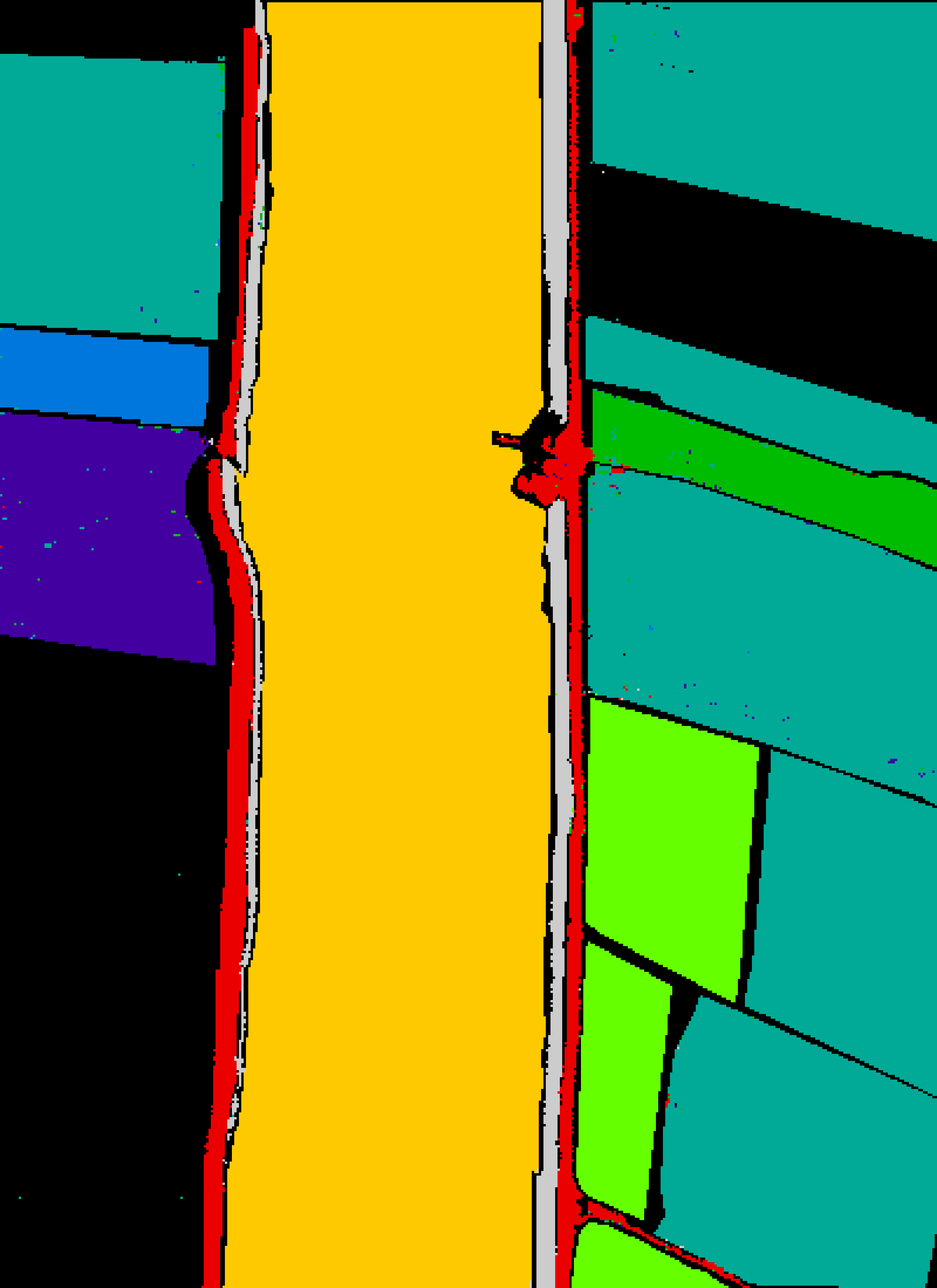}
            \caption{CNN2D}
        \end{subfigure}
        \begin{subfigure}{0.15\textwidth}
            \centering
            \includegraphics[width=0.99\textwidth]{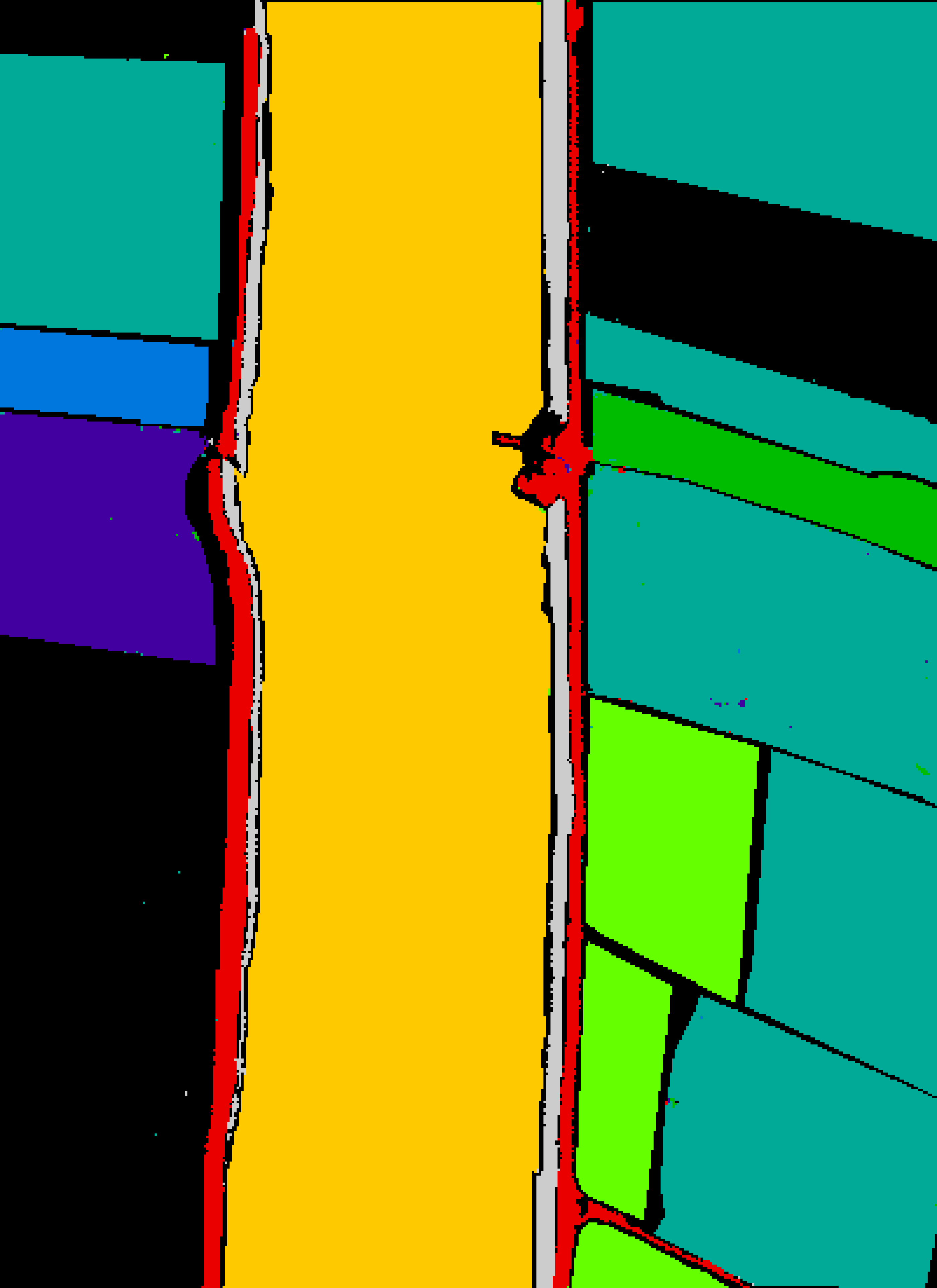}
            \caption{CNN3D}
        \end{subfigure}
        \begin{subfigure}{0.15\textwidth}
            \centering
            \includegraphics[width=0.99\textwidth]{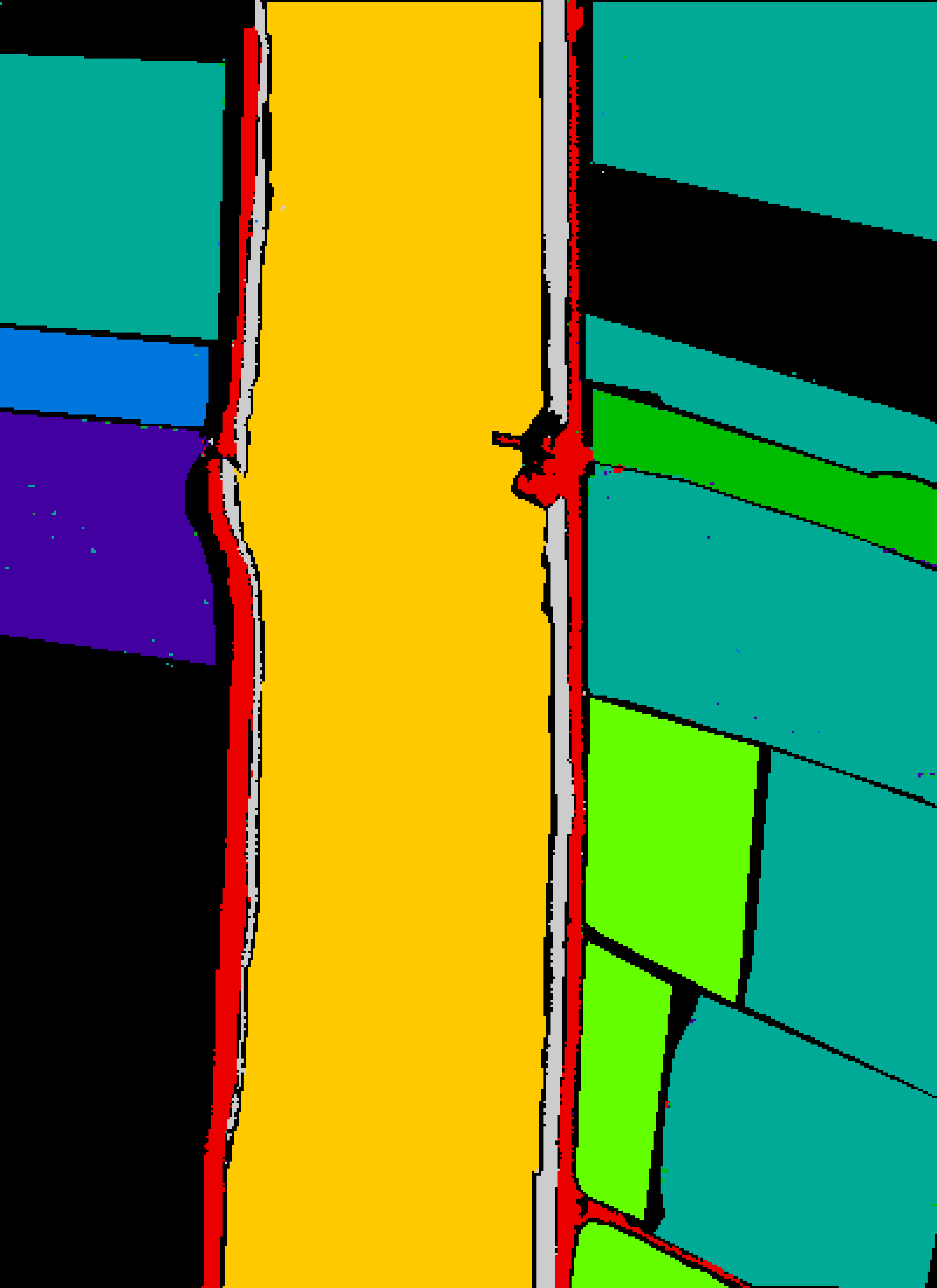}
            \caption{HybCNN}
        \end{subfigure}
        \begin{subfigure}{0.15\textwidth}
            \centering
            \includegraphics[width=0.99\textwidth]{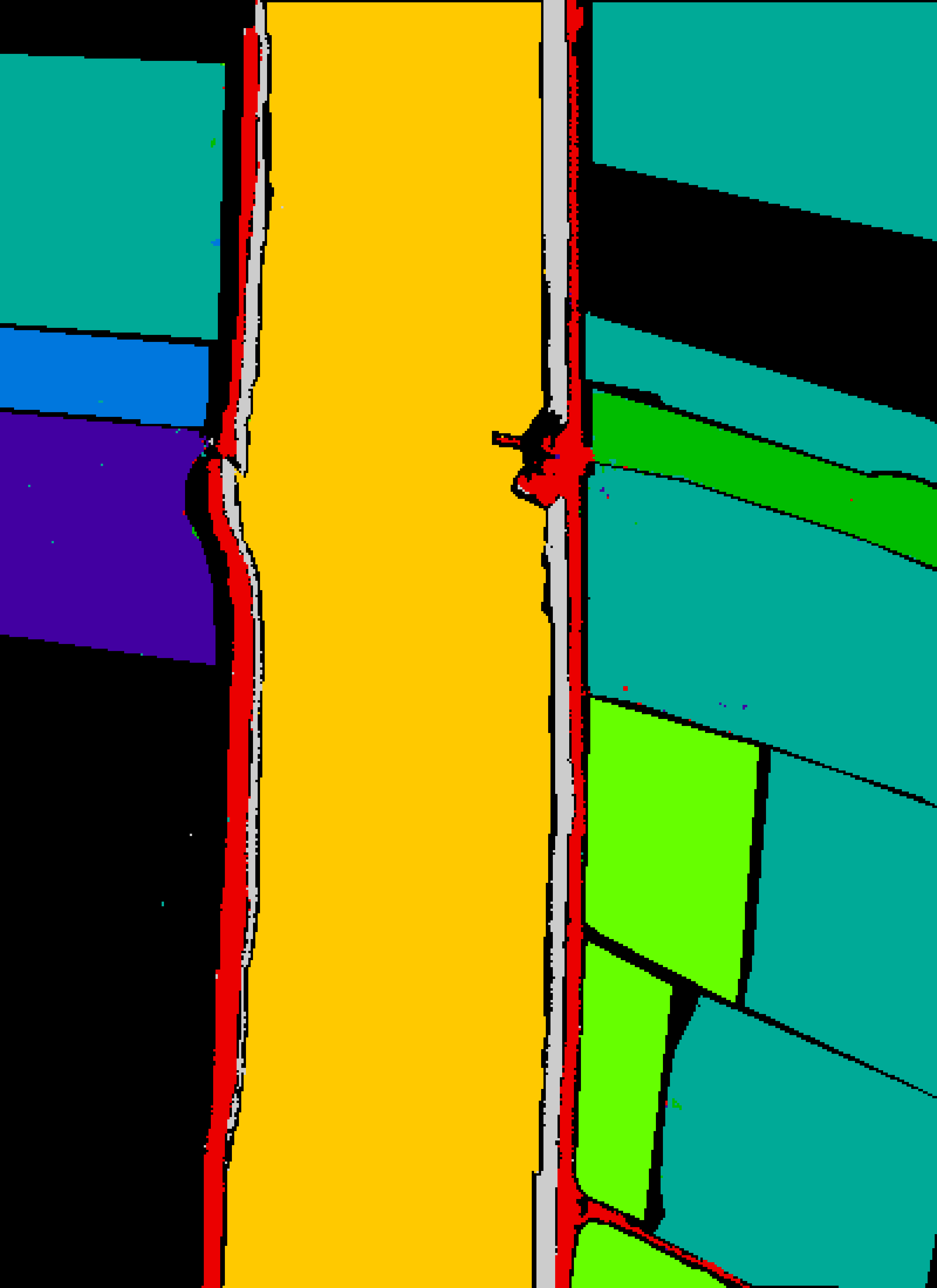}
            \caption{IN2D}
        \end{subfigure}
        \begin{subfigure}{0.15\textwidth}
            \centering
            \includegraphics[width=0.99\textwidth]{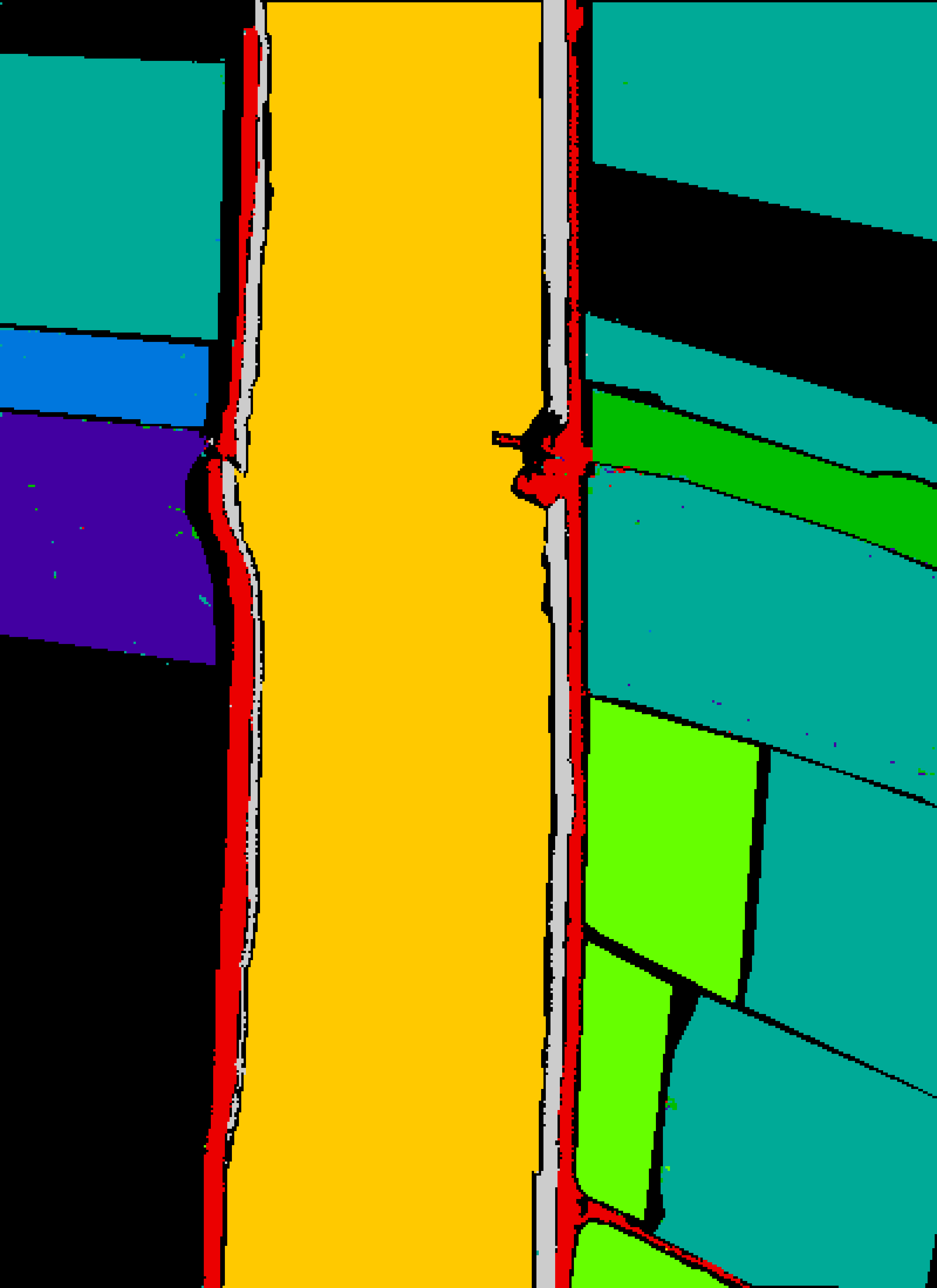}
            \caption{IN3D}
        \end{subfigure}
        \begin{subfigure}{0.15\textwidth}
            \centering
            \includegraphics[width=0.99\textwidth]{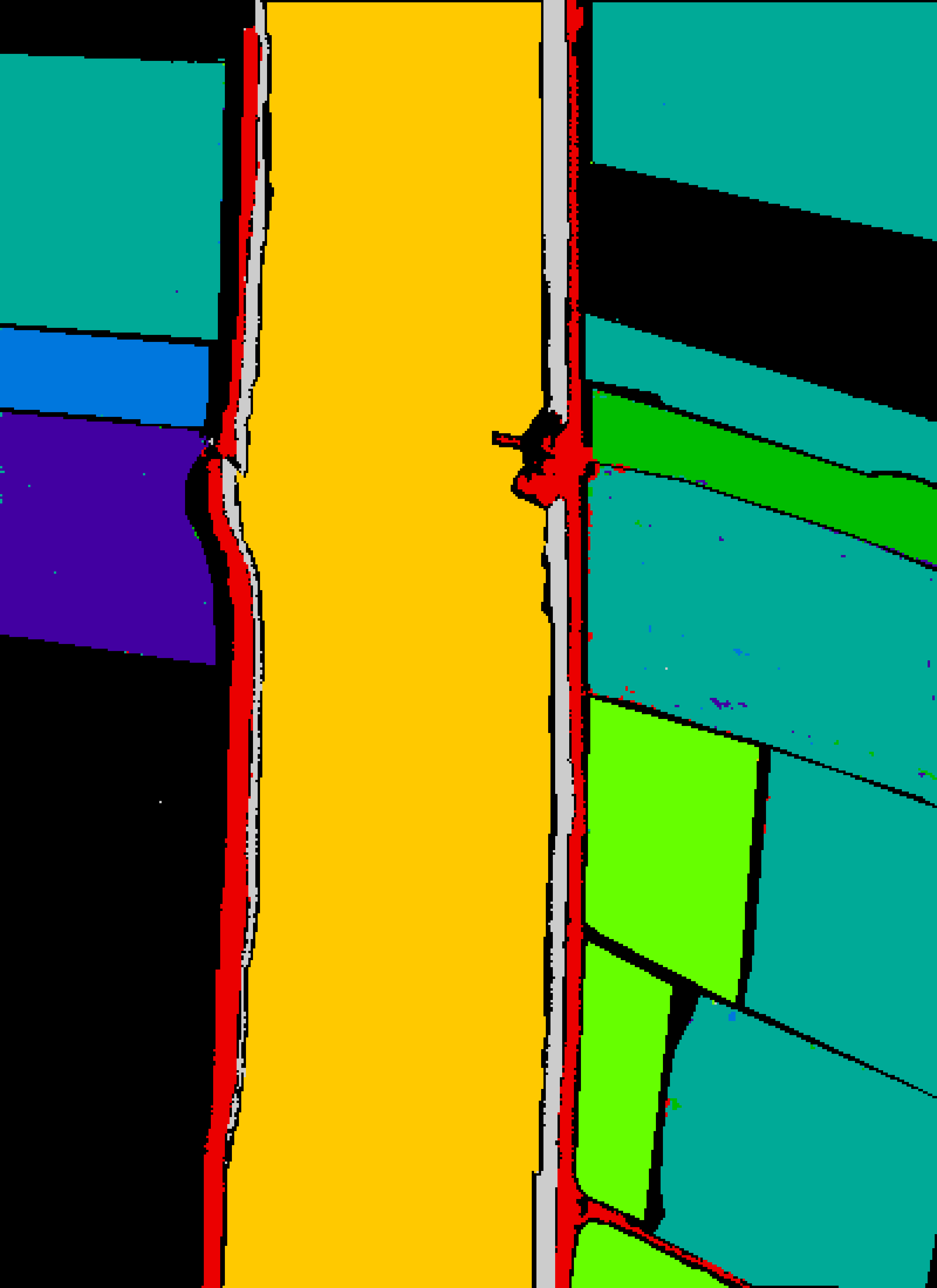}
            \caption{HybIN}
        \end{subfigure}
        \begin{subfigure}{0.15\textwidth}
            \centering
            \includegraphics[width=0.99\textwidth]{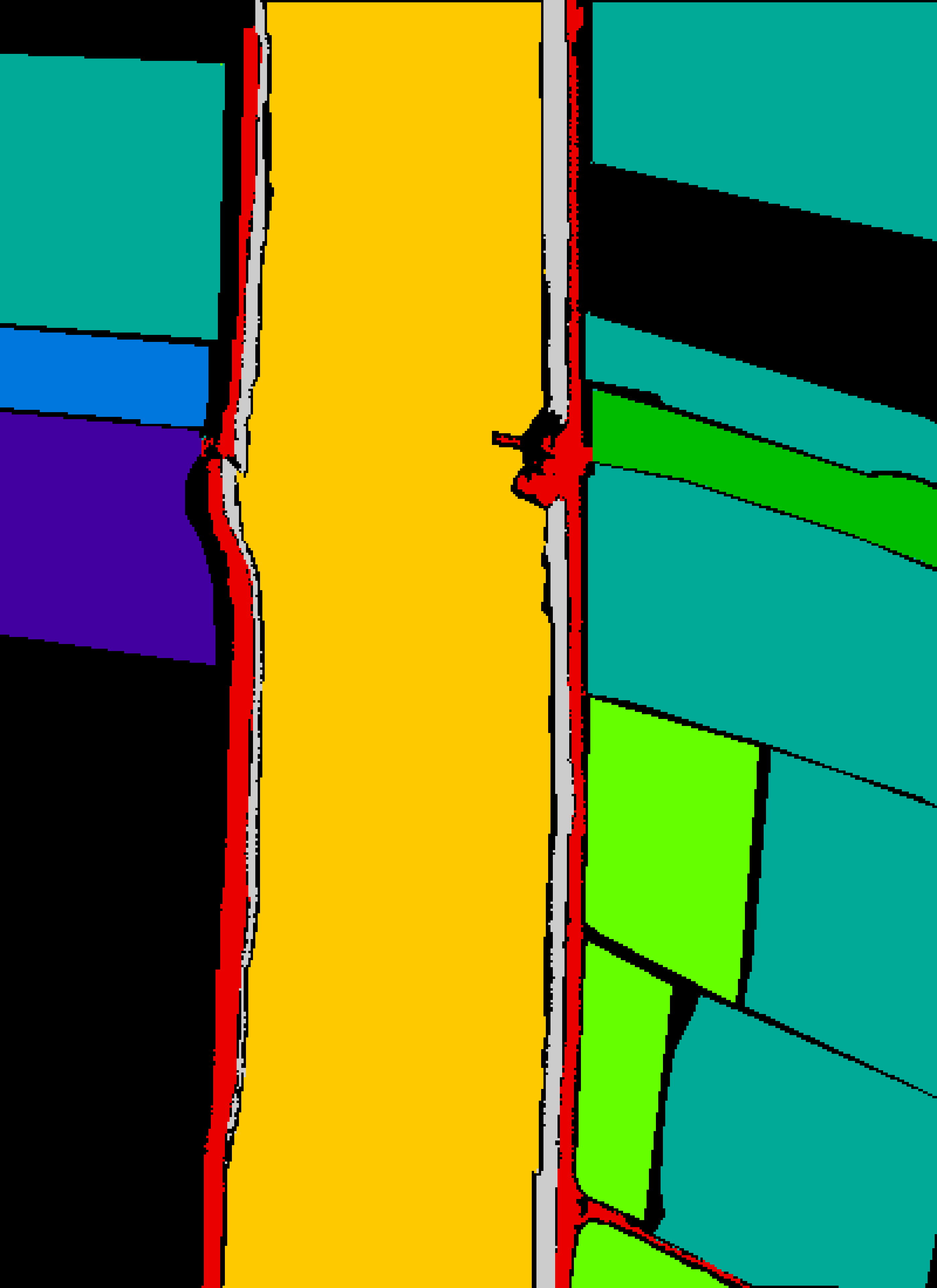}
            \caption{MorpCNN}
        \end{subfigure}
        \begin{subfigure}{0.15\textwidth}
            \centering
            \includegraphics[width=0.99\textwidth]{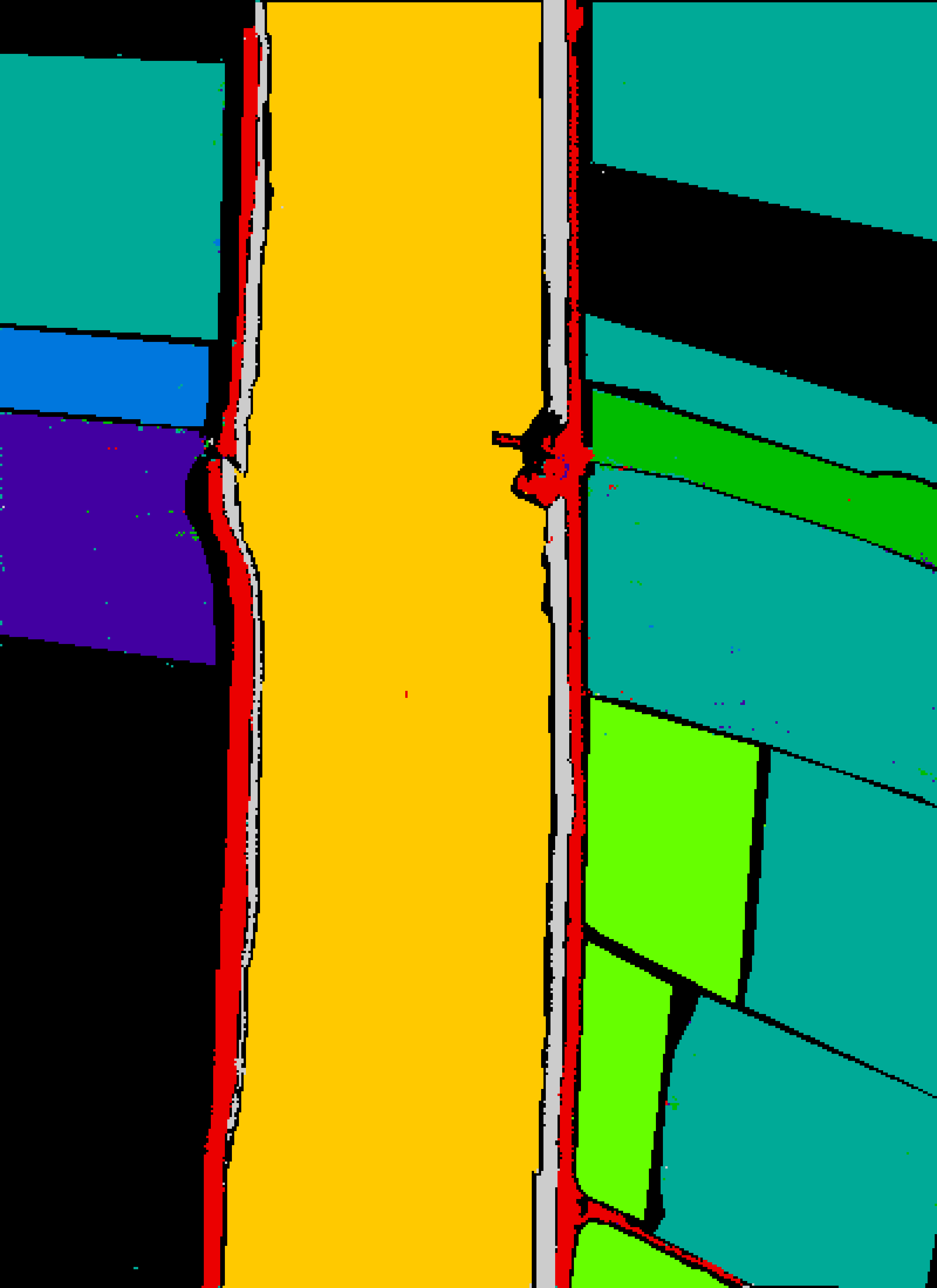}
            \caption{Hybrid-ViT}
        \end{subfigure}
        \begin{subfigure}{0.15\textwidth}
            \centering
            \includegraphics[width=0.99\textwidth]{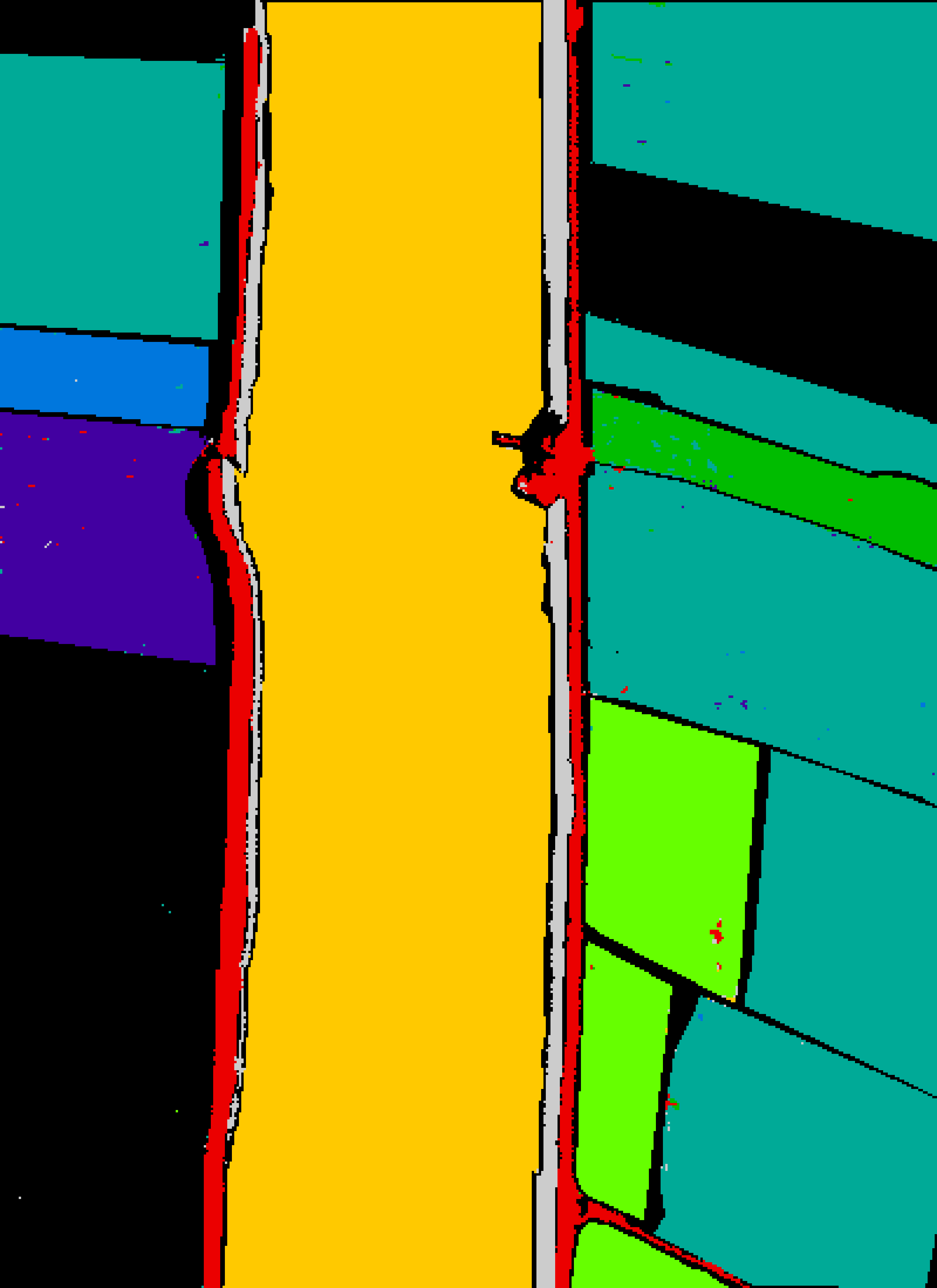}
            \caption{Hir-Transformer}
        \end{subfigure}
        \begin{subfigure}{0.15\textwidth}
            \centering
            \includegraphics[width=0.99\textwidth]{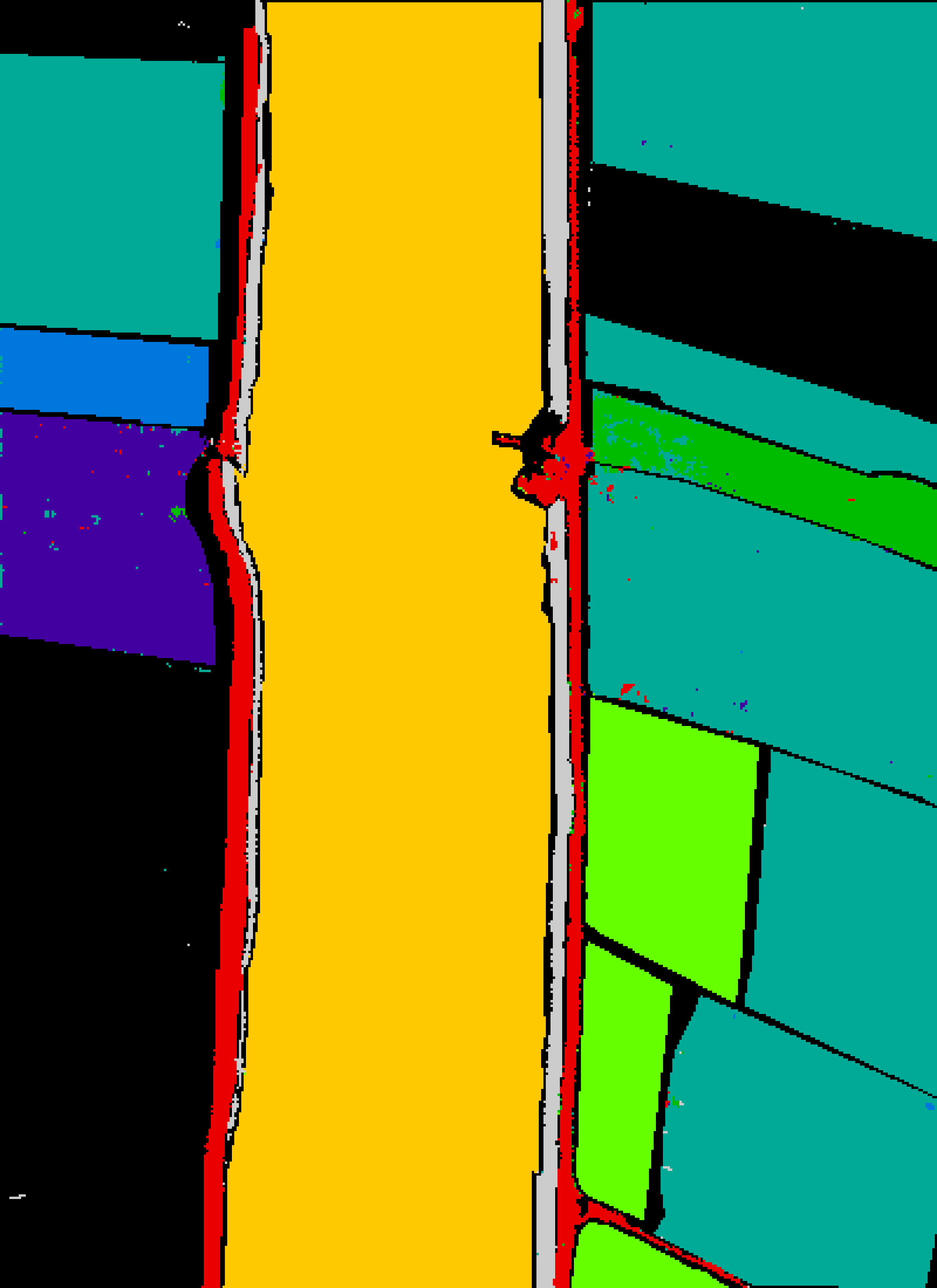}
            \caption{SSMamba}
        \end{subfigure}
        \begin{subfigure}{0.15\textwidth}
            \centering
            \includegraphics[width=0.99\textwidth]{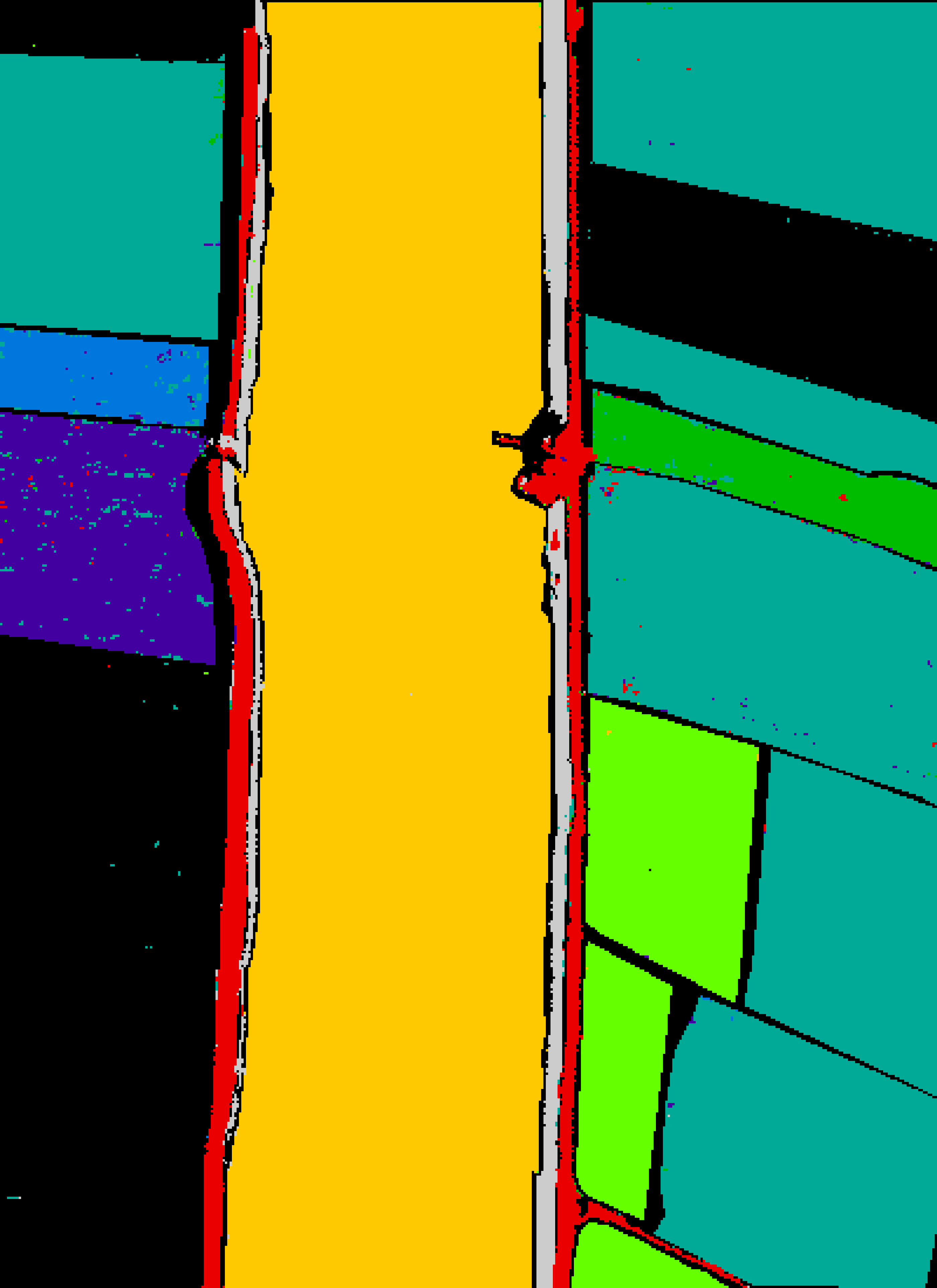}
            \caption{SMM}
        \end{subfigure}
        \begin{subfigure}{0.15\textwidth}
            \centering
            \includegraphics[width=0.99\textwidth]{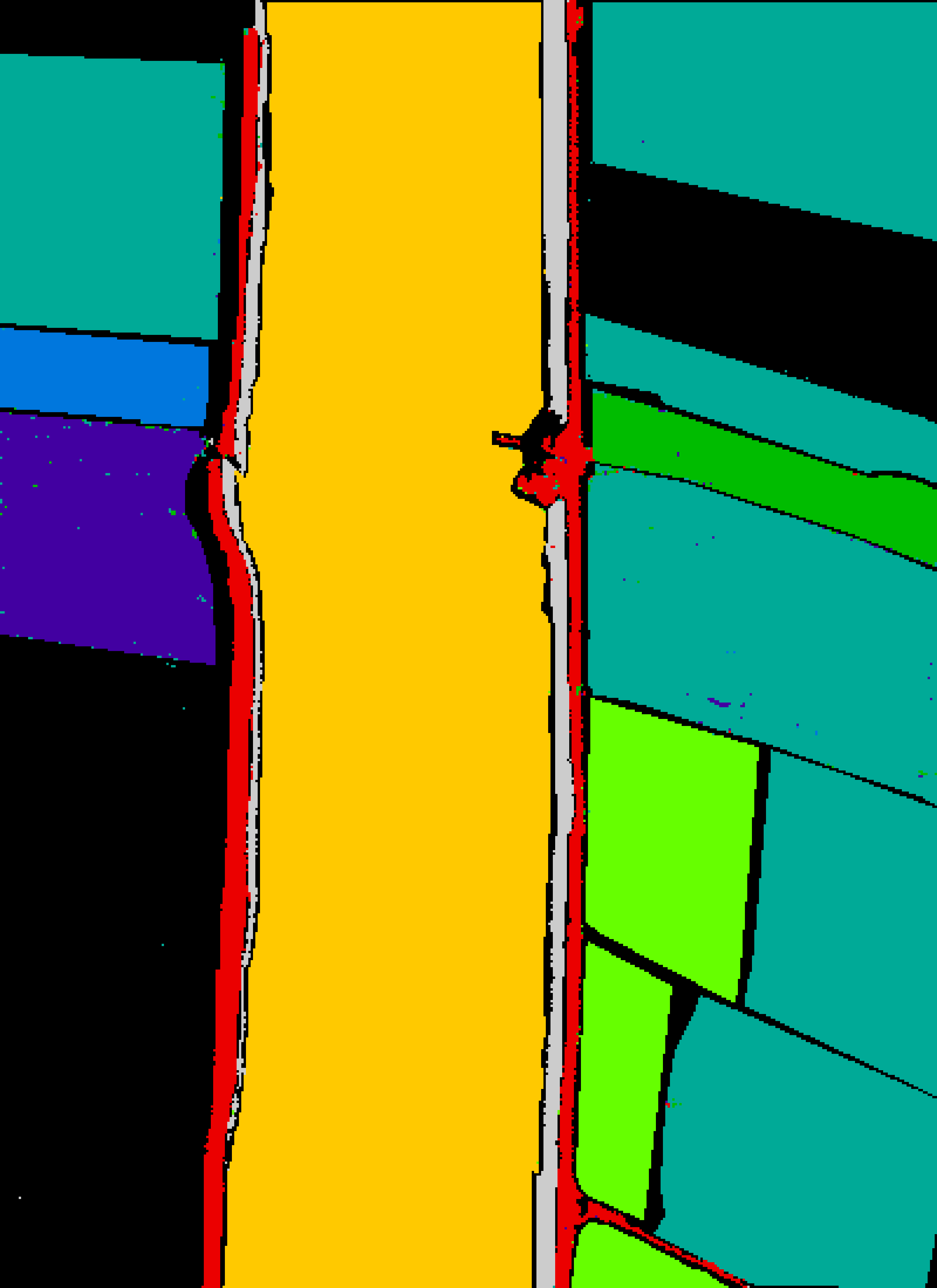}
            \caption{SSMM}
        \end{subfigure}
    \caption{\textbf{LK Dataset:} The predicted ground truth maps for various competing methods alongside the proposed variants of the MorpMamba model. While many competing methods achieved similar accuracy levels, they demonstrated limited parameter efficiency, rendering them less suitable for deployment on resource-constrained devices compared to MorpMamba.}
    \label{LK_results}
\end{figure*}
\begin{figure*}[!htb]
    \centering
        \begin{subfigure}{0.15\textwidth}
            \includegraphics[width=0.99\textwidth]{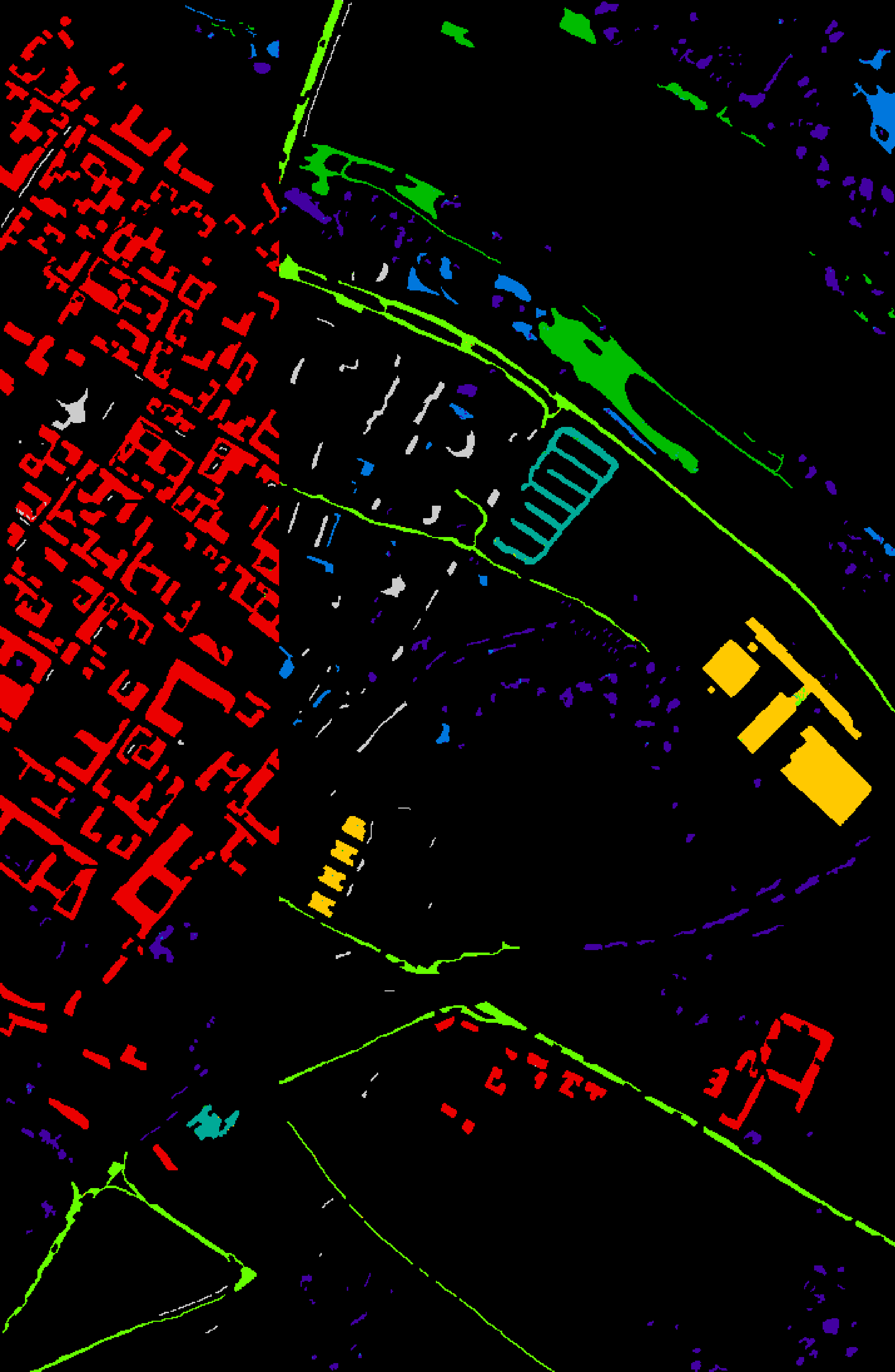}
            \caption{CNN2D}
        \end{subfigure}
        \begin{subfigure}{0.15\textwidth}
            \centering
            \includegraphics[width=0.99\textwidth]{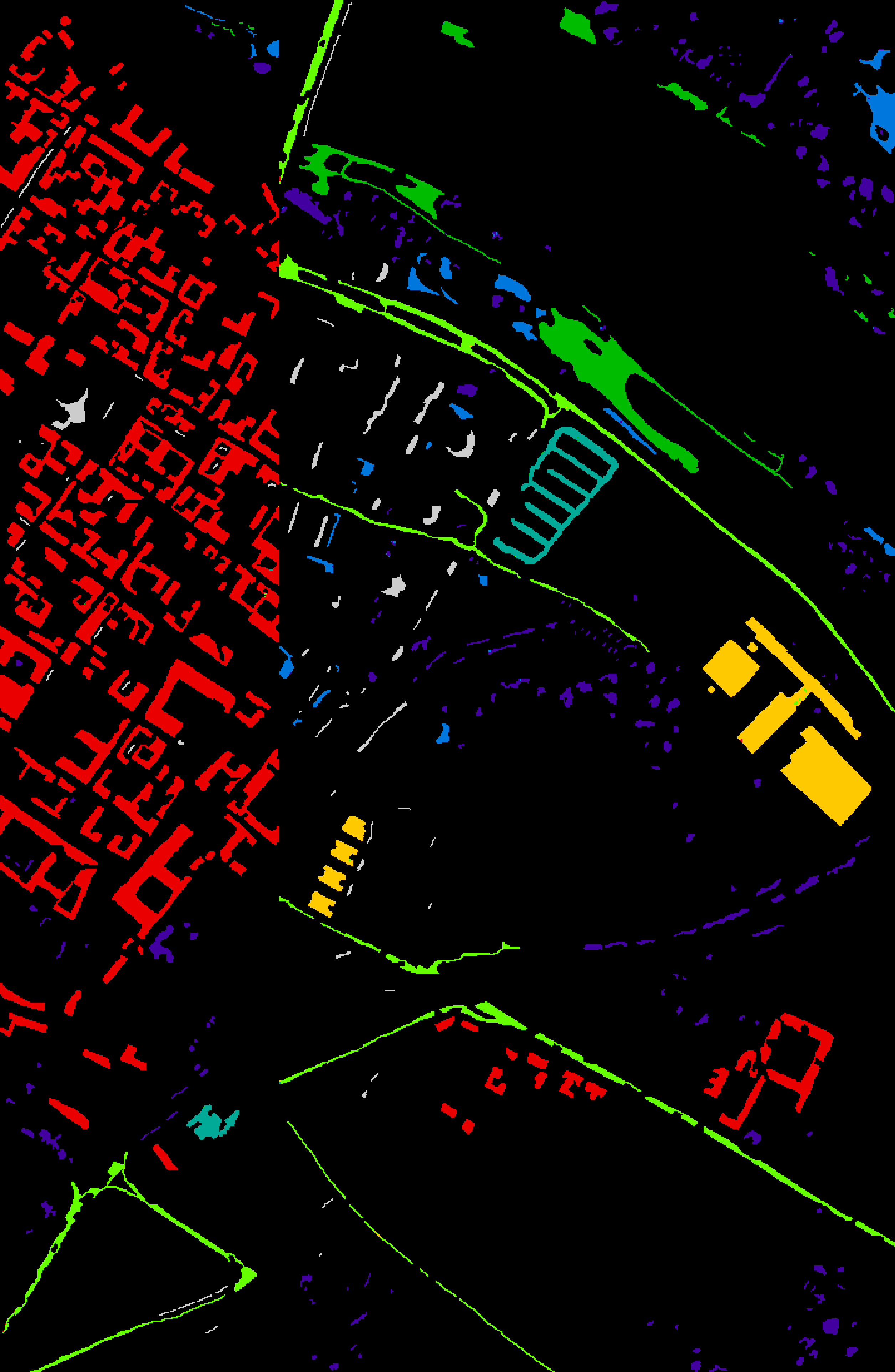}
            \caption{CNN3D}
        \end{subfigure}
        \begin{subfigure}{0.15\textwidth}
            \centering
            \includegraphics[width=0.99\textwidth]{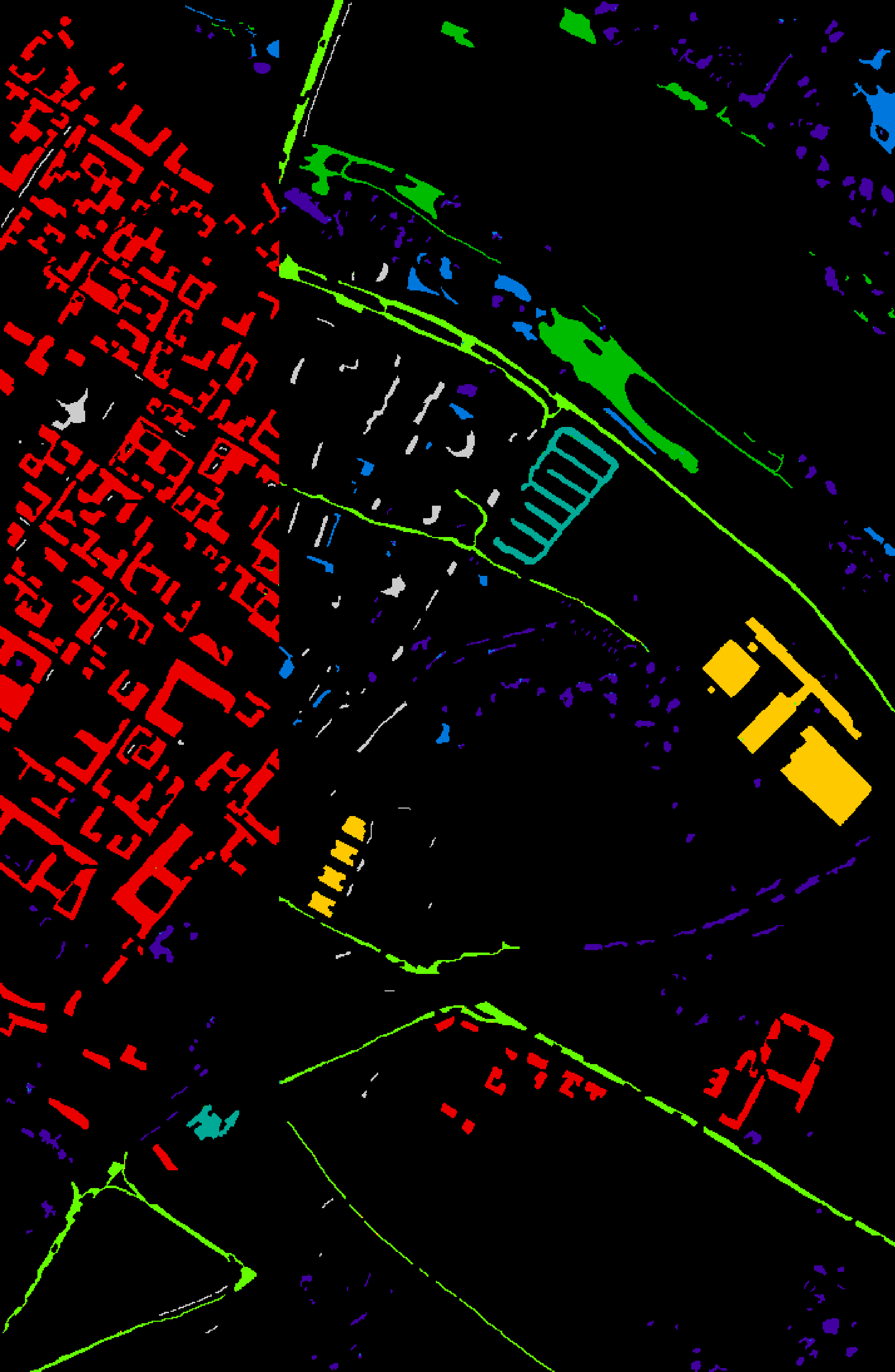}
            \caption{HybCNN}
        \end{subfigure}
        \begin{subfigure}{0.15\textwidth}
            \centering
            \includegraphics[width=0.99\textwidth]{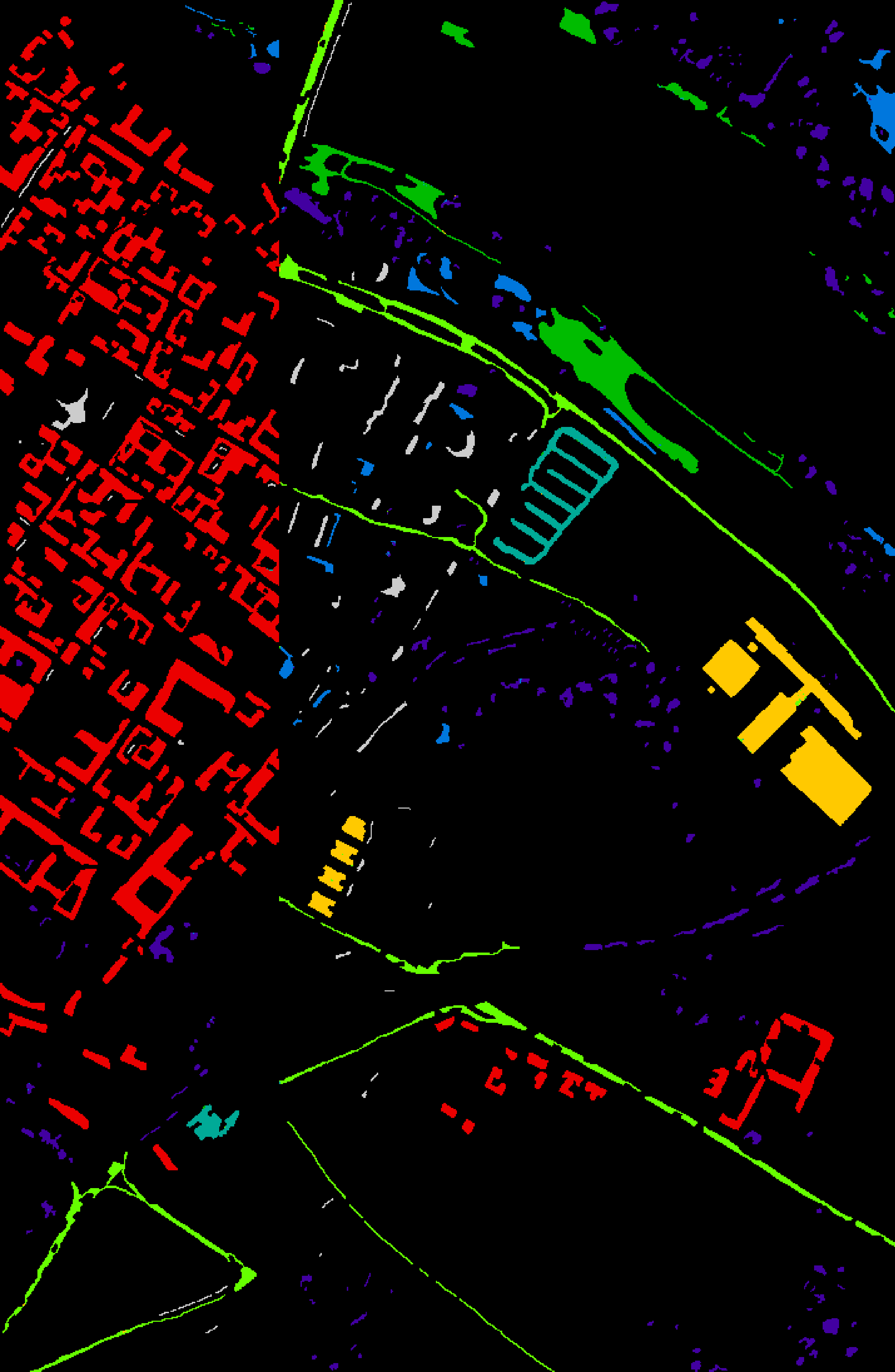}
            \caption{IN2D}
        \end{subfigure}
        \begin{subfigure}{0.15\textwidth}
            \centering
            \includegraphics[width=0.99\textwidth]{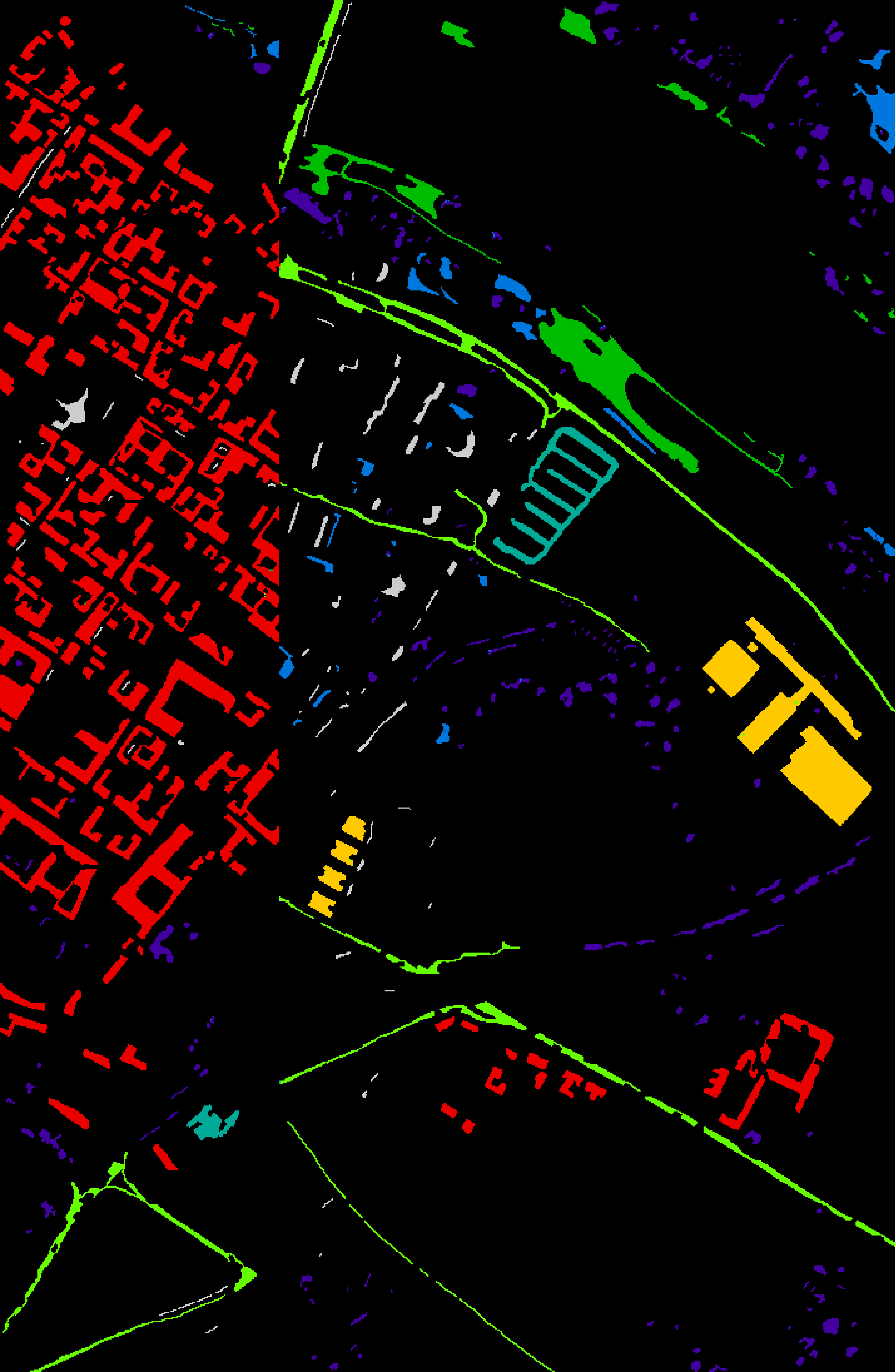}
            \caption{IN3D}
        \end{subfigure}
        \begin{subfigure}{0.15\textwidth}
            \centering
            \includegraphics[width=0.99\textwidth]{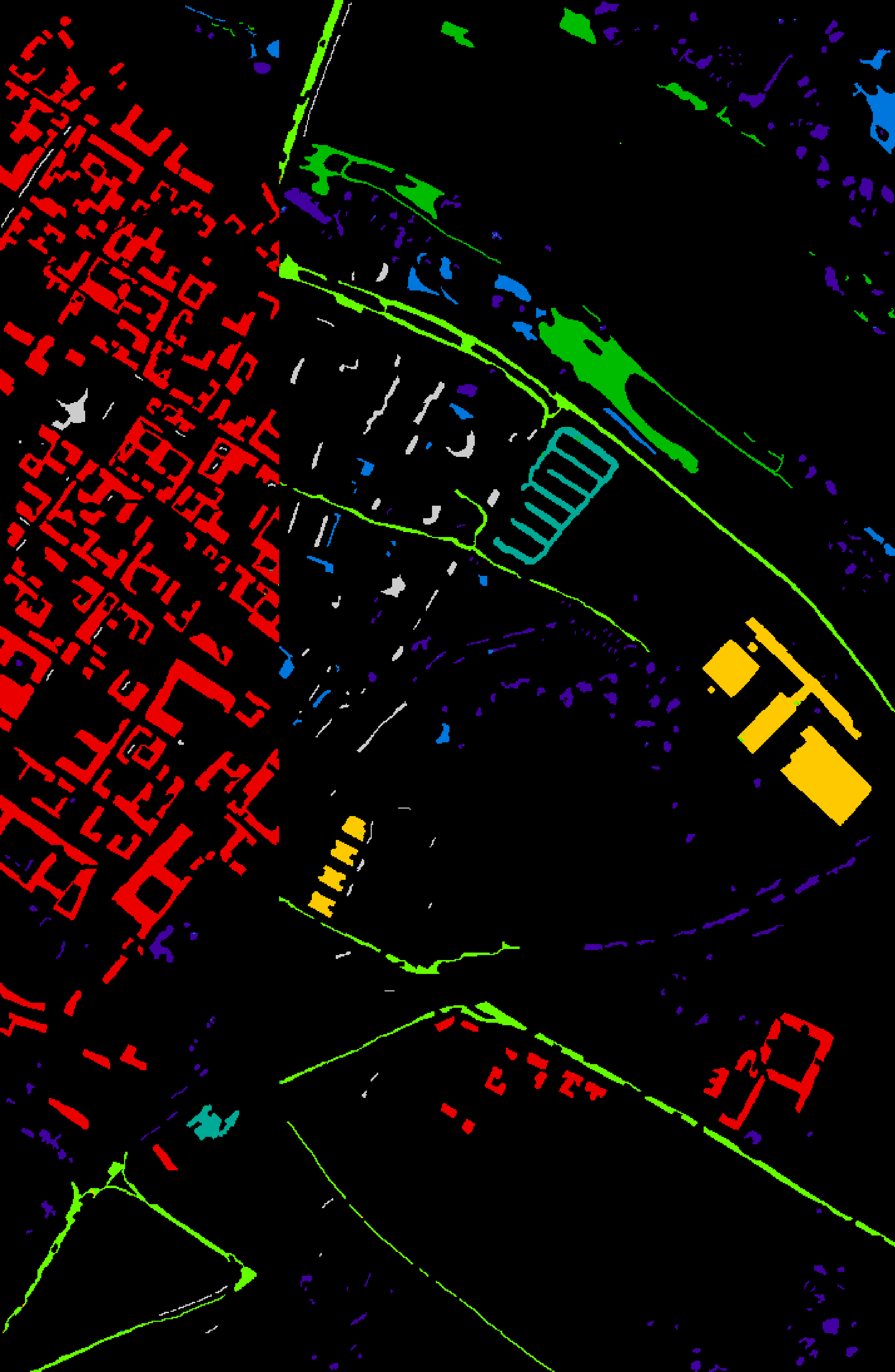}
            \caption{HybIN}
        \end{subfigure}
        \begin{subfigure}{0.15\textwidth}
            \centering
            \includegraphics[width=0.99\textwidth]{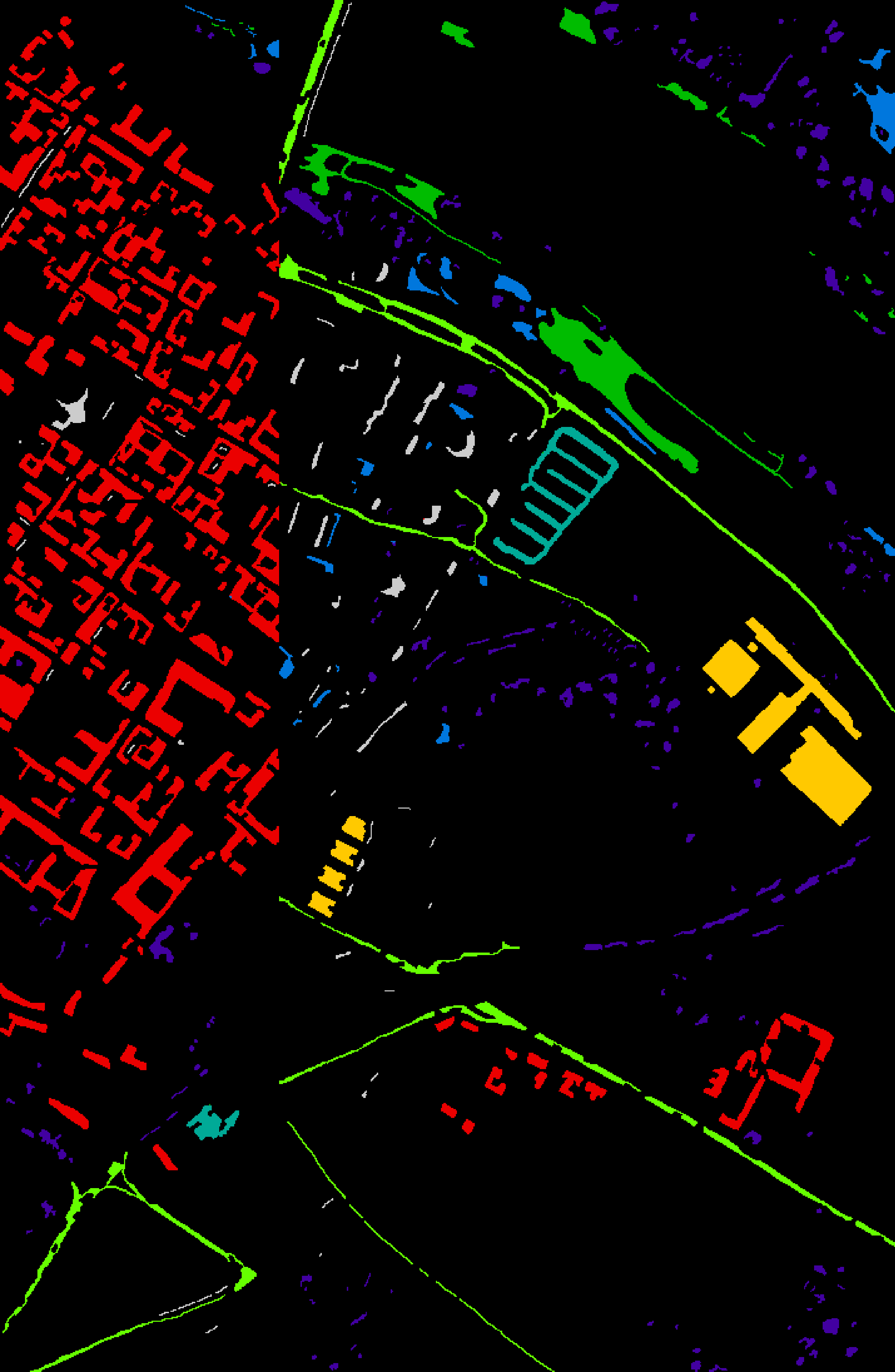}
            \caption{MorpCNN}
        \end{subfigure}
        \begin{subfigure}{0.15\textwidth}
            \centering
            \includegraphics[width=0.99\textwidth]{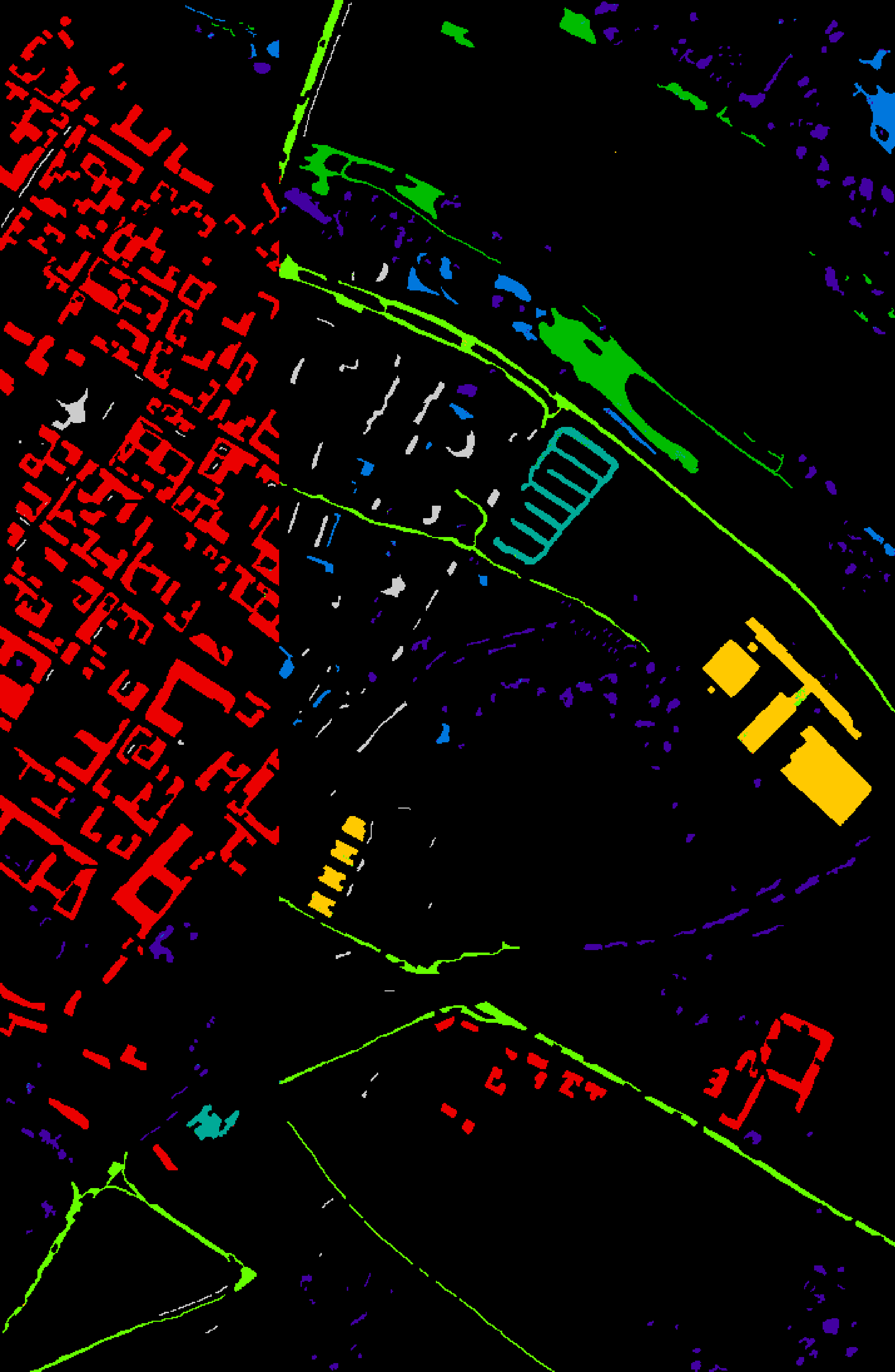}
            \caption{Hybrid-ViT}
        \end{subfigure}
        \begin{subfigure}{0.15\textwidth}
            \centering
            \includegraphics[width=0.99\textwidth]{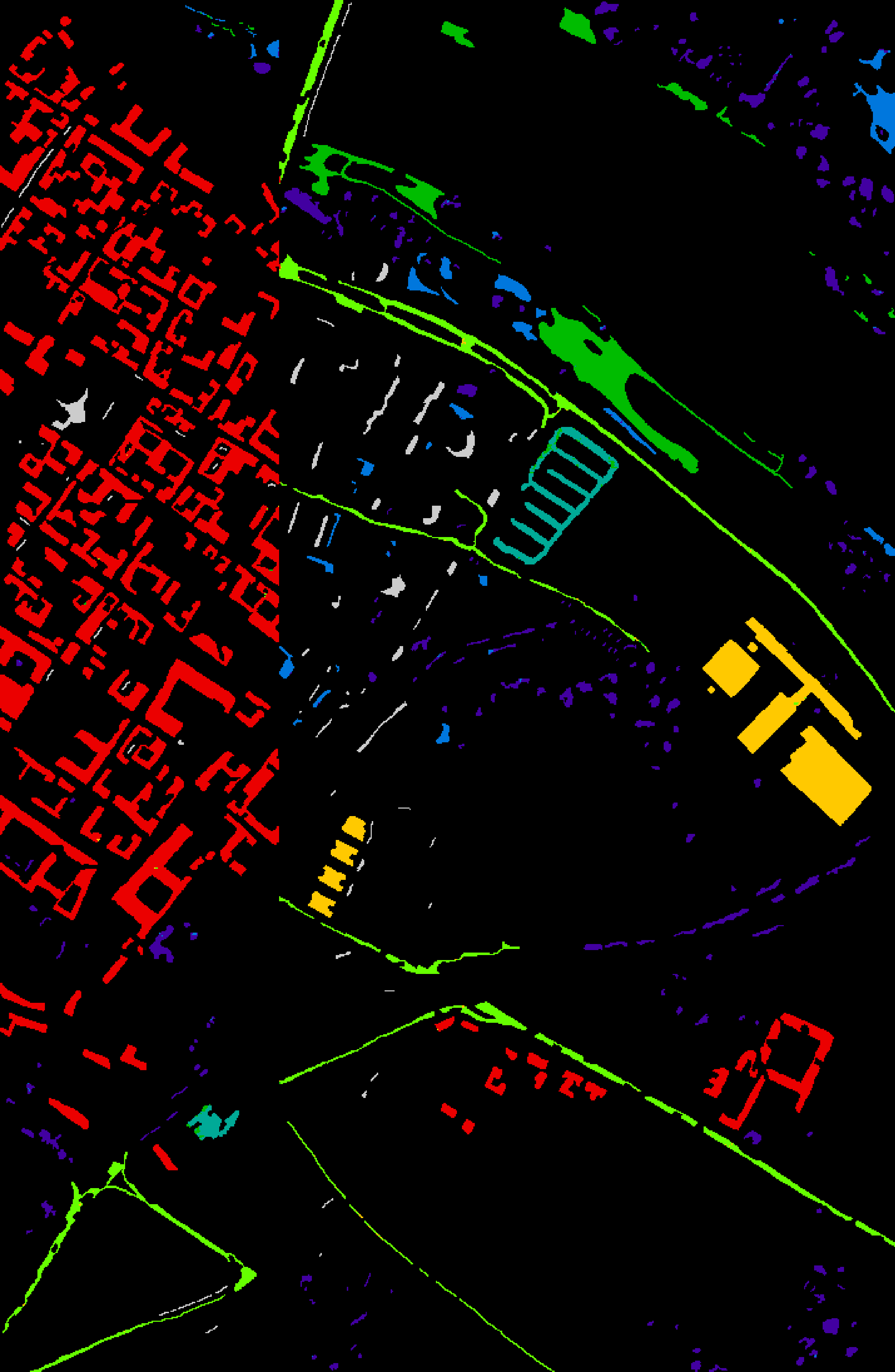}
            \caption{Hir-Transformer}
        \end{subfigure}
        \begin{subfigure}{0.15\textwidth}
            \centering
            \includegraphics[width=0.99\textwidth]{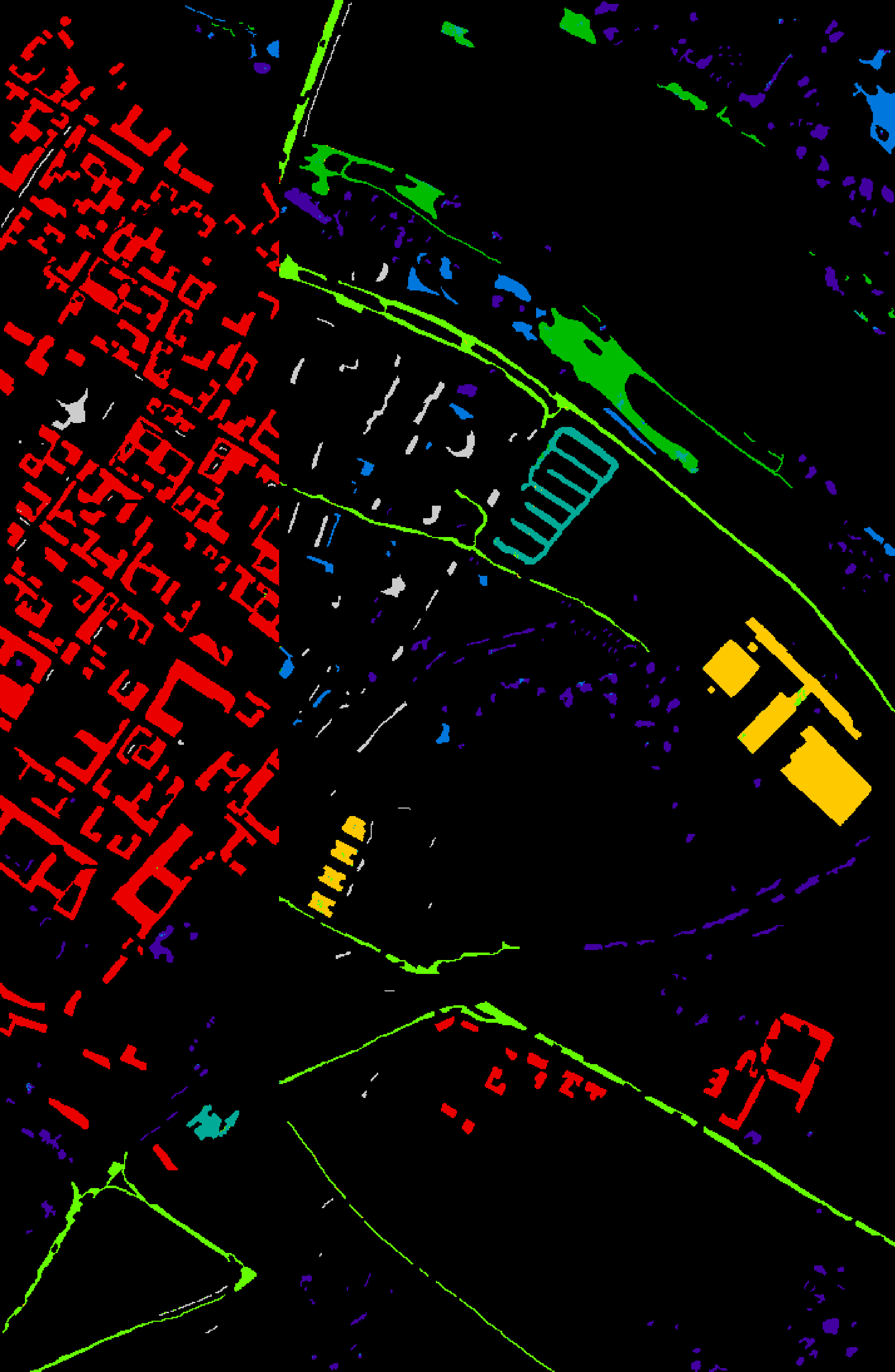}
            \caption{SSMamba}
        \end{subfigure}
        \begin{subfigure}{0.15\textwidth}
            \centering
            \includegraphics[width=0.99\textwidth]{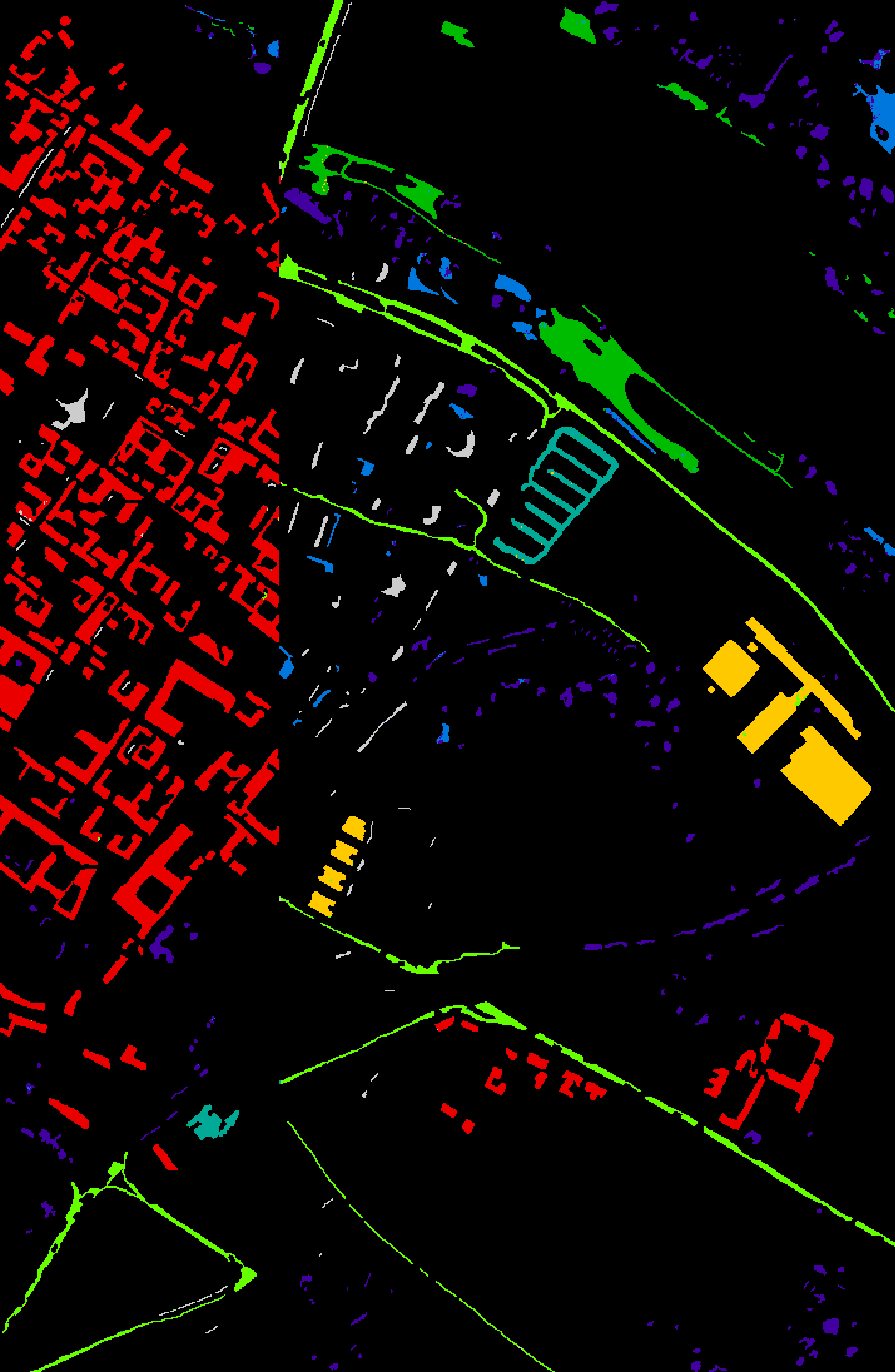}
            \caption{SMM}
        \end{subfigure}
        \begin{subfigure}{0.15\textwidth}
            \centering
            \includegraphics[width=0.99\textwidth]{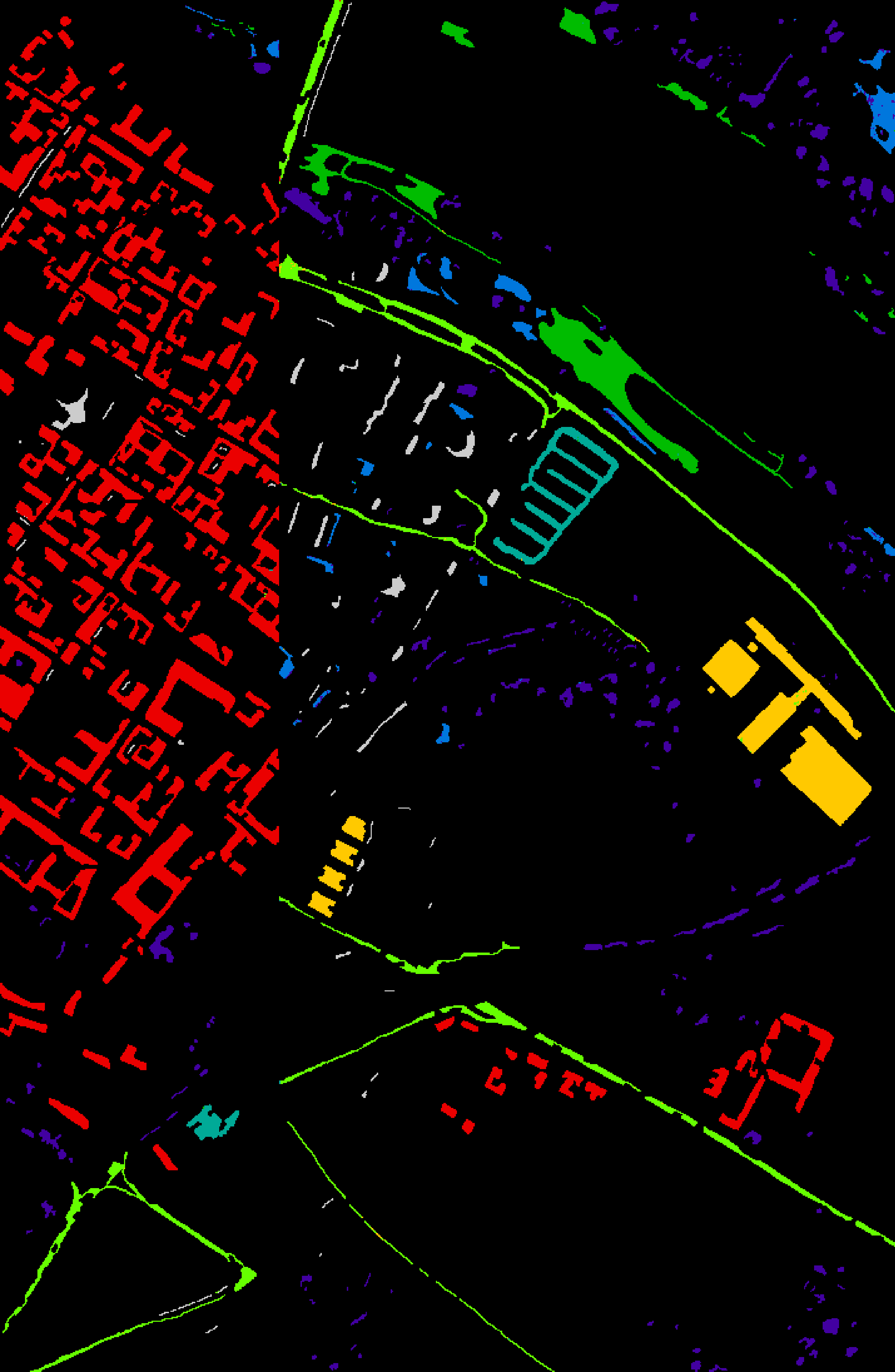}
            \caption{SSMM}
        \end{subfigure}
    \caption{\textbf{PC dataset:} The predicted ground truth maps for various competing methods alongside the proposed variants of the MorpMamba model.}
    \label{PC_results}
\end{figure*}
\begin{figure*}[!htb]
    \centering
        \begin{subfigure}{0.15\textwidth}
            \includegraphics[width=0.99\textwidth]{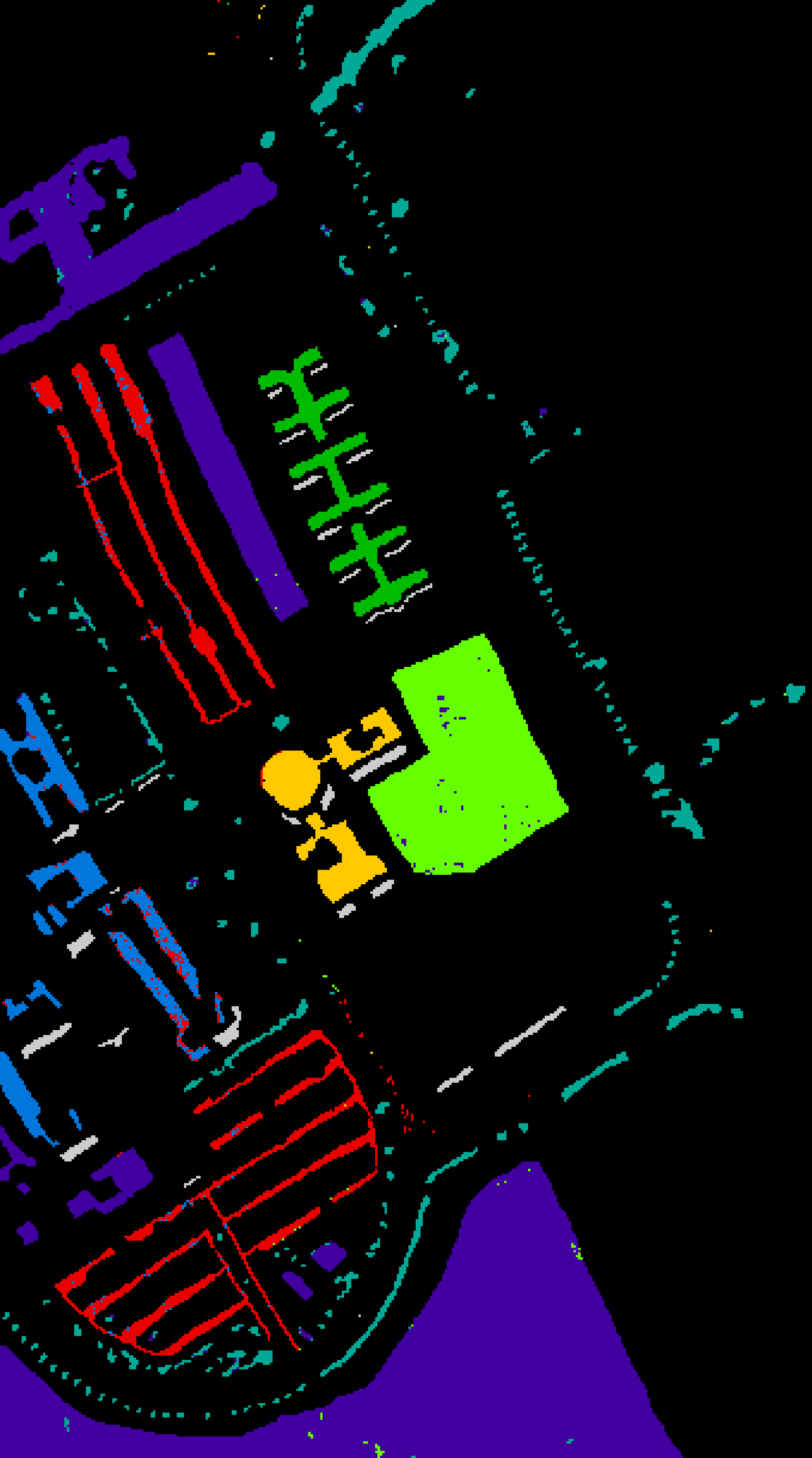}
            \caption{CNN2D}
        \end{subfigure}
        \begin{subfigure}{0.15\textwidth}
            \centering
            \includegraphics[width=0.99\textwidth]{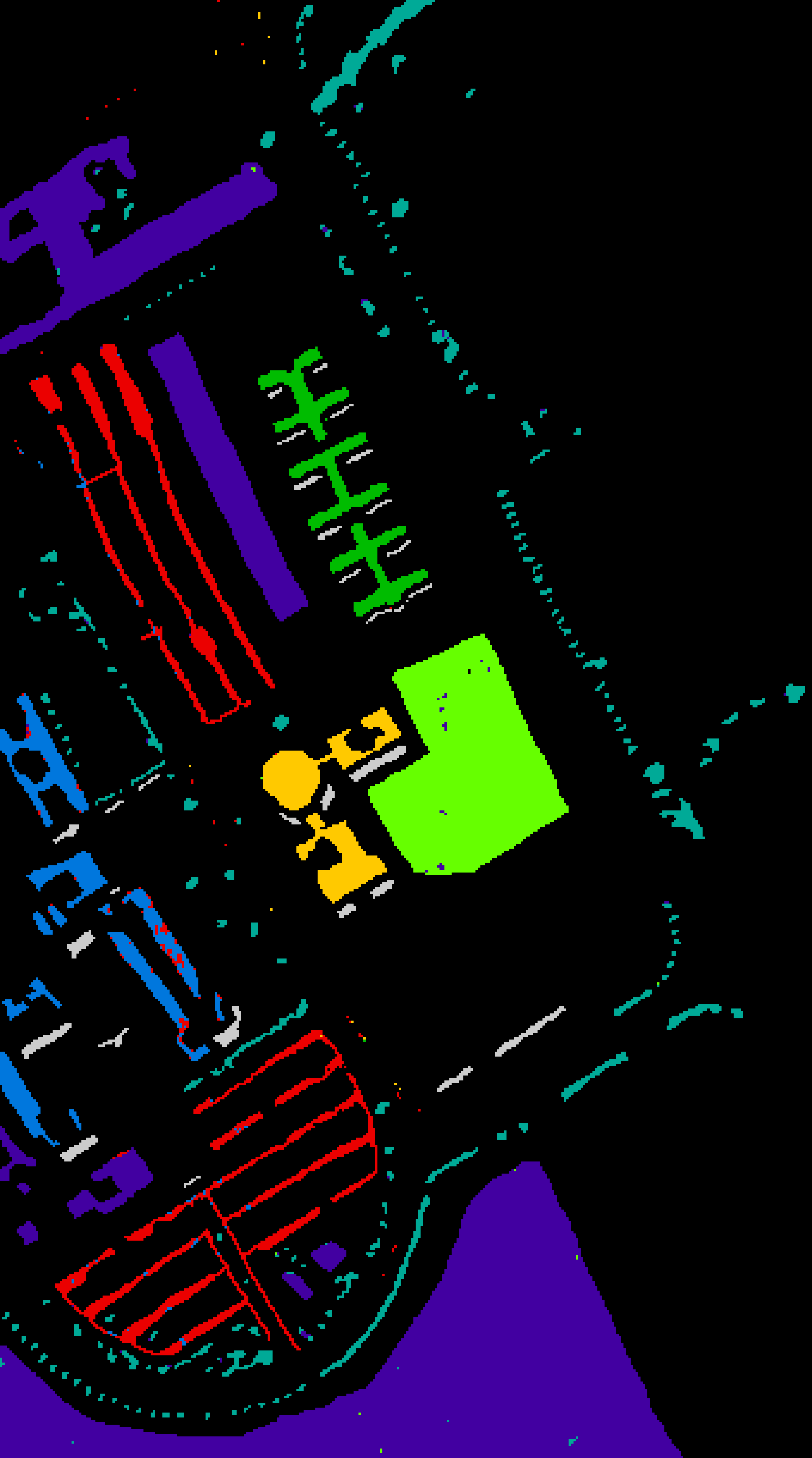}
            \caption{CNN3D}
        \end{subfigure}
        \begin{subfigure}{0.15\textwidth}
            \centering
            \includegraphics[width=0.99\textwidth]{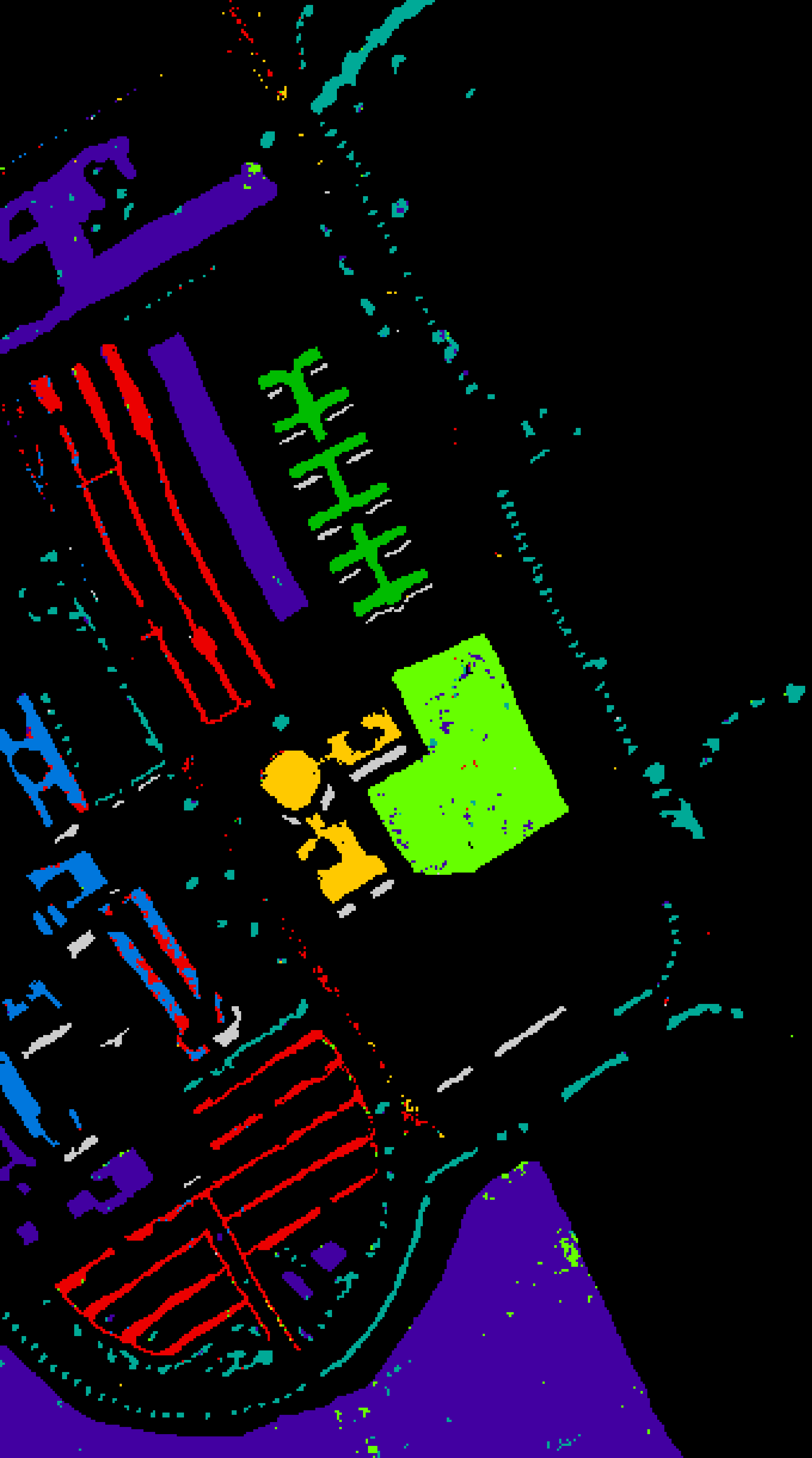}
            \caption{HybCNN}
        \end{subfigure}
        \begin{subfigure}{0.15\textwidth}
            \centering
            \includegraphics[width=0.99\textwidth]{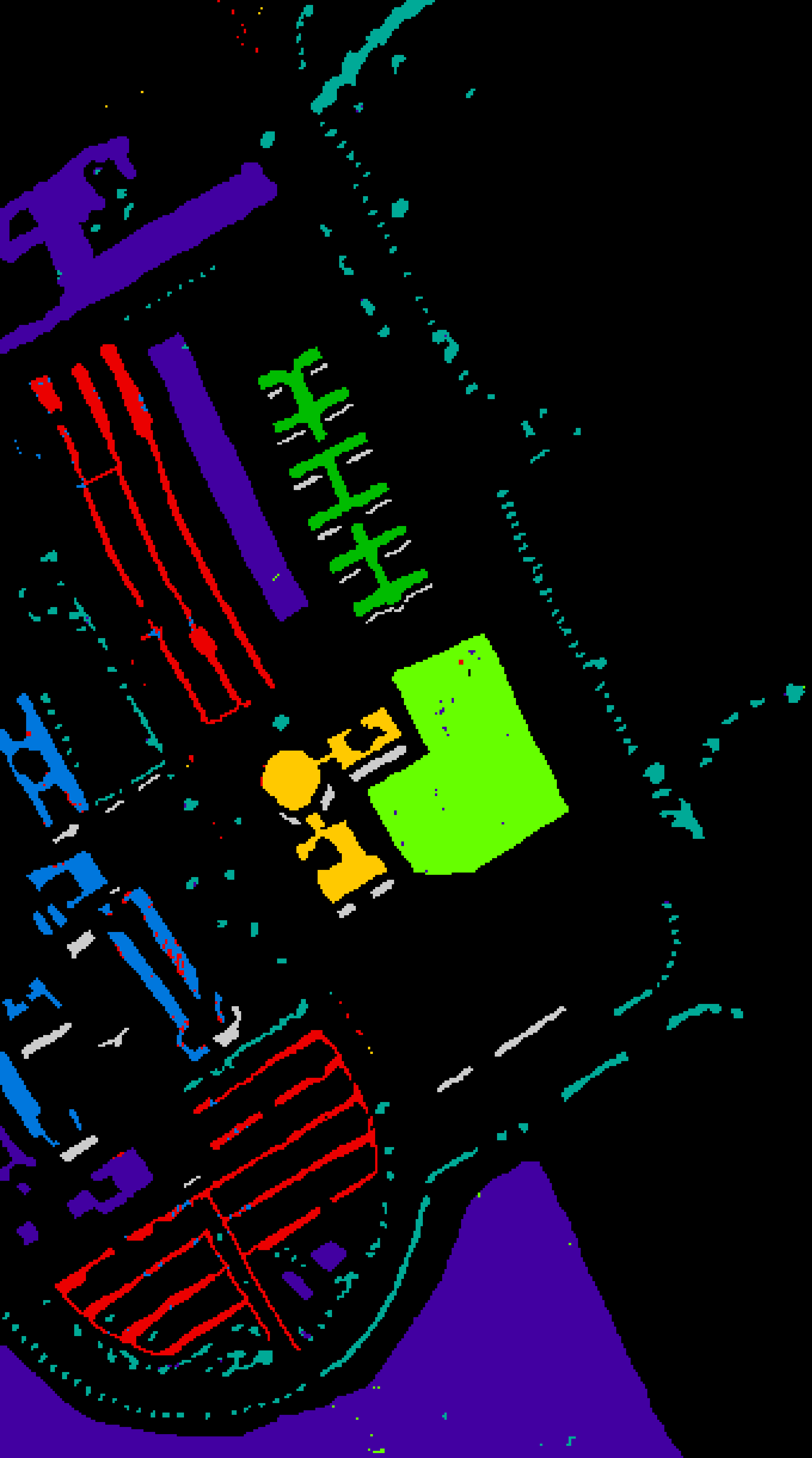}
            \caption{IN2D}
        \end{subfigure}
        \begin{subfigure}{0.15\textwidth}
            \centering
            \includegraphics[width=0.99\textwidth]{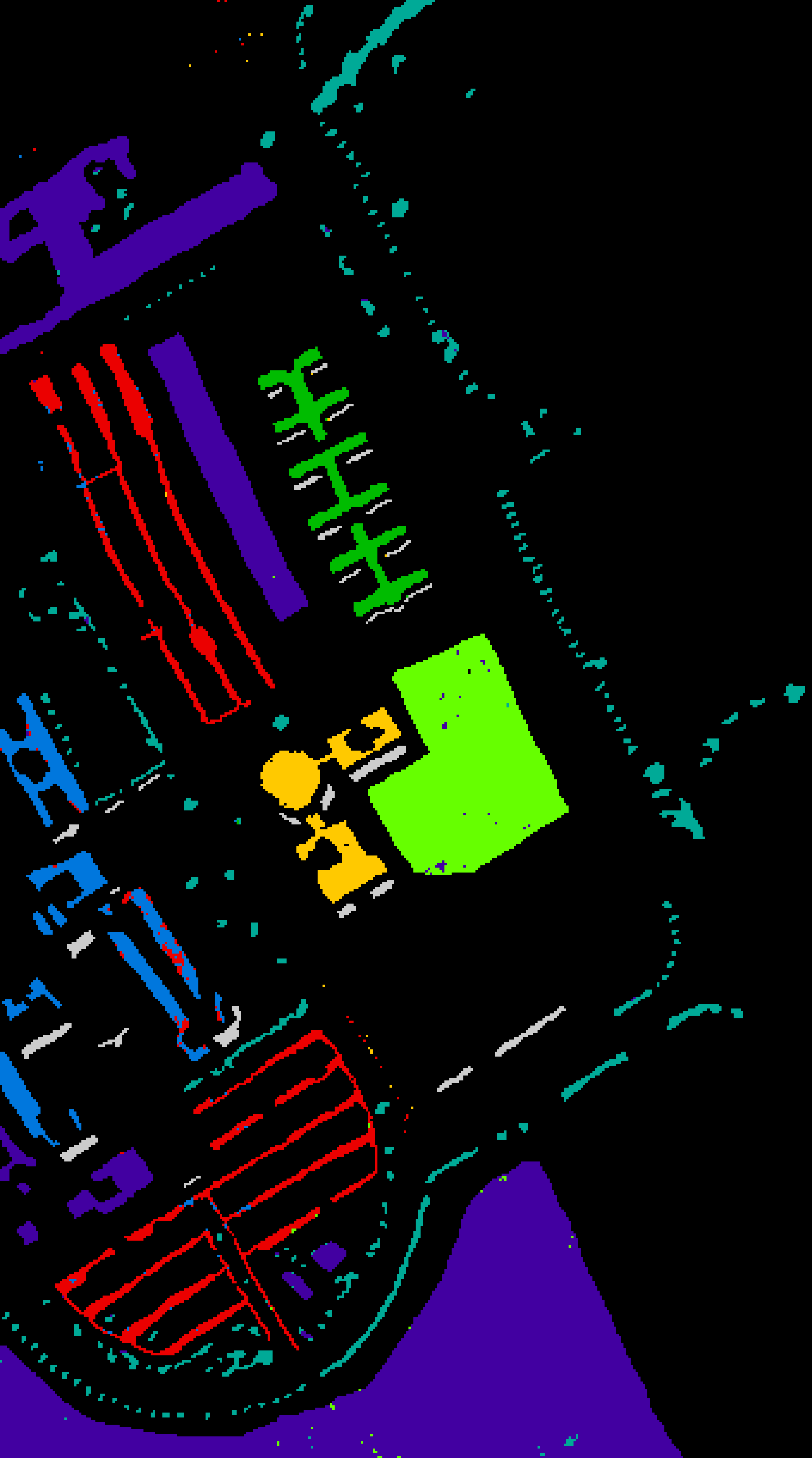}
            \caption{IN3D}
        \end{subfigure}
        \begin{subfigure}{0.15\textwidth}
            \centering
            \includegraphics[width=0.99\textwidth]{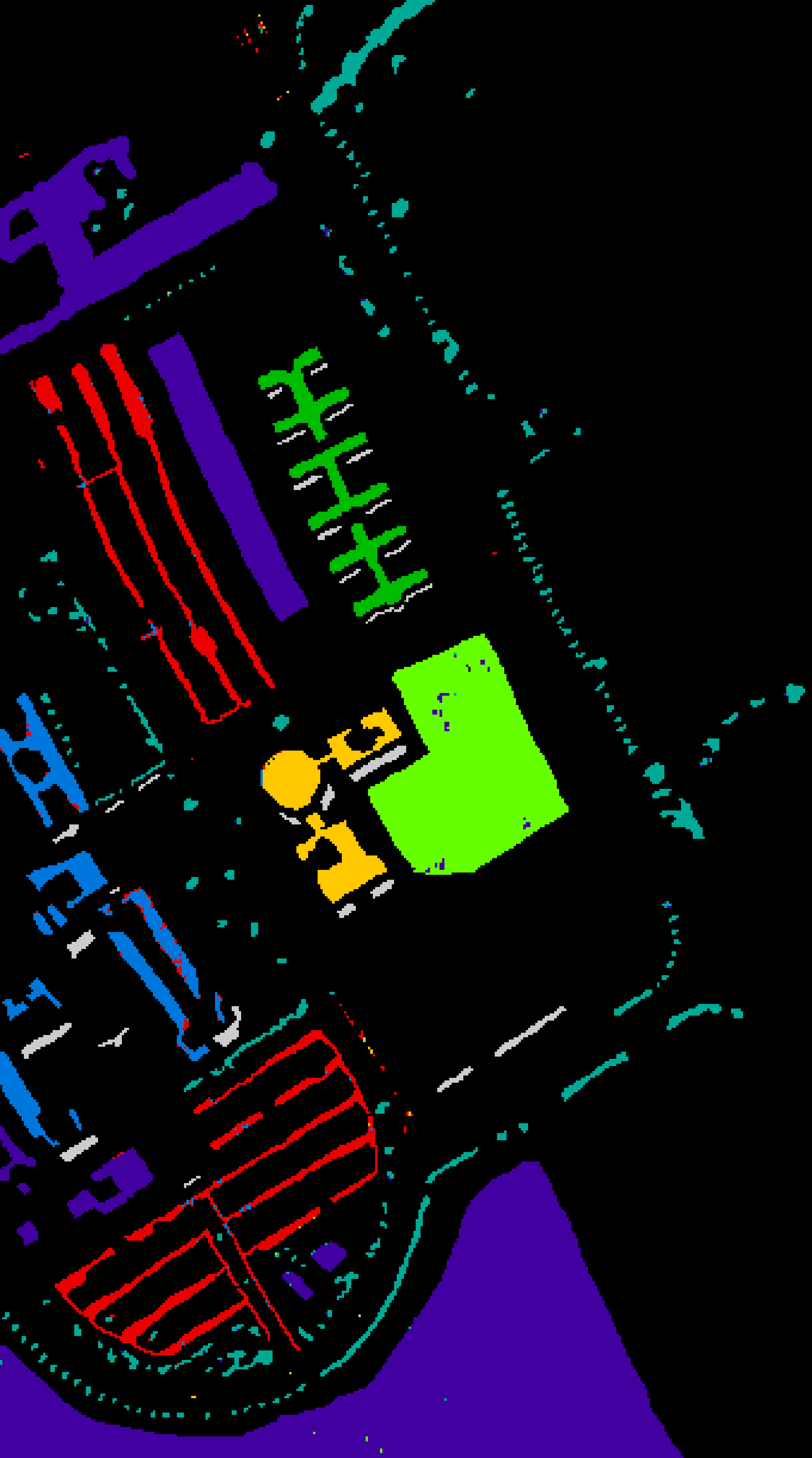}
            \caption{HybIN}
        \end{subfigure}
        \begin{subfigure}{0.15\textwidth}
            \centering
            \includegraphics[width=0.99\textwidth]{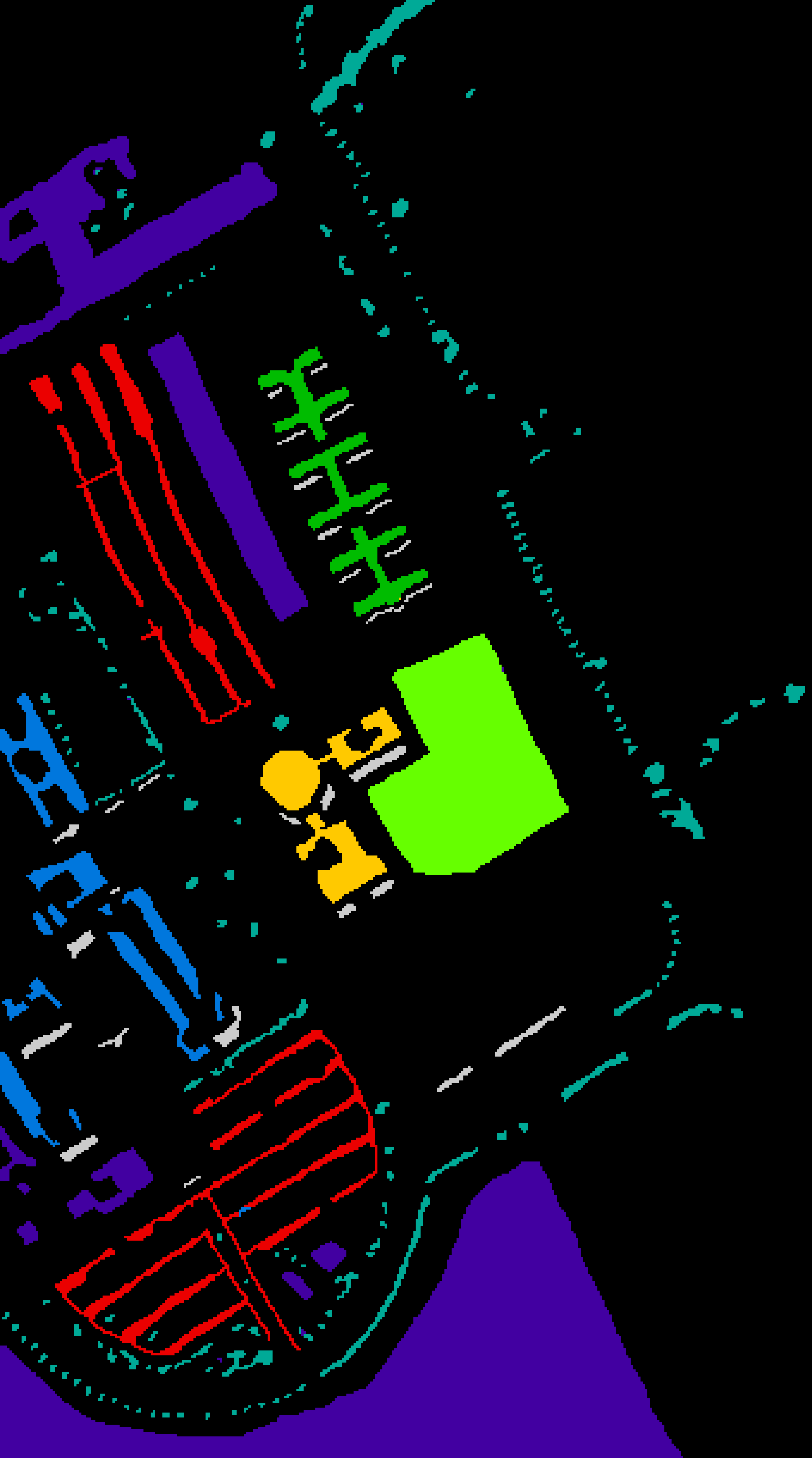}
            \caption{MorpCNN}
        \end{subfigure}
        \begin{subfigure}{0.15\textwidth}
            \centering
            \includegraphics[width=0.99\textwidth]{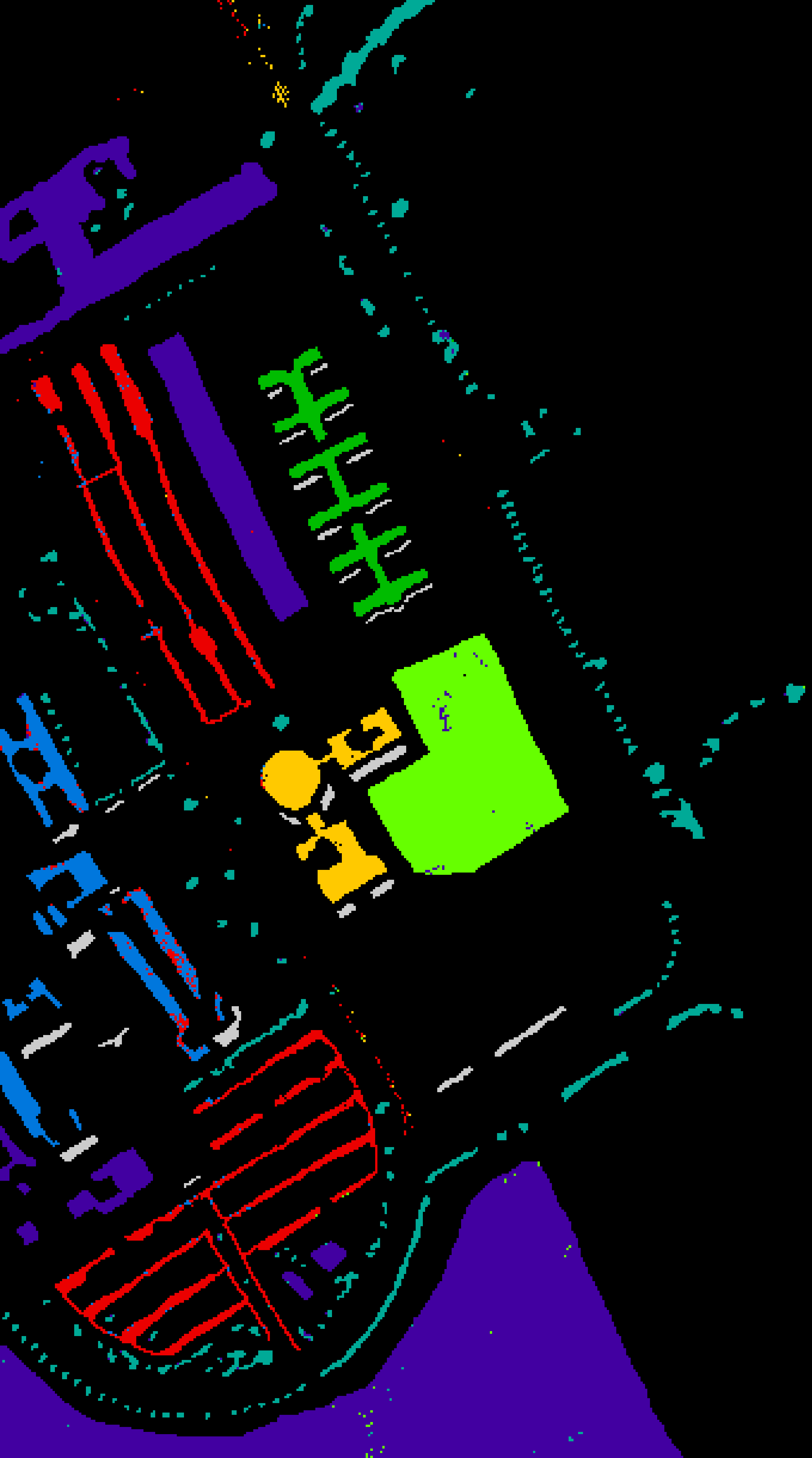}
            \caption{Hybrid-ViT}
        \end{subfigure}
        \begin{subfigure}{0.15\textwidth}
            \centering
            \includegraphics[width=0.99\textwidth]{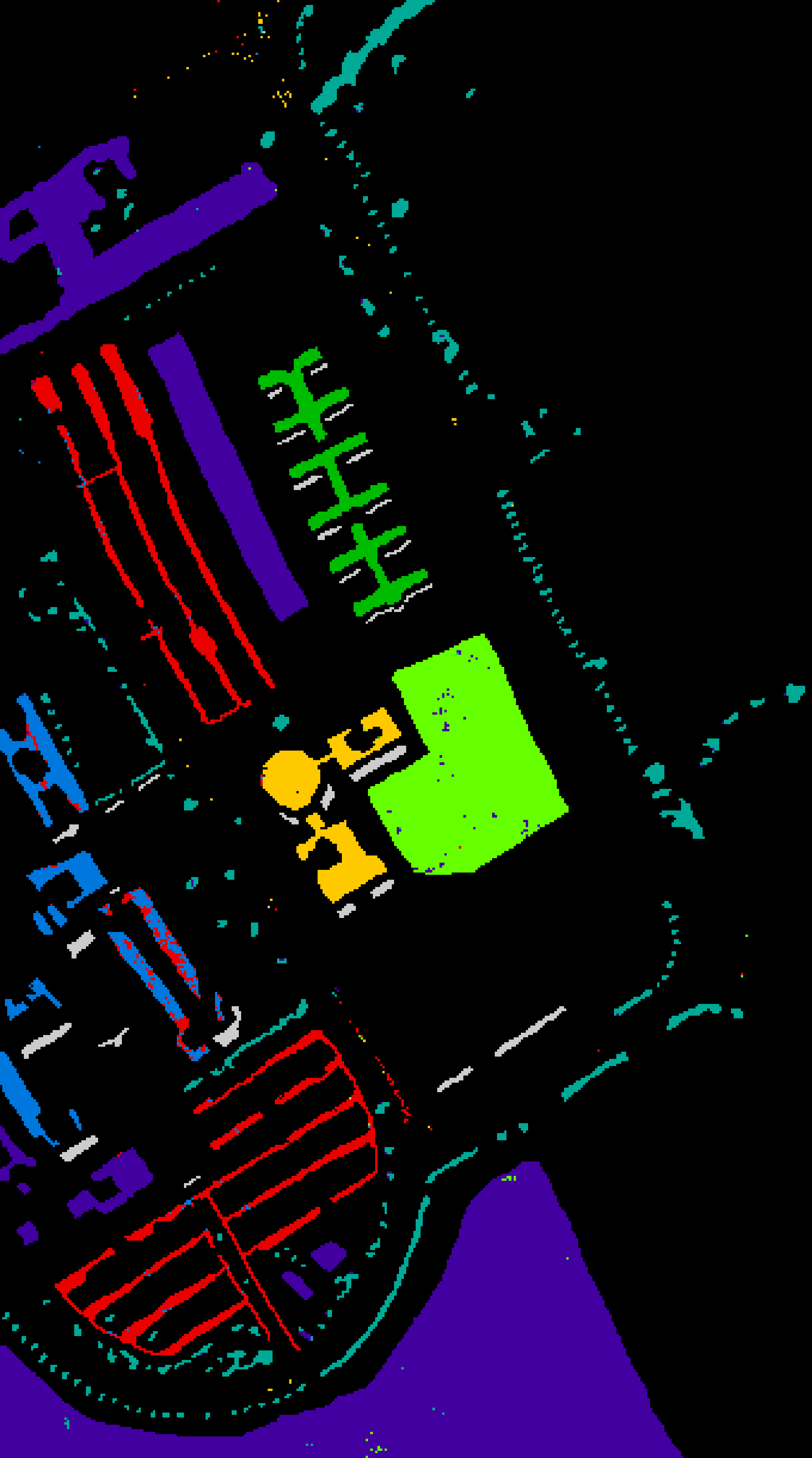}
            \caption{Hir-Transformer}
        \end{subfigure}
        \begin{subfigure}{0.15\textwidth}
            \centering
            \includegraphics[width=0.99\textwidth]{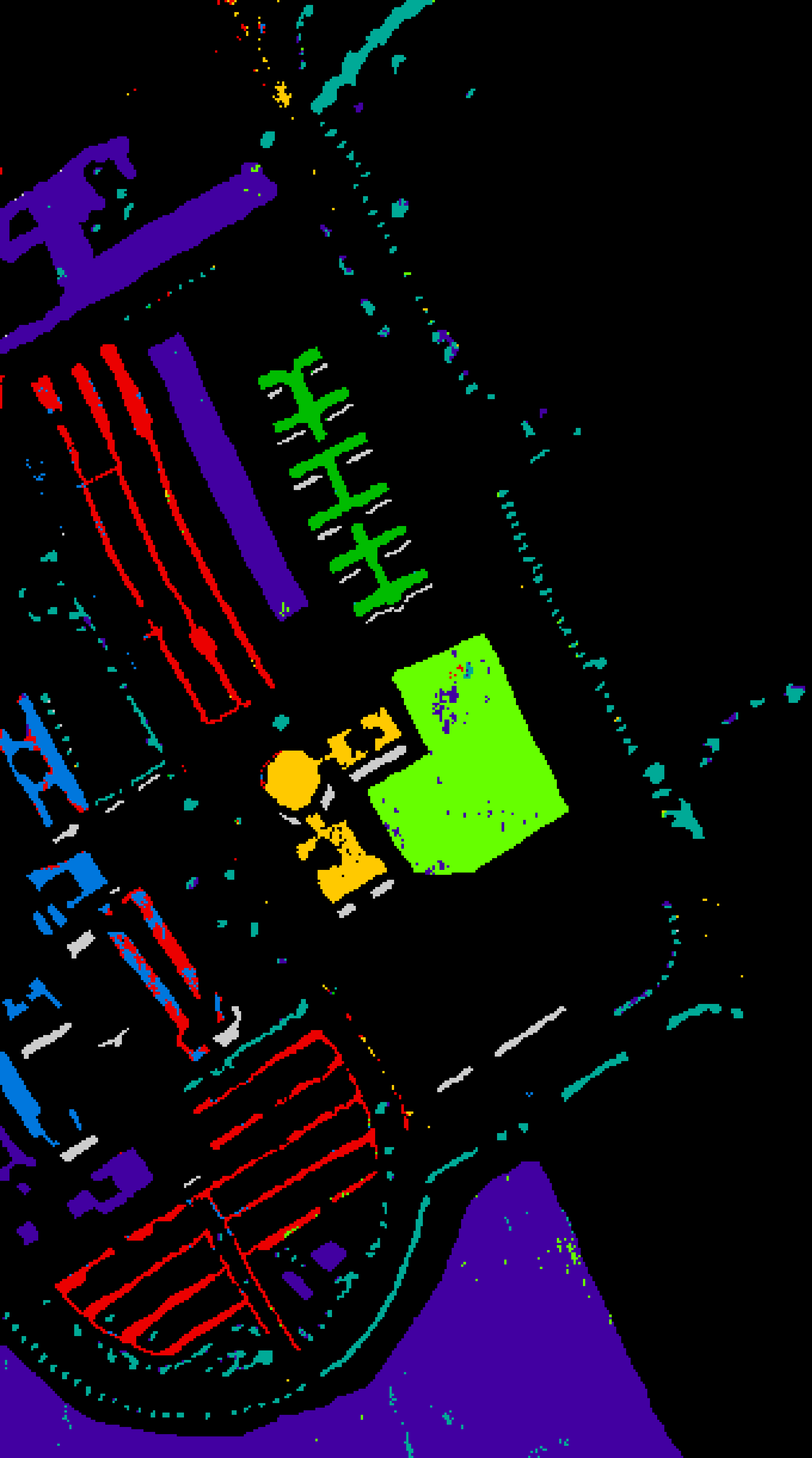}
            \caption{SSMamba}
        \end{subfigure}
        \begin{subfigure}{0.15\textwidth}
            \centering
            \includegraphics[width=0.99\textwidth]{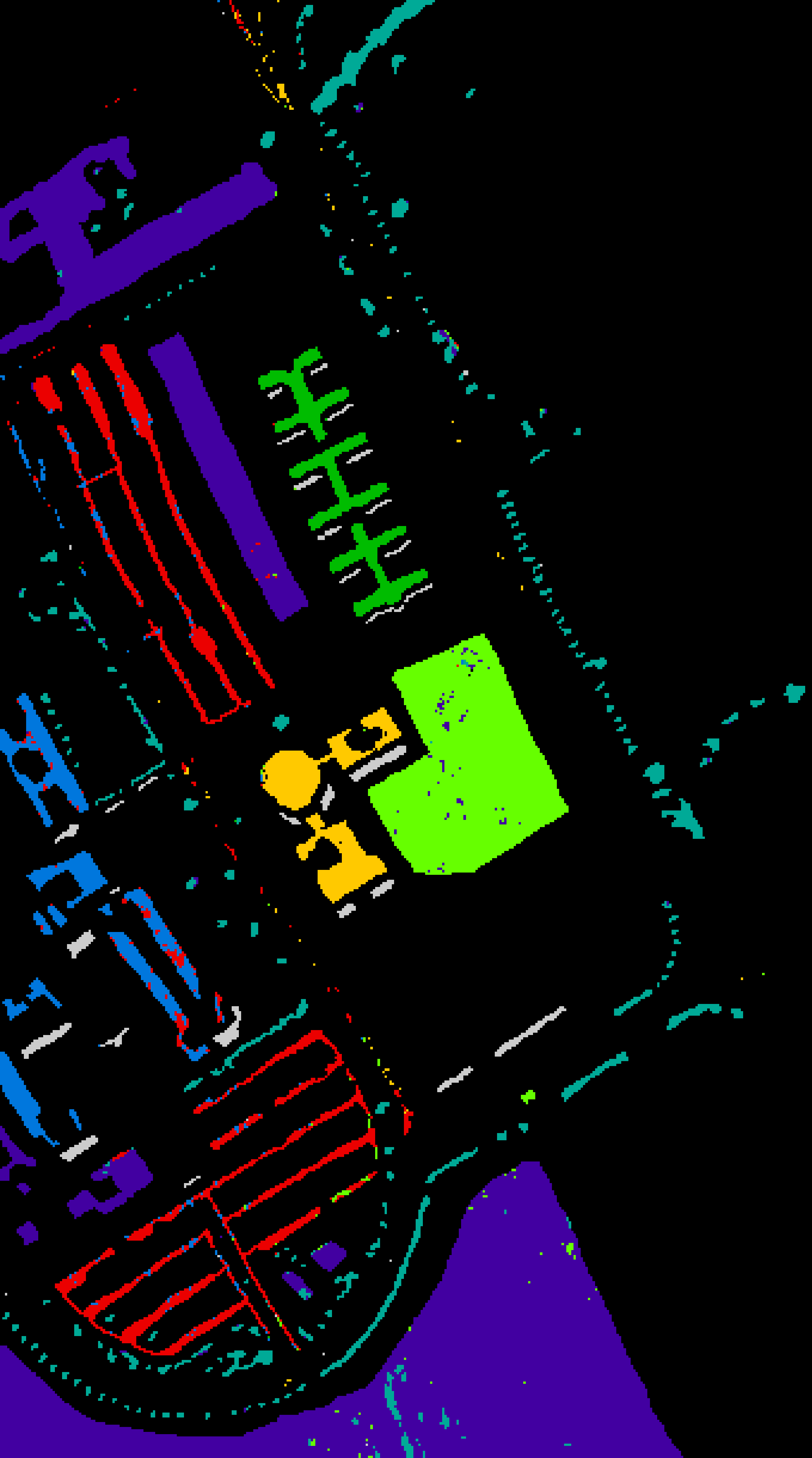}
            \caption{SMM}
        \end{subfigure}
        \begin{subfigure}{0.15\textwidth}
            \centering
            \includegraphics[width=0.99\textwidth]{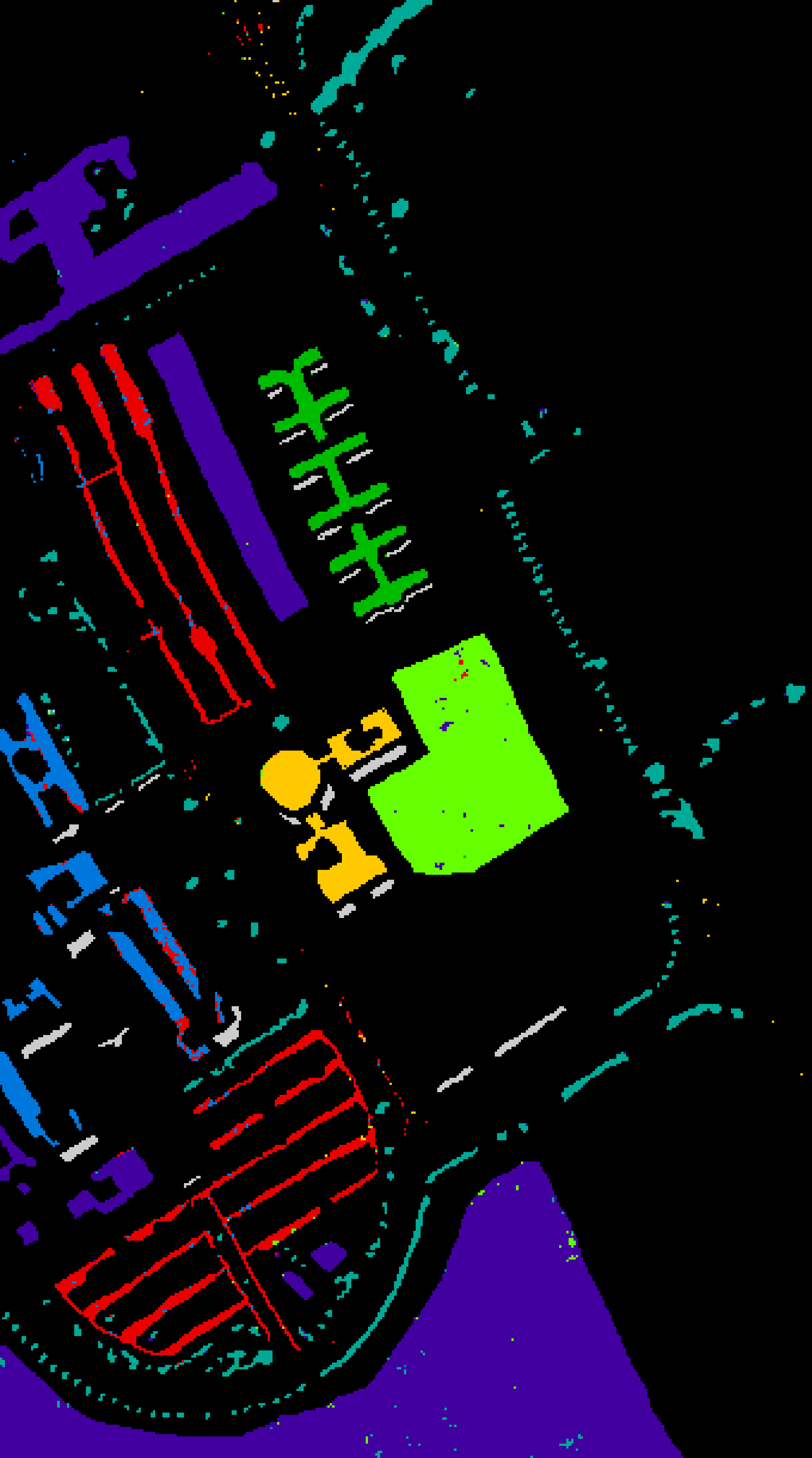}
            \caption{SSMM}
        \end{subfigure}
    \caption{\textbf{PU dataset:} The predicted ground truth maps for various competing methods alongside the proposed variants of the MorpMamba model.}
    \label{PU_results}
\end{figure*}
\begin{figure*}[!htb]
    \centering
        \begin{subfigure}{0.15\textwidth}
            \includegraphics[width=0.99\textwidth]{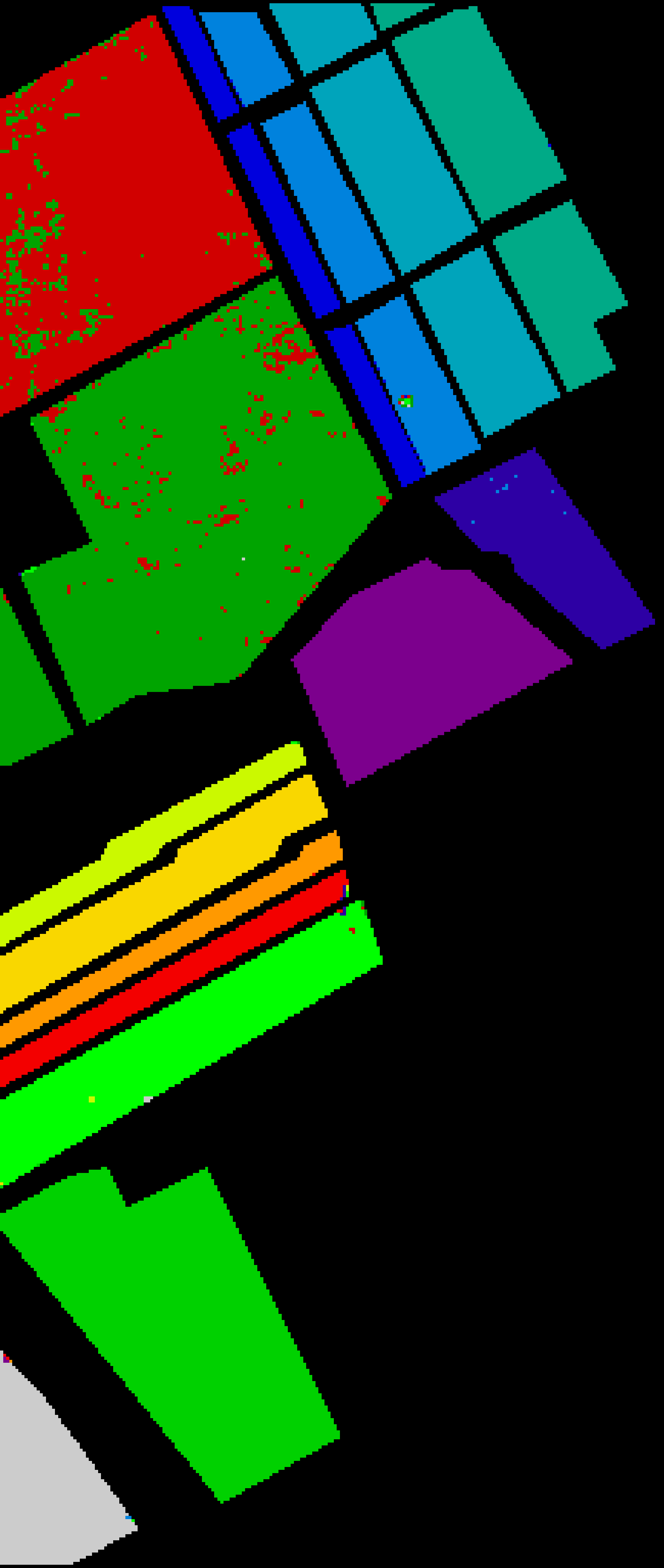}
            \caption{CNN2D}
        \end{subfigure}
        \begin{subfigure}{0.15\textwidth}
            \centering
            \includegraphics[width=0.99\textwidth]{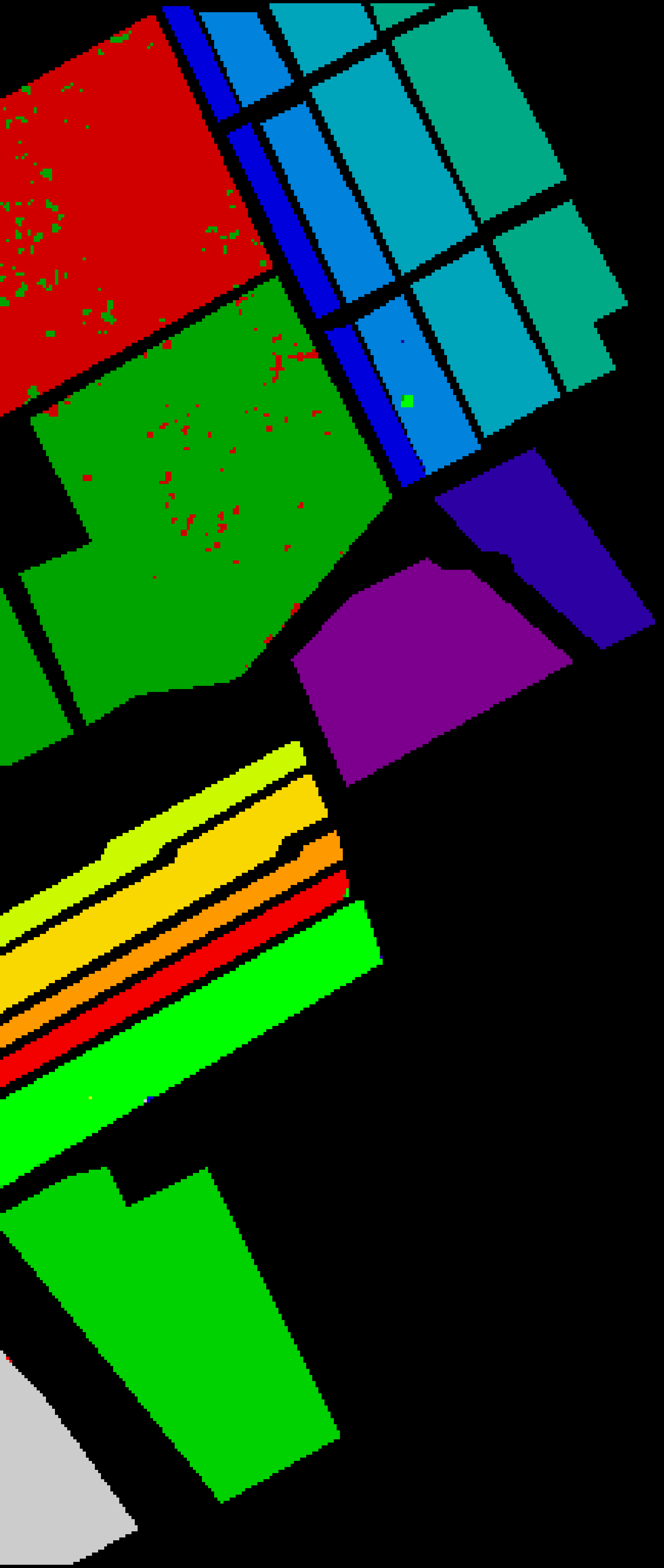}
            \caption{CNN3D}
        \end{subfigure}
        \begin{subfigure}{0.15\textwidth}
            \centering
            \includegraphics[width=0.99\textwidth]{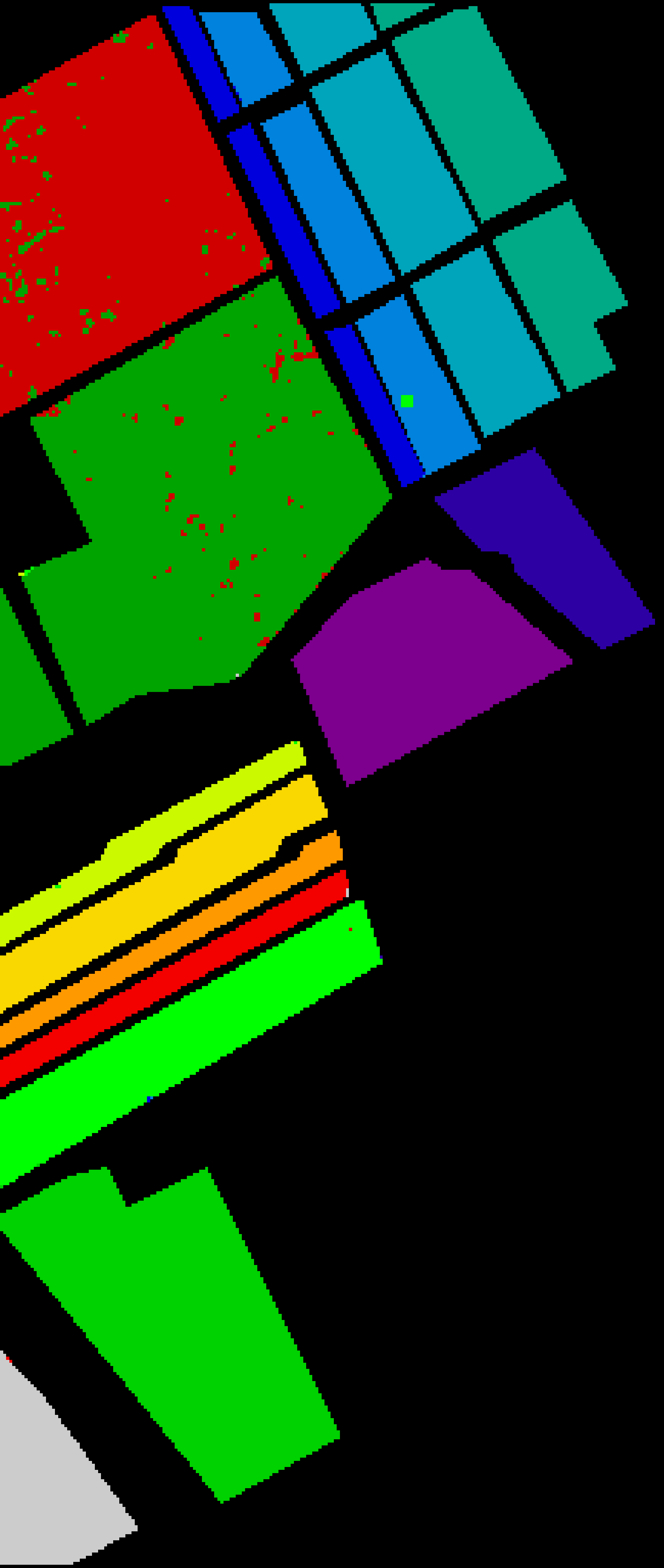}
            \caption{HybCNN}
        \end{subfigure}
        \begin{subfigure}{0.15\textwidth}
            \centering
            \includegraphics[width=0.99\textwidth]{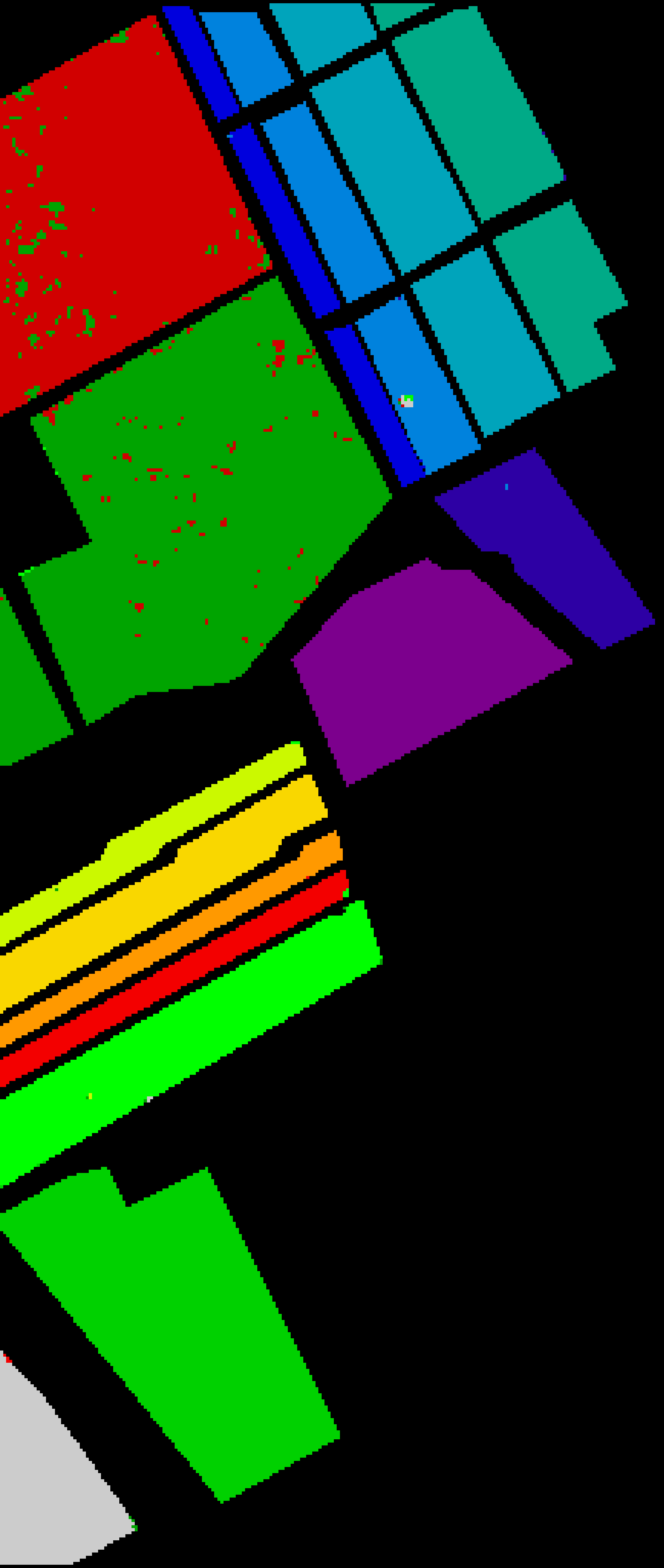}
            \caption{IN2D}
        \end{subfigure}
        \begin{subfigure}{0.15\textwidth}
            \centering
            \includegraphics[width=0.99\textwidth]{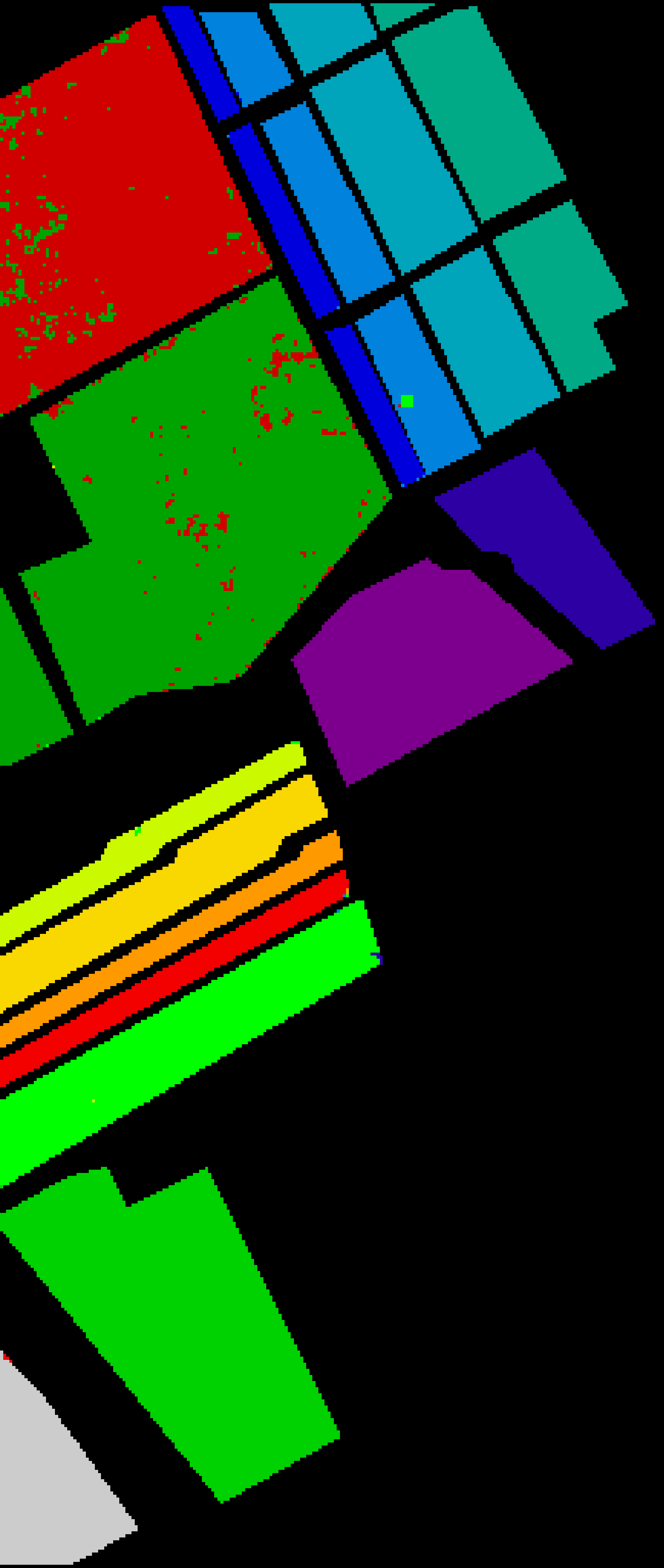}
            \caption{IN3D}
        \end{subfigure}
        \begin{subfigure}{0.15\textwidth}
            \centering
            \includegraphics[width=0.99\textwidth]{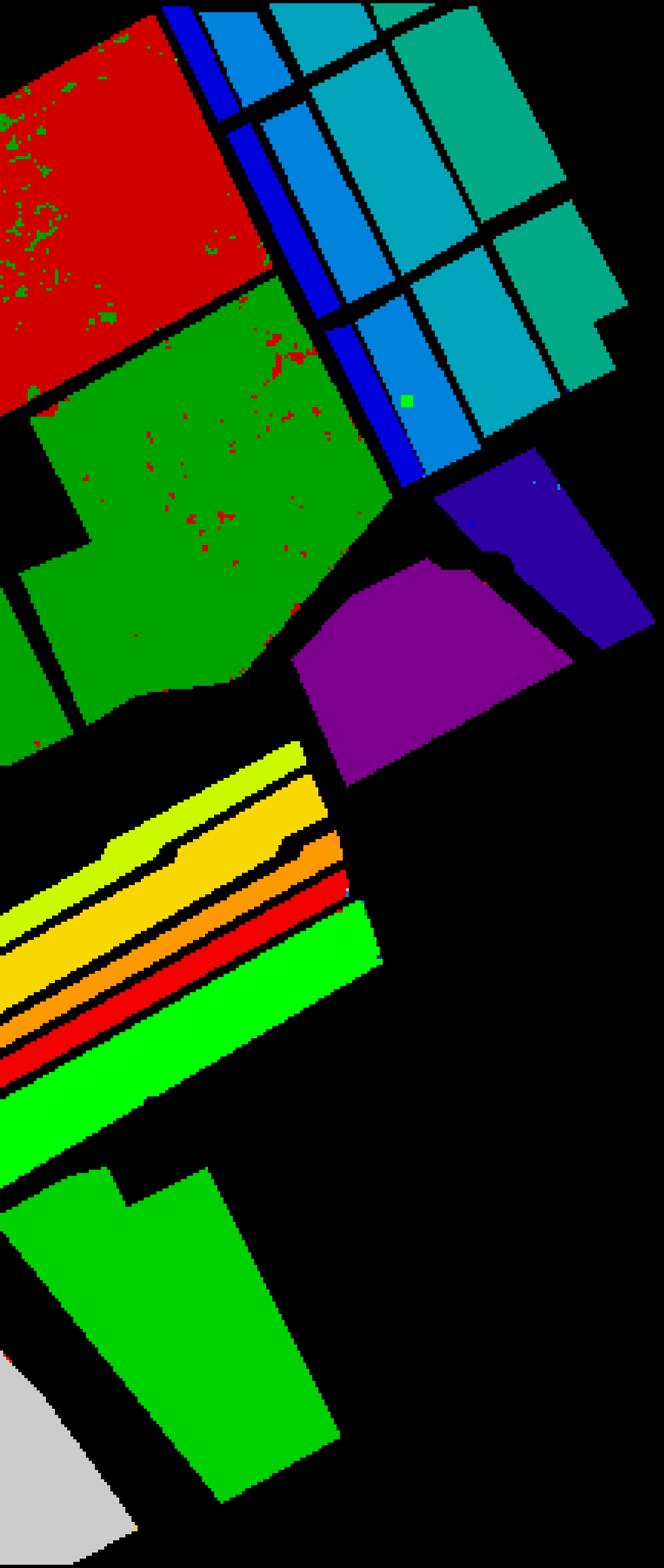}
            \caption{HybIN}
        \end{subfigure}
        \begin{subfigure}{0.15\textwidth}
            \centering
            \includegraphics[width=0.99\textwidth]{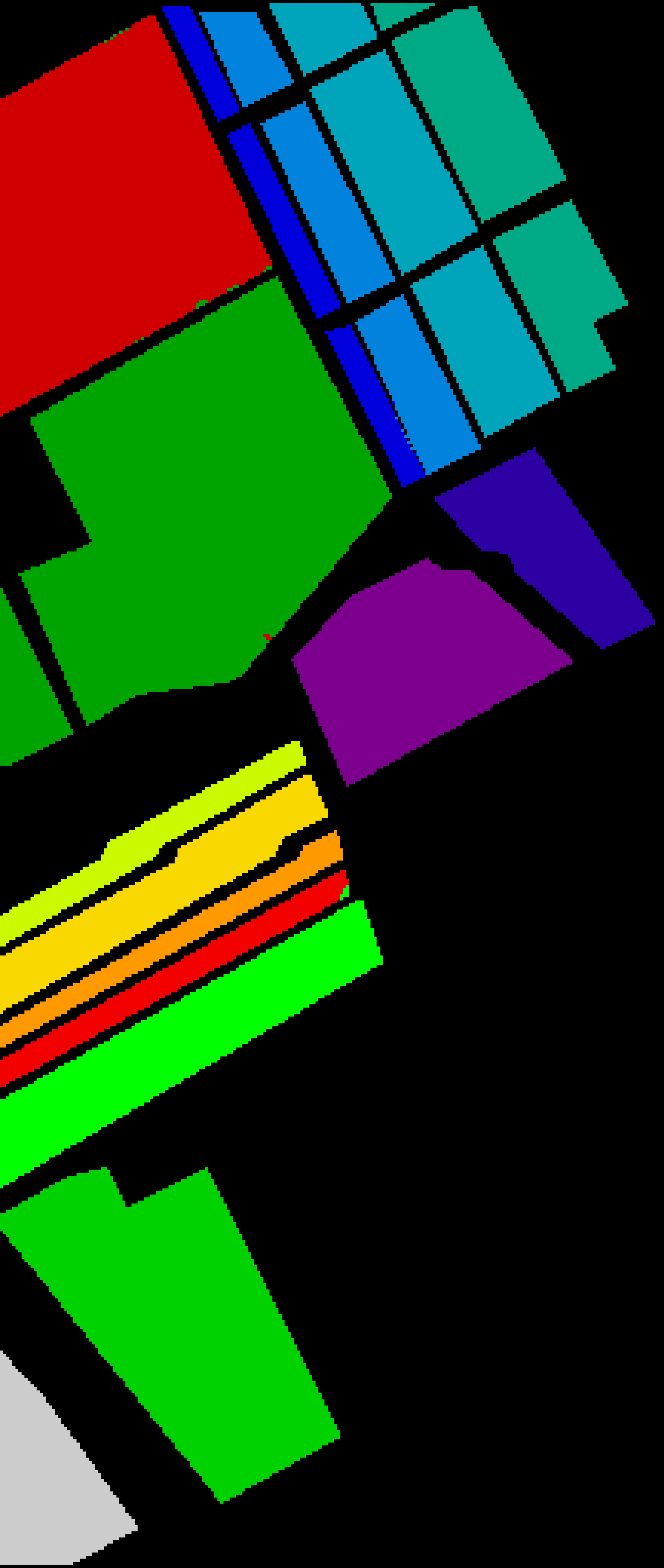}
            \caption{MorpCNN}
        \end{subfigure}
        \begin{subfigure}{0.15\textwidth}
            \centering
            \includegraphics[width=0.99\textwidth]{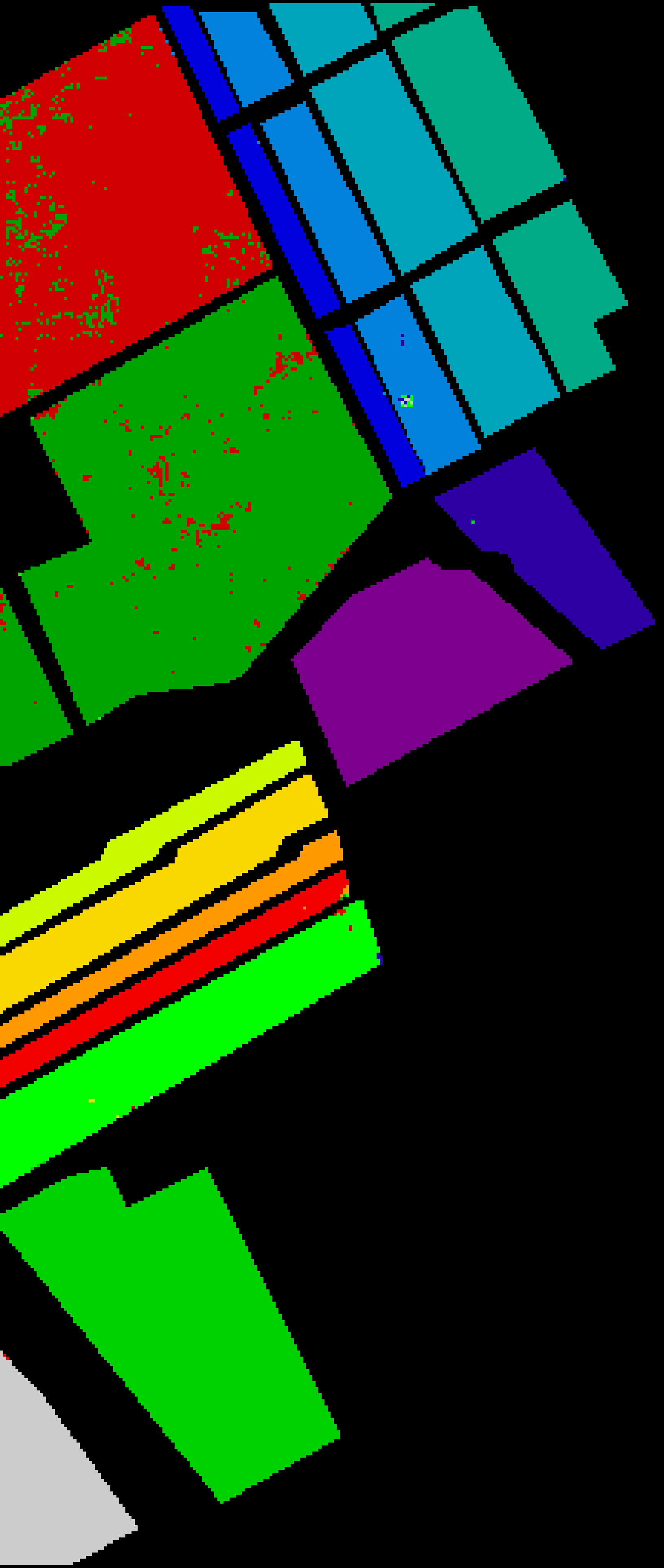}
            \caption{Hybrid-ViT}
        \end{subfigure}
        \begin{subfigure}{0.15\textwidth}
            \centering
            \includegraphics[width=0.99\textwidth]{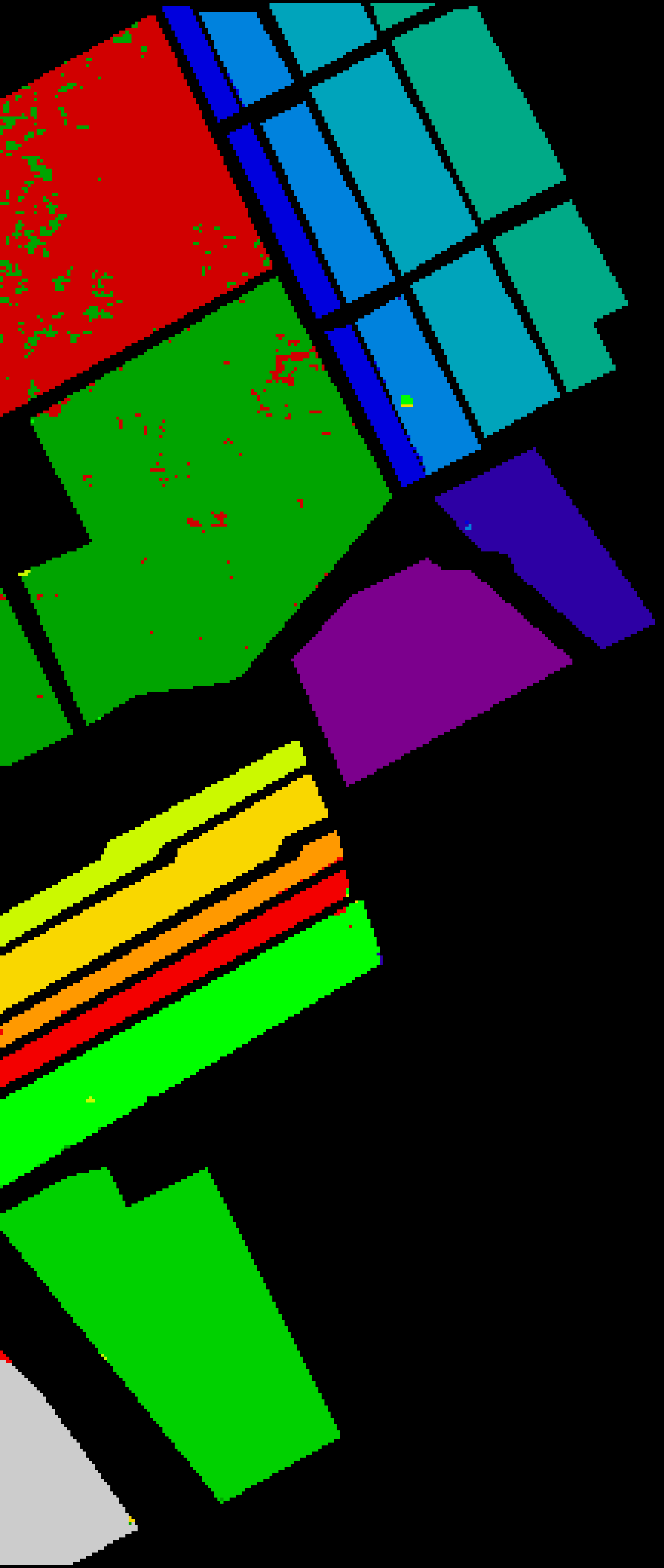}
            \caption{Hir-Transformer}
        \end{subfigure}
        \begin{subfigure}{0.15\textwidth}
            \centering
            \includegraphics[width=0.99\textwidth]{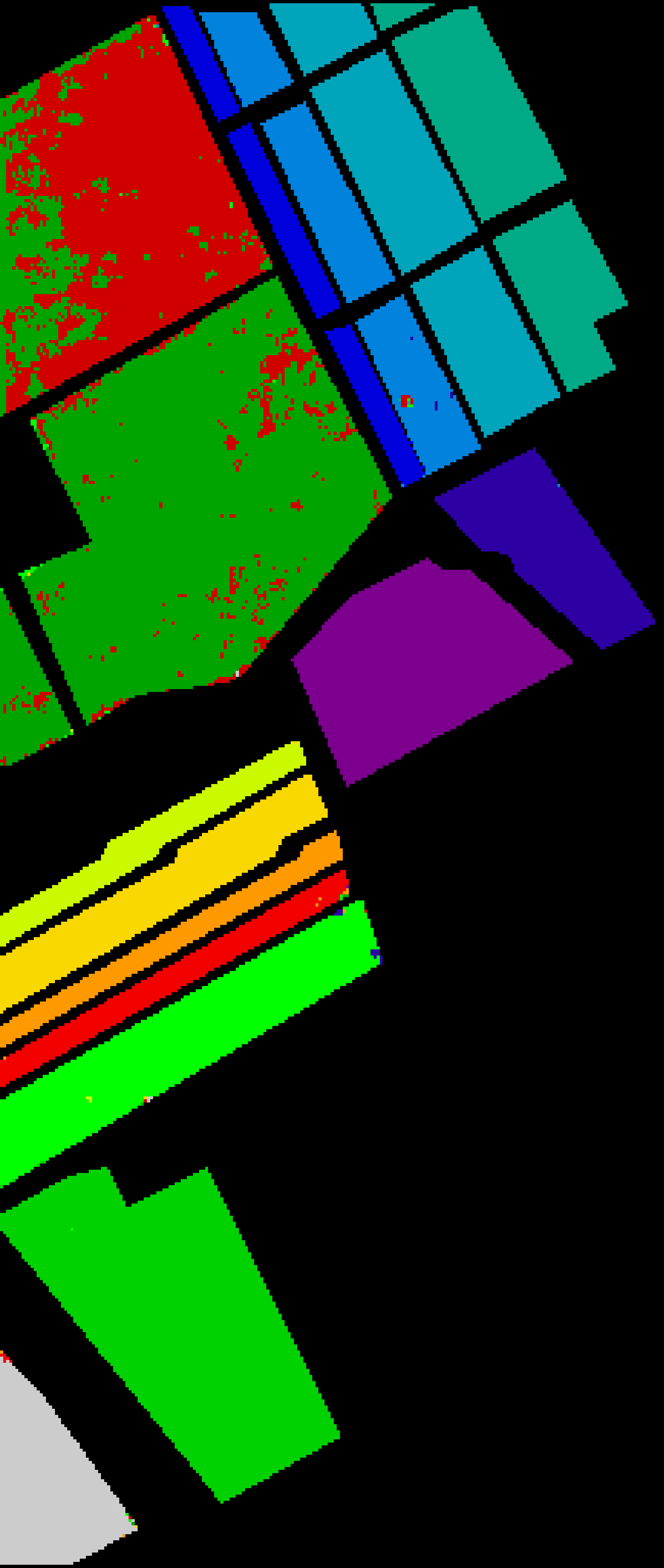}
            \caption{SSMamba}
        \end{subfigure} 
        \begin{subfigure}{0.15\textwidth}
            \centering
            \includegraphics[width=0.99\textwidth]{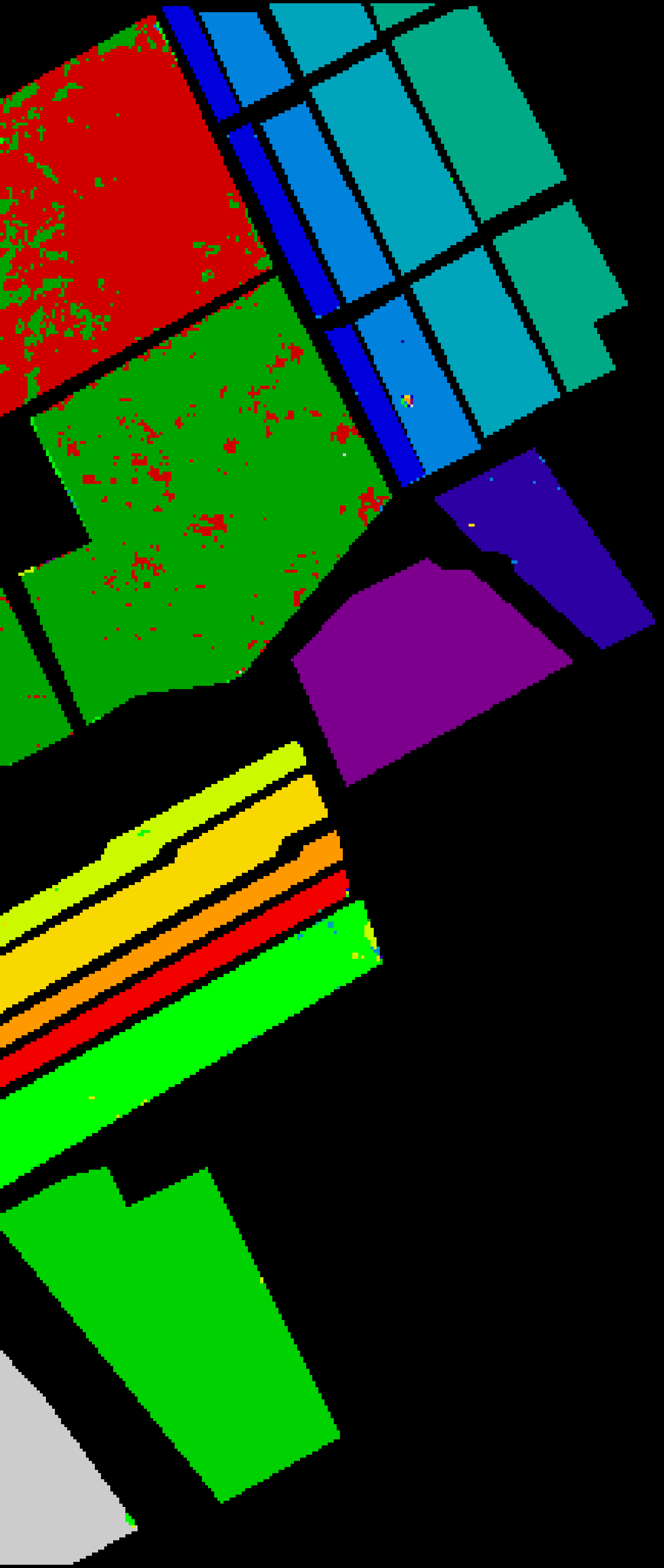}
            \caption{SSM}
        \end{subfigure} 
        \begin{subfigure}{0.15\textwidth}
            \centering
            \includegraphics[width=0.99\textwidth]{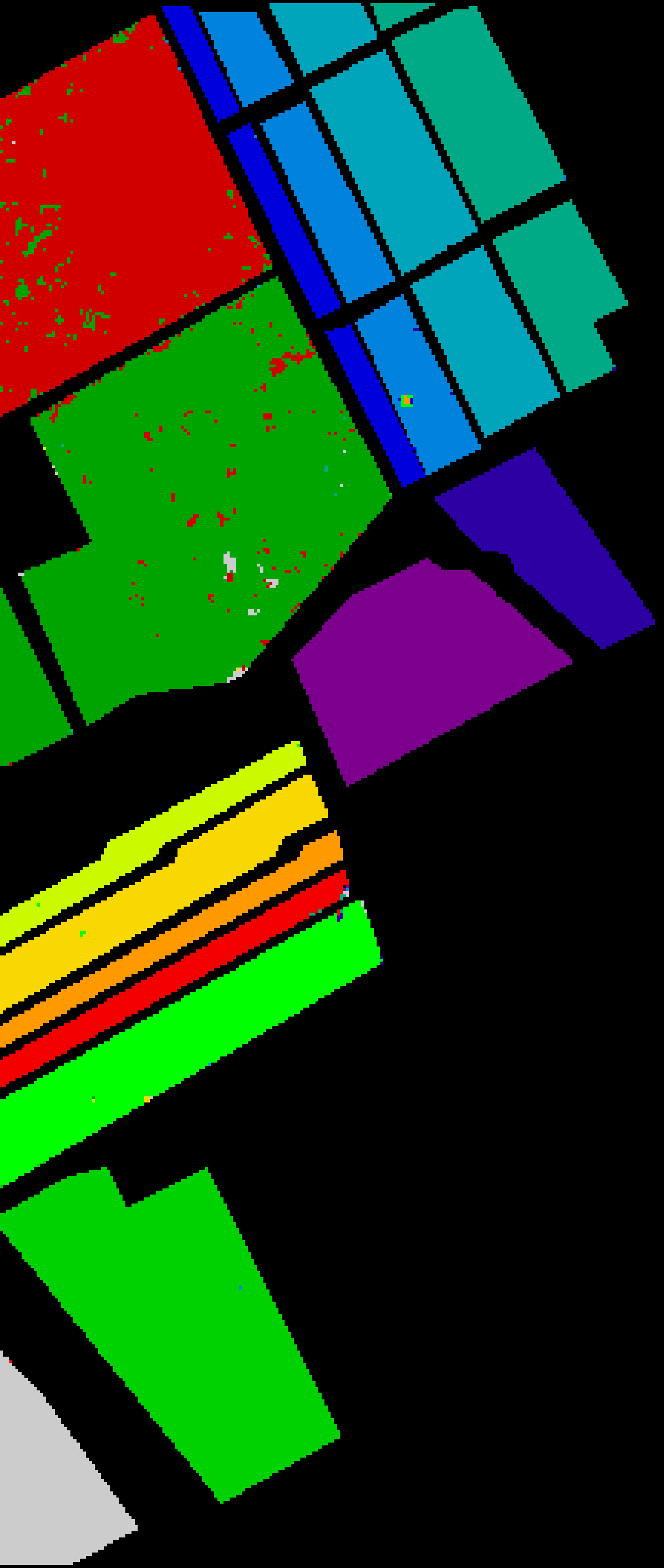}
            \caption{SSMM}
        \end{subfigure}
    \caption{\textbf{SA dataset:} The predicted ground truth maps for various competing methods alongside the proposed variants of the MorpMamba model.}
    \label{SA_results}
\end{figure*}
\begin{figure*}[!htb]
    \centering
        \begin{subfigure}{0.32\textwidth}
            \includegraphics[width=0.99\textwidth]{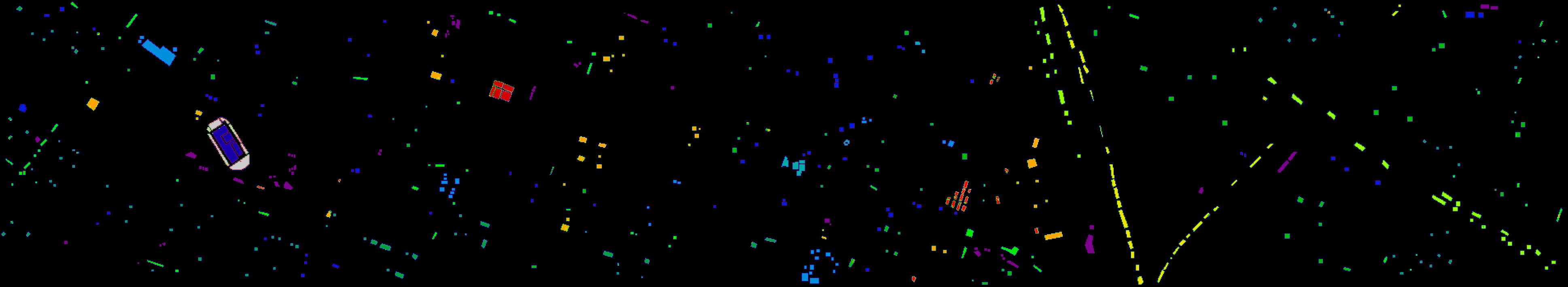}
            \caption{CNN2D}
        \end{subfigure}
        \begin{subfigure}{0.32\textwidth}
            \centering
            \includegraphics[width=0.99\textwidth]{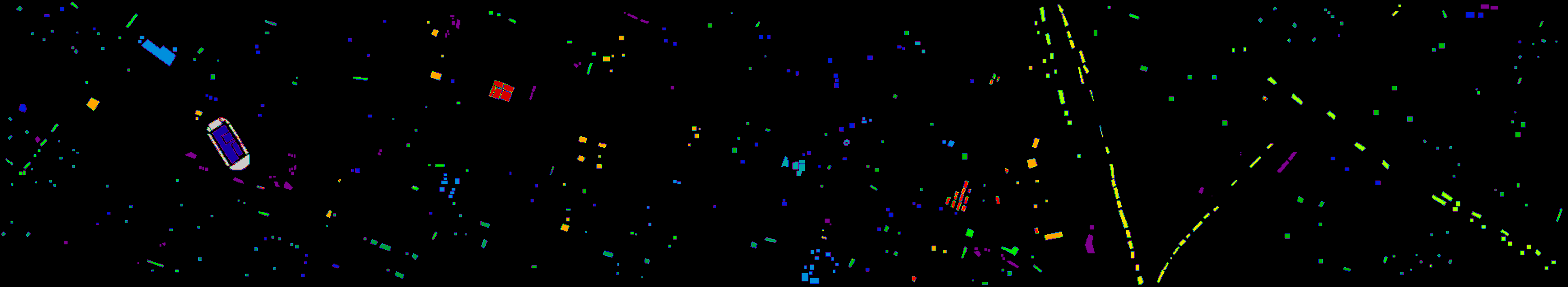}
            \caption{CNN3D}
        \end{subfigure}
        \begin{subfigure}{0.32\textwidth}
            \centering
            \includegraphics[width=0.99\textwidth]{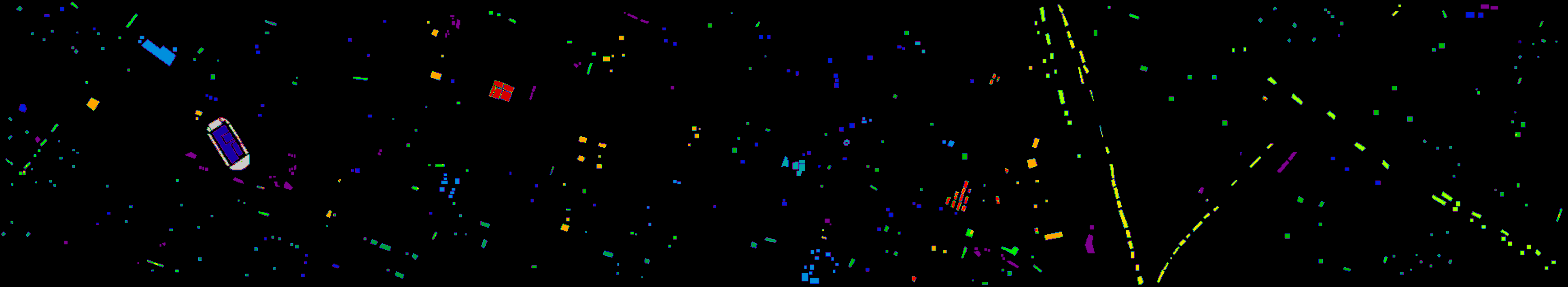}
            \caption{HybCNN}
        \end{subfigure}
        \begin{subfigure}{0.32\textwidth}
            \centering
            \includegraphics[width=0.99\textwidth]{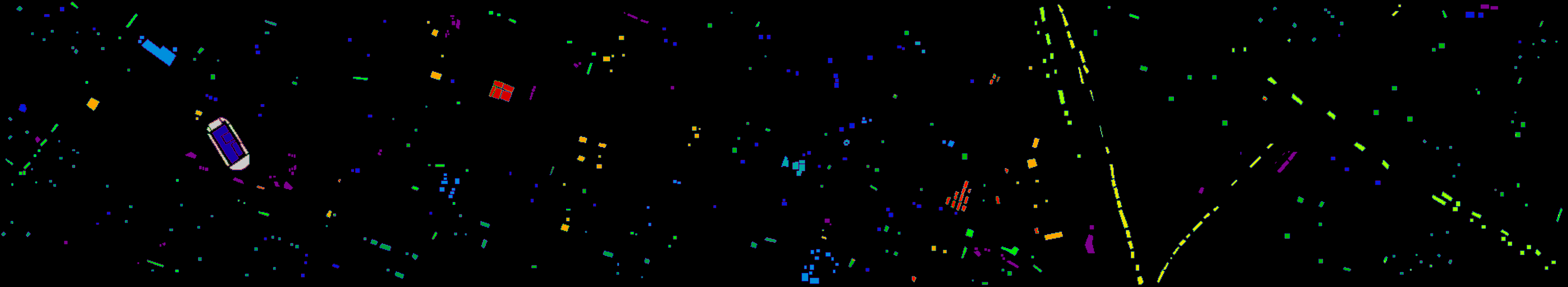}
            \caption{IN2D}
        \end{subfigure}
        \begin{subfigure}{0.32\textwidth}
            \centering
            \includegraphics[width=0.99\textwidth]{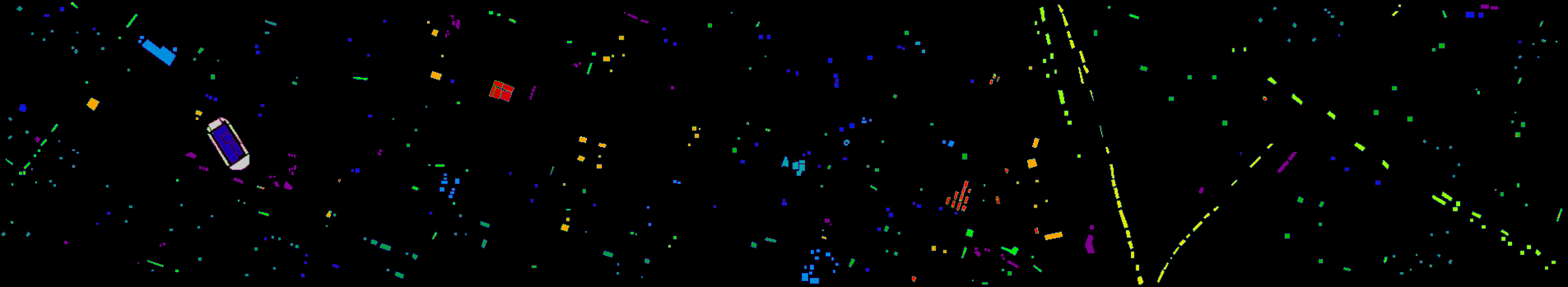}
            \caption{IN3D}
        \end{subfigure}
        \begin{subfigure}{0.32\textwidth}
            \centering
            \includegraphics[width=0.99\textwidth]{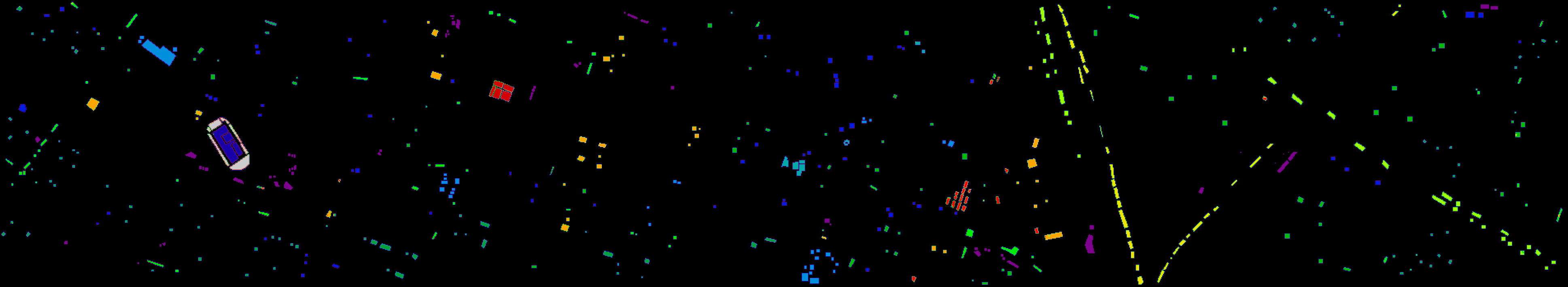}
            \caption{HybIN}
        \end{subfigure}
        \begin{subfigure}{0.32\textwidth}
            \centering
            \includegraphics[width=0.99\textwidth]{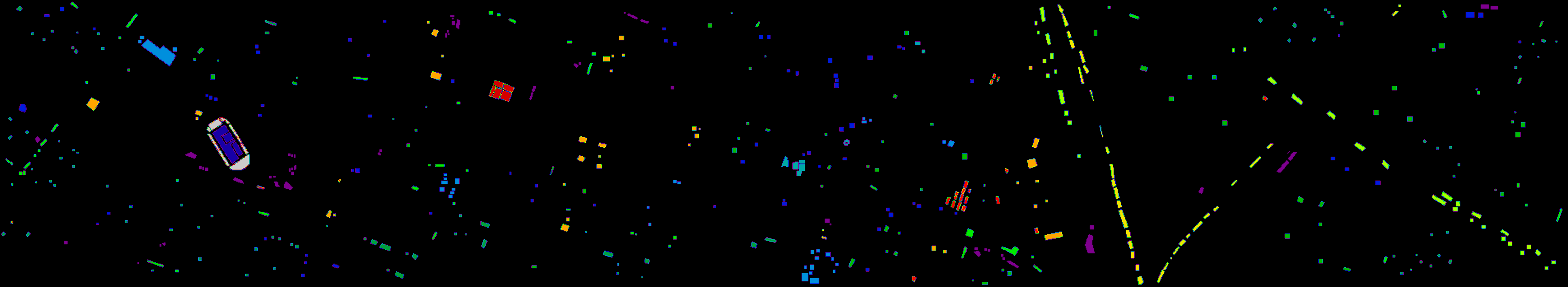}
            \caption{MorpCNN}
        \end{subfigure}
        \begin{subfigure}{0.32\textwidth}
            \centering
            \includegraphics[width=0.99\textwidth]{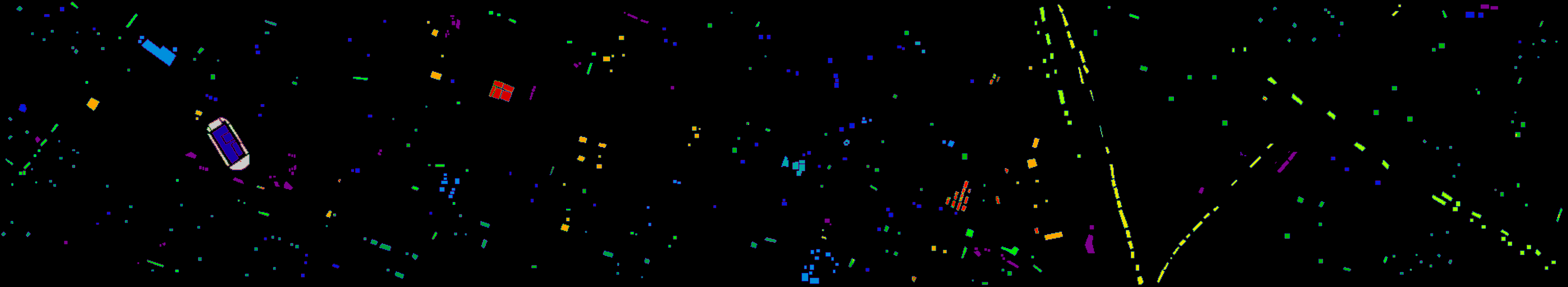}
            \caption{Hybrid-ViT}
        \end{subfigure}
        \begin{subfigure}{0.32\textwidth}
            \centering
            \includegraphics[width=0.99\textwidth]{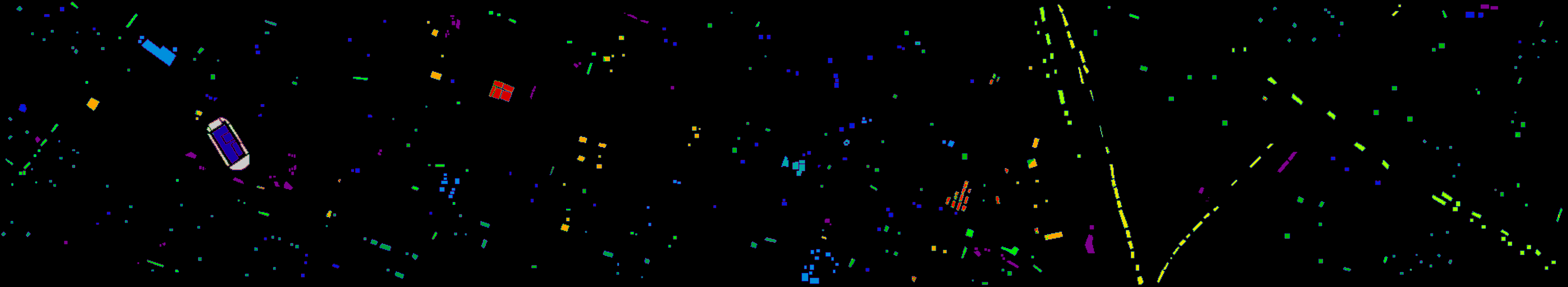}
            \caption{Hir-Transformer}
        \end{subfigure}
        \begin{subfigure}{0.32\textwidth}
            \centering
            \includegraphics[width=0.99\textwidth]{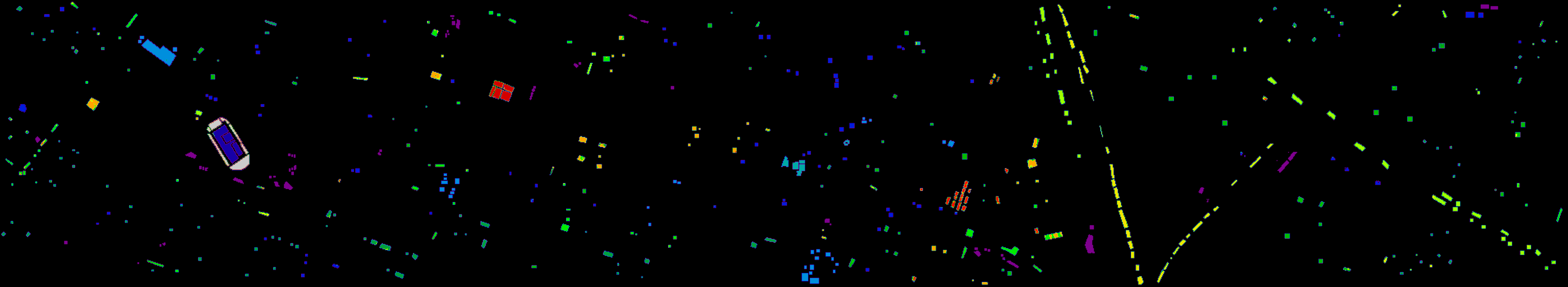}
            \caption{SSMamba}
        \end{subfigure}
        \begin{subfigure}{0.32\textwidth}
            \centering
            \includegraphics[width=0.99\textwidth]{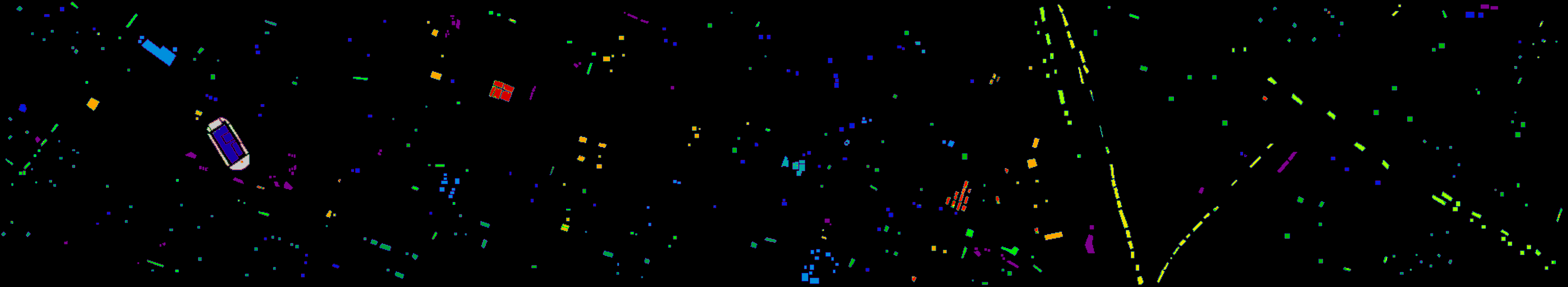}
            \caption{SSM}
        \end{subfigure}
        \begin{subfigure}{0.32\textwidth}
            \centering
            \includegraphics[width=0.99\textwidth]{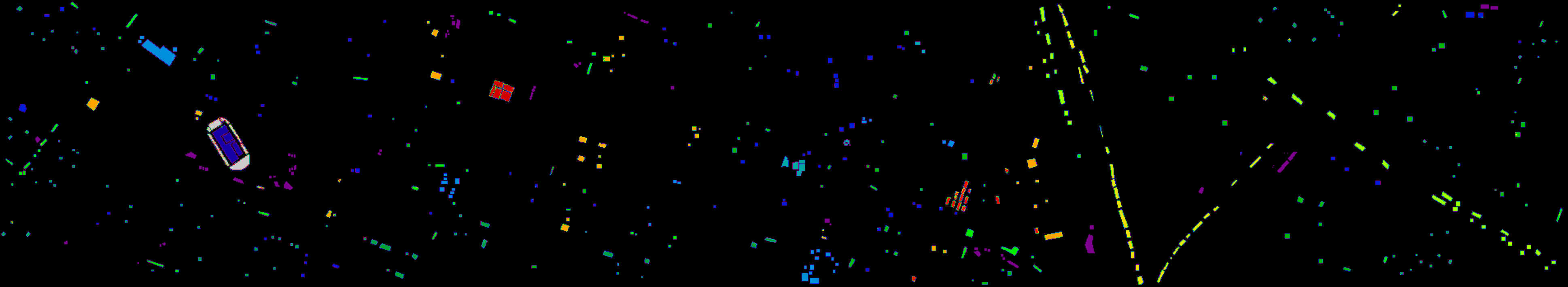}
            \caption{SSMM}
        \end{subfigure}
    \caption{\textbf{UH dataset:} The predicted ground truth maps for various competing methods alongside the proposed variants of the MorpMamba model.}
    \label{UH_results}
\end{figure*}

In short, the results from Table \ref{Tab2} demonstrate that the inclusion of morphological operations (erosion and dilation) in MorpMamba significantly improves its classification performance. This is evidenced by higher OA, AA, and $\kappa$ scores across all datasets. Additionally, the t-SNE visualizations further confirm the model’s superior feature learning capabilities.

\section{Comparison with SOTA Methods and Discussion}
\label{QRD2}

This section presents a detailed comparative analysis of the proposed MorpMamba model against state-of-the-art (SOTA) HSI classification models. The comparison focuses on key performance metrics: OA, AA, and the $\kappa$ coefficient. Additionally, we evaluate the computational efficiency by analyzing the number of parameters required for each model. Table \ref{my_label} summarizes these metrics across five prominent HSI datasets: the UH, LK, PU, PC, and SA. Furthermore, Figures \ref{LK_results}, \ref{PC_results}, \ref{PU_results}, \ref{SA_results}, and \ref{UH_results} illustrate the predicted ground truth maps for all competing methods, including the proposed MorpMamba model. In this context, OA represents the percentage of correctly classified samples out of the total samples, AA is the mean accuracy across all classes, and $\kappa$ is a statistical measure of agreement adjusted for a chance.

The comparative models included in our analysis are the 2D CNN \cite{yang2018hyperspectral}, 3D CNN \cite{ahmad2020fast}, HybCNN (a hybrid of 2D and 3D convolutions) \cite{9767615}, 2D IN (2D inception network) \cite{xiong2018ai}, 3D IN (3D inception network) \cite{10087213}, HybIN (a hybrid of 2D and 3D inception network) \cite{shwetha2019hybrid, firat2023hybrid}, MorpCNN \cite{9451651}, Hybrid-ViT (a hybrid vision Transformer) \cite{2330979}, Hir-Transformer (a hierarchical Transformer) \cite{10681622}, SSMamba, spatial morphological Mamba (SMM -- the proposed SMM model that integrates only spatial morphological operations with the Mamba architecture), and spatial-spectral morphological Mamba (SSMM -- the proposed SSMM model that integrates spatial-spectral morphological operations with the Mamba architecture). The detailed configurations for each competing method, including the number of layers and filters per layer, are applied according to the specifications provided in their respective papers.

To ensure a fair comparison, all methods were evaluated under consistent experimental conditions. A patch size of 4 was used, and 15 spectral bands were selected. The dataset was divided into training (20\%), validation (30\%), and test (50\%) sets. All models were trained for 50 epochs with a batch size of 256, using the Adam optimizer with a learning rate of 0.001. For the UH dataset, MorpMamba achieved an OA of 98.28\%, an AA of 97.91\%, and a $\kappa$ of 98.14 as shown in Table \ref{my_label}. Although the 3D CNN model exhibited a slightly higher OA of 99.01\% and AA of 98.81\%, MorpMamba demonstrated its competitive edge with significantly fewer parameters (67,013 compared to 4,042,751).

Similarly, on the LK dataset, MorpMamba attained an OA of 99.70\%, an AA of 99.25\%, and a $\kappa$ of 99.61. This performance is closely aligned with top-performing models like the 3D CNN (OA of 99.81\%) while showcasing remarkable computational efficiency with substantially fewer parameters (66,239 compared to 4,041,977). For the PU dataset, MorpMamba achieved an OA of 97.67\%, an AA of 96.93\%, and a $\kappa$ of 96.91, maintaining competitive performance against the 3D CNN model (OA of 98.70\%) with a lower parameter count (66,239). In the PC dataset, MorpMamba recorded an OA of 99.71\%, an AA of 98.85\%, and a $\kappa$ of 99.59, closely competing with the 3D CNN model (OA of 99.87\%) while having fewer parameters (66,239). Finally, on the SA dataset, MorpMamba achieved an OA of 98.52\%, an AA of 99.25\%, and a $\kappa$ of 98.35\%. While the 3D CNN slightly outperformed in OA (98.86\%), MorpMamba’s performance is particularly impressive given its reduced parameter count (67,142).

The MorpMamba model stands out due to its superior computational efficiency compared to Transformer-based models, which typically suffer from quadratic complexity as the sequence length increases. By maintaining linear complexity, MorpMamba ensures scalability to larger datasets while delivering competitive accuracy. The incorporation of morphological operations significantly enhances the robustness and stability of the model, effectively mitigating noise and highlighting structural features within HSI data. This strategic approach addresses challenges associated with high-dimensional data, resulting in consistent performance across diverse datasets. In summary, the comparative analysis underscores the strengths of the MorpMamba model in HSI classification. It consistently achieves high accuracy and efficiency, outperforming SOTA models while maintaining a lower computational footprint.

\section{Computational Complexity}

The computational complexity of the proposed MorpMamba model, as compared to various SOTA methods, can be analyzed by evaluating each major component in the architecture. The Erosion and Dilation layers, which utilize depthwise convolution to process the input tensor, exhibit a complexity of $O(C \cdot H \cdot W \cdot k^2)$ for each operation, where $C$ denotes the number of channels, and $k$ is the kernel size. Consequently, the spectral-spatial token generation combines the complexities of both spatial and spectral morphological operations, resulting in a total complexity of $O(C^2 \cdot H \cdot W \cdot k^2)$. This reflects the increased dimensionality when treating spectral bands as spatial dimensions during processing. The Multi-Head Self-Attention mechanism introduces a significant computational overhead, with a complexity of $O(N^2 \cdot D)$, where $N$ is the number of tokens and $D$ is the feature dimension. This stems from the need to compute attention scores for each token against all other tokens in the sequence, which can be particularly demanding for large datasets. Additionally, the token enhancement which dynamically adjusts the importance of spatial and spectral tokens, adds a complexity of $O(N \cdot D)$. The SSM further contributes with the complexity $O(T \cdot N \cdot D)$, $T$ represents the number of time steps or iterations for state updates. 

Overall, the dominant term in the MorpMamba model's complexity can be summarized as $O(N^2 \cdot D)$, indicating that while the model effectively handles the intricate relationships within high-dimensional hyperspectral data, it maintains a scalable architecture that facilitates improved classification accuracy. In contrast, the Mamba model without morphological operations operates at a lower complexity of $O(N \cdot D)$. This highlights the computational efficiency gained by integrating morphological operations, which not only enhance robustness and stability by reducing noise but also capture vital structural features in HSIs. Therefore, MorpMamba's ability to balance high accuracy and reduced parameter count positions it as a leading model for HSI classification, particularly in scenarios demanding computational efficiency.

\section{Conclusion and Future Research Directions}

This work introduced spatial morphological Mamba (SMM) and spatial-spectral morphological Mamba (SSMM) in short MorpMamba, a novel framework for HSI classification that integrates morphological spatial and spatial-spectral operations with the state space model architecture. MorpMamba demonstrated SOTA performance across multiple HSI datasets, achieving superior classification results in terms of OA, AA, and $\kappa$ coefficient, while significantly reducing computational overhead compared to CNN, transformer, and mamba-based models. The incorporation of morphological operations such as erosion and dilation in the tokenization process enhances the model's ability to extract both fine-grained details and global structural information in both spatial and spectral dimensions. The token enhancement module further refines these features by dynamically adjusting the significance of spatial and spectral tokens, resulting in more robust and context-aware representations. Additionally, the use of multi-head self-attention enables the model to effectively capture intricate relationships within the data, and the state space module models temporal dependencies efficiently.

Future research directions focus on advancing MorpMamba's capabilities in several impactful areas such as Domain Adaptation and Meta-Learning approaches will be systematically implemented to enhance MorpMamba's generalization across diverse HSI datasets originating from varied geographic regions and sensor platforms. Combining MorpMamba with additional remote sensing modalities, such as LiDAR and SAR, will enable the development of richer, multi-modal Earth observation models for advanced applications in environmental monitoring and Earth sciences. The extension of MorpMamba to multi-temporal HSI data will support tasks such as change detection and time-series analysis, leveraging its capacity to model temporal dynamics effectively. By pursuing these directions, MorpMamba will not only advance the field of HSI but also contribute to network science and multi-modal remote sensing.

\bibliographystyle{IEEEtran}
\bibliography{Sam}
\end{document}